\documentclass[10pt]{article} 
\usepackage[preprint]{tmlr}

\usepackage{hyperref}
\usepackage{url}

\usepackage{lipsum}
\usepackage{amsmath,amsfonts,amsthm}
\usepackage{graphicx}
\usepackage{epstopdf}
\usepackage{algorithmic}
\ifpdf
  \DeclareGraphicsExtensions{.eps,.pdf,.png,.jpg}
\else
  \DeclareGraphicsExtensions{.eps}
\fi

\usepackage{xspace}
\usepackage{bbm} 
\usepackage{placeins}
\usepackage{hyphenat} 
\usepackage{cleveref}

\usepackage[format=plain,labelformat=simple,labelsep=period,font=small,compatibility=false]{caption}
\usepackage[font=footnotesize,skip=3pt,subrefformat=parens]{subcaption}

\usepackage[inline]{enumitem}
\setlist[enumerate]{leftmargin=.5in}
\setlist[itemize]{leftmargin=.5in}

\newtheorem{theorem}{Theorem}
\newtheorem{proposition}{Proposition}
\newtheorem{lemma}{Lemma}
\newtheorem{corollary}{Corollary}

\theoremstyle{definition}
\newtheorem{definition}{Definition}
\newtheorem{hypothesis}{Hypothesis}
\newtheorem{remark}{Remark}

\crefname{theorem}{Theorem}{Theorems}
\crefname{proposition}{Proposition}{Propositions}
\crefname{lemma}{Lemma}{Lemmas}
\crefname{corollary}{Corollary}{Corollaries}
\crefname{definition}{Definition}{Definitions}
\crefname{hypothesis}{Hypothesis}{Hypotheses}
\crefname{remark}{Remark}{Remarks}
\crefname{section}{Section}{Sections}
\crefname{appendix}{Appendix}{Appendices}

\usepackage{amsopn}

\makeatletter
\DeclareRobustCommand\onedot{\futurelet\@let@token\@onedot}
\def\@onedot{\ifx\@let@token.\else.\null\fi\xspace}

\def\eg{\emph{e.g}\onedot} 
\def\ie{\emph{i.e}\onedot}

\makeatother

\newcommand{\iter}[2]{#1^{(#2)}}
\newcommand{\fouriertransf}[1]{\widehat{#1}}
\newcommand{\flipped}[1]{\overline{#1}}
\newcommand{\avgseq}[1]{{}\widetilde{#1}}

\newcommand{\mathC}{\mathbb{C}}
\newcommand{\mathE}{\mathbb{E}}

\newcommand{\mathN}{\mathbb{N}}
\newcommand{\mathP}{\mathbb{P}}

\newcommand{\mathR}{\mathbb{R}}
\newcommand{\mathS}{\mathbb{S}}

\newcommand{\mathZ}{\mathbb{Z}}
\newcommand{\mathone}{\mathbbm{1}}

\newcommand{\MH}{\mathcal H}

\newcommand{\MJ}{\mathcal J}

\newcommand{\MT}{\mathcal T}

\newcommand{\MV}{\mathcal V}

\newcommand{\FA}{\mathfrak A}

\newcommand{\FS}{\mathfrak S}

\newcommand{\FX}{\mathfrak X}
\newcommand{\FY}{\mathfrak Y}

\newcommand{\bh}{\boldsymbol h}

\newcommand{\bu}{\boldsymbol u}
\newcommand{\bv}{\boldsymbol v}

\newcommand{\BBD}{\mathbf D}

\newcommand{\BBG}{\mathbf G}

\newcommand{\BBW}{\mathbf W}
\newcommand{\BBX}{\mathbf X}

\newcommand{\BFS}{\boldsymbol\FS}
\newcommand{\BFX}{\boldsymbol\FX}
\newcommand{\BFY}{\boldsymbol\FY}

\newcommand{\rmd}{\mathrm{d}}
\newcommand{\rme}{\mathrm{e}}
\newcommand{\rmg}{\mathrm{g}}
\newcommand{\rmh}{\mathrm{h}}

\newcommand{\RMD}{\mathrm{D}}

\newcommand{\RMG}{\mathrm{G}}

\newcommand{\RMI}{\mathrm{I}}

\newcommand{\RMU}{\mathrm{U}}
\newcommand{\RMV}{\mathrm{V}}
\newcommand{\RMW}{\mathrm{W}}
\newcommand{\RMX}{\mathrm{X}}
\newcommand{\RMY}{\mathrm{Y}}

\newcommand{\SFF}{\mathsf{F}}
\newcommand{\SFG}{\mathsf{G}}
\newcommand{\SFH}{\mathsf{H}}
\newcommand{\SFM}{\mathsf{M}}
\newcommand{\SFP}{\mathsf{P}}
\newcommand{\SFQ}{\mathsf{Q}}
\newcommand{\SFR}{\mathsf{R}}
\newcommand{\SFS}{\mathsf{S}}

\newcommand{\SFZ}{\mathsf{Z}}

\newcommand{\SFDelta}{\mathsf{\Delta}}

\newcommand{\btau}{\boldsymbol \tau}

\newcommand{\bzero}{\boldsymbol 0}

\newcommand{\iif}{\;\Longleftrightarrow\;}

\newcommand{\red}[1]{\textcolor{red}{#1}}

\newcommand{\dtcwpt}{DT-$\mathC$WPT\xspace}

\newcommand{\range}[2]{\left\{#1,\, \dots,\, #2\right\}}
\newcommand{\bigrange}[2]{\bigl\{#1 ,\, \dots,\, #2\bigr\}}

\newcommand{\zeroto}[1]{\range{0}{#1}}
\newcommand{\bigzeroto}[1]{\bigrange{0}{#1}}
\newcommand{\binaryset}{\{0,\, 1\}}

\newcommand{\pair}[2]{\left(#1,\, #2\right)}

\newcommand{\set}[2]{\left\{ #1 \;\middle|\; #2 \right\}}
\newcommand{\bigset}[2]{\bigl\{ #1 \bigm| #2 \bigr\}}
\newcommand{\Bigset}[2]{\Bigl\{ #1 \Bigm| #2 \Bigr\}}

\newcommand{\interval}[2]{\left[#1,\, #2\right]}
\newcommand{\intervalexcll}[2]{\left]#1,\, #2\right]}
\newcommand{\intervalexclr}[2]{\left[#1,\, #2\right[}

\newcommand{\zeropi}{\interval{0}{\pi}}
\newcommand{\zerotwopi}{\interval{0}{2\pi}}

\newcommand{\zerotwopiexcl}{\intervalexclr{0}{2\pi}}
\newcommand{\zeroexcltwopi}{\intervalexcll{0}{2\pi}}

\newcommand{\mpipi}{\interval{-\pi}{\pi}}

\newcommand{\zerotopi}{\interval{0}{\pi}}

\newcommand{\tuple}[2]{\{#1,\, #2\}}

\newcommand{\nonzeroset}[1]{#1 \setminus \{0\}}
\newcommand{\nonzeroMathN}{\nonzeroset{\mathN}}
\newcommand{\nonzeroMathR}{\nonzeroset{\mathR}}
\newcommand{\nonzeroMathC}{\nonzeroset{\mathC}}
\newcommand{\nonzeroMathRplus}{\mathR_{>0}}

\newcommand{\Proba}{\mathP}
\newcommand{\Expval}{\mathE}

\newcommand{\condproba}[2]{\Proba\set{#1}{#2}}
\newcommand{\bigcondproba}[2]{\Proba\bigset{#1}{#2}}
\newcommand{\Bigcondproba}[2]{\Proba\Bigset{#1}{#2}}

\newcommand{\condexpval}[2]{\Expval\left[ #1 \;\middle|\; #2 \right]}

\newcommand{\Bigcondexpval}[2]{\Expval\Bigl[ #1 \Bigm| #2 \Bigr]}

\newcommand{\innerprod}[2]{\left\langle #1,\, #2 \right\rangle}

\newcommand{\smallO}{o}

\renewcommand{\arraystretch}{1.5} 

\renewcommand{\texttt}[1]{%
	\begingroup
	\ttfamily
	\begingroup\lccode`~=`/\lowercase{\endgroup\def~}{/\discretionary{}{}{}}%
	\begingroup\lccode`~=`[\lowercase{\endgroup\def~}{[\discretionary{}{}{}}%
	\begingroup\lccode`~=`.\lowercase{\endgroup\def~}{.\discretionary{}{}{}}%
	\catcode`/=\active\catcode`[=\active\catcode`.=\active
	\scantokens{#1\noexpand}%
	\endgroup
} 

\newcommand{\norm}[1]{\left\|#1\right\|}

\newcommand{\normone}[1]{\norm{#1}_1}
\newcommand{\normtwo}[1]{\norm{#1}_2}
\newcommand{\norminfty}[1]{\norm{#1}_{\infty}}
\newcommand{\normLtwo}[1]{\norm{#1}_{L^2}}

\newcommand{\bignorm}[1]{\bigl\|#1\bigr\|}

\newcommand{\bignormtwo}[1]{\bignorm{#1}_2}
\newcommand{\bignorminfty}[1]{\bignorm{#1}_{\infty}}
\newcommand{\bignormLtwo}[1]{\bignorm{#1}_{L^2}}

\newcommand{\Bignorm}[1]{\Bigl\|#1\Bigr\|}

\newcommand{\BignormLtwo}[1]{\Bignorm{#1}_{L^2}}

\newcommand\indep{\protect\mathpalette{\protect\independenT}{\perp}}
\def\independenT#1#2{\mathrel{\rlap{$#1#2$}\mkern2mu{#1#2}}}

\newcommand{\obo}{1 \times 1}

\newcommand{\qqand}{\qquad\mbox{and}\qquad}
\newcommand{\qqor}{\qquad\mbox{or}\qquad}

\newcommand{\qqwith}{\qquad\mbox{with}\qquad}

\newcommand{\todo}[1][]{%
  	\red{\ifthenelse{\equal{#1}{}}%
    	{[TODO]}%
    	{[TODO: #1]}%
  	}\xspace
}

\DeclareMathOperator{\supp}{supp}

\DeclareMathOperator{\sign}{sgn}

\DeclareMathOperator{\maxpool}{MaxPool}

\DeclareMathOperator{\Real}{Re}

\DeclareMathOperator{\idxmod}{mod}

\DeclareMathOperator{\idxmax}{max}

\DeclareMathOperator{\idxlum}{lum}

\newcommand{\topright}{\nearrow}
\newcommand{\bottomright}{\searrow}
\newcommand{\topleft}{\nwarrow}
\newcommand{\bottomleft}{\swarrow}

\newcommand{\rmax}{$\mathR$Max\xspace}
\newcommand{\cmod}{$\mathC$Mod\xspace}

\newcommand{\xcoord}{$x$\xspace}

\newcommand{\sigmaAlgebra}{$\sigma$-algebra\xspace}

\newcommand{\subspace}{\MV}

\newcommand{\discreteSubspace}{\MJ}

\newcommand{\ball}[1]{B_{#1}}
\newcommand{\probaspace}{E}

\newcommand{\unitcircle}{\mathS^1}

\newcommand{\measurablesetUnitcircle}{\FA}

\newcommand{\measurablesetRealaxis}{\FX}
\newcommand{\measurablesetRealposmultaxes}{\BFS}
\newcommand{\measurablesetRealmultaxes}{\BFX}
\newcommand{\measurablesetRealmultaxesbis}{\BFY}

\newcommand{\setof}[1]{\zeroto{#1 - 1}}
\newcommand{\bigsetof}[1]{\bigzeroto{#1 - 1}}

\newcommand{\poolinggrid}{\range{-\gridhalfsize}{\gridhalfsize}^2}

\newcommand{\linftyBall}{\ball{\infty}}

\newcommand{\fourierWindowGaborContinuous}{\linftyBall(\supportsizeContinuous/2)}
\newcommand{\fourierWindowSampling}{\linftyBall(\pi/\samplinterv)}

\newcommand{\spatialWindow}{\linftyBall(\spatialWindowsize)}
\newcommand{\spatialWindowDiscretegrid}{\linftyBall(\spatialWindowsizeDiscretegrid)}

\newcommand{\shannonspaceParam}[1]{\iter{\subspace}{#1}} 
\newcommand{\shannonspaceSamplinterv}{\shannonspaceParam{\samplinterv}}
\newcommand{\shannonspaceSubsamplinterv}{\shannonspaceParam{\subsamplinterv}}

\newcommand{\setofFBs}{\zeroto{3}} 

\newcommand{\nexamples}{N}

\newcommand{\inchannels}{K}
\newcommand{\outchannels}{L}
\newcommand{\normdegree}{p}

\newcommand{\sub}{m} 

\newcommand{\samplinterv}{s}
\newcommand{\imgsize}{N}

\newcommand{\gridhalfsize}{q}

\newcommand{\depth}{J}

\newcommand{\supportsizeContinuous}{\varepsilon}
\newcommand{\supportsizeDiscrete}{\kappa}
\newcommand{\spatialWindowsize}{r}

\newcommand{\subsamplinterv}{\samplinterv'}

\newcommand{\subDepth}{\sub_\depth}

\newcommand{\supportsizeDiscreteDepth}{\supportsizeDiscrete_\depth}

\newcommand{\spatialWindowsizeDiscretegrid}{\spatialWindowsize_\gridhalfsize}
\newcommand{\nevalpointsGrid}{\nevalpoints_\gridhalfsize}

\newcommand{\selectExample}{n}

\newcommand{\selectInchannel}{k}
\newcommand{\selectOutchannel}{l}
\newcommand{\selectDepth}{j}
\newcommand{\selectEvalpoint}{i}

\newcommand{\selectFB}{m}
\newcommand{\selectFBOned}{i}
\newcommand{\selectFBOnedbis}{j}

\newcommand{\scalarindex}{n}
\newcommand{\scalarindexbis}{p}
\newcommand{\colormixval}{\mu}
\newcommand{\permutval}{\sigma}

\newcommand{\angleval}{\eta}
\newcommand{\anglevalbis}{\omega}
\newcommand{\pointunitcircle}{z}
\newcommand{\nevalpoints}{n}
\newcommand{\wavelength}{\lambda}
\newcommand{\freqval}{\nu}
\newcommand{\freqvalbis}{\xi}
\newcommand{\freqvalMpipi}{\theta}
\newcommand{\freqvalMpipibis}{\omega}
\newcommand{\boundVectorindex}{n}

\newcommand{\timeval}{t}

\newcommand{\spatialval}{x}
\newcommand{\spatialvalbis}{y}

\newcommand{\energyratio}{\rho}
\newcommand{\localityparam}{\tau}

\newcommand{\colormixvalSelectOutIn}{\colormixval_{\selectOutInchannel}}

\newcommand{\diffPhase}{{\delta\!\phase}}

\newcommand{\selectOutInchannel}{\selectOutchannel\selectInchannel}

\newcommand{\energyratioSelect}{\energyratio_{\selectOutchannel}}

\newcommand{\colormixvec}{\boldsymbol{\colormixval}}

\newcommand{\permutvec}{\boldsymbol{\permutval}}

\newcommand{\permutvecDepthSelect}{\iter{\permutvec}{\depth}_\selectOutchannel}

\newcommand{\translvecContinuous}{\bh}
\newcommand{\translvecDiscrete}{\bu}
\newcommand{\translvecDiscreteBis}{\bv}
\newcommand{\unitvec}{\bu}

\newcommand{\freqvec}{\boldsymbol{\freqval}}
\newcommand{\freqvecbis}{\boldsymbol{\freqvalbis}}
\newcommand{\freqvecMpipi}{\boldsymbol{\freqvalMpipi}} 
\newcommand{\freqvecMpipiSelect}{\freqvecMpipi_\selectOutchannel}
\newcommand{\freqvecMpipibis}{\boldsymbol{\freqvalMpipibis}}
\newcommand{\freqvecMpipiMultires}[1]{\iter{\freqvecMpipi}{#1}}

\newcommand{\spatialvec}{\boldsymbol{\spatialval}}
\newcommand{\spatialvecbis}{\boldsymbol{\spatialvalbis}}
\newcommand{\vectorindex}{\boldsymbol{\scalarindex}}
\newcommand{\vectorindexbis}{\boldsymbol{\scalarindexbis}}
\newcommand{\searchvec}{\translvecContinuous}

\newcommand{\spatialvecSelect}{\spatialvec_{\vectorindex}}
\newcommand{\searchvecSelect}{\searchvec_{\vectorindexbis}}
\newcommand{\searchvecMaxSelect}{\searchvec^{\idxmax}_{\vectorindex}}
\newcommand{\searchvecApproxMaxSelect}{{\searchvec'}^{\idxmax}_{\vectorindex}}

\newcommand{\waveletOned}{\psi}

\newcommand{\complexphasefun}{u}
\newcommand{\modulusfun}{v}

\newcommand{\modulusmultfun}{\boldsymbol\modulusfun}

\newcommand{\inpfun}{F}
\newcommand{\inpfunbis}{\inpfun_1}
\newcommand{\wavelet}{\varPsi}
\newcommand{\scalingfun}{\varPhi}
\newcommand{\complexPhase}{Z}
\newcommand{\phase}{H}
\newcommand{\cosfun}{G}

\newcommand{\exactfun}{A}

\newcommand{\approxfun}{{}\widetilde{\exactfun}}
\newcommand{\exactfunInterp}{\exactfun_{\inpimg}}
\newcommand{\approxfunInterp}{\approxfun_{\inpimg}}
\newcommand{\lowfreqfun}{\inpfun_0}

\newcommand{\fourierLowfreqfun}{\fouriertransf{\lowfreqfun}}

\newcommand{\fourierStochLowfreqfunInterp}{\fouriertransf{\stochInpfun}_{0,\, \inpimg}}
\newcommand{\interpshannon}[2]{{\inpfun_{#1}^{(#2)}}}

\newcommand{\shanScalingfunParam}[1]{\iter{\scalingfun}{#1}} 
\newcommand{\shanScalingfunSamplinterv}{\shanScalingfunParam{\samplinterv}}
\newcommand{\fourierInpfun}{\fouriertransf{\inpfun}}

\newcommand{\evalExactfunInterp}{
    \exactfunInterp\bigl(
		\spatialvecSelect,\, \searchvecMaxSelect
	\bigr)
}
\newcommand{\maxExactfunInterp}{\exactfunInterp^{\idxmax}}
\newcommand{\evalMaxExactfunInterp}{\maxExactfunInterp(\spatialvecSelect)}
\newcommand{\evalApproxfunInterp}{
    \approxfunInterp\bigl(
		\spatialvecSelect,\, \searchvecApproxMaxSelect
	\bigr)
}
\newcommand{\maxApproxfunInterp}{\approxfunInterp^{\idxmax}}
\newcommand{\evalMaxApproxfunInterp}{\maxApproxfunInterp(\spatialvecSelect)}

\newcommand{\flippedWavelet}{\flipped{\wavelet}}

\newcommand{\fourierWaveletOned}{\fouriertransf{\waveletOned}}
\newcommand{\fourierWavelet}{\fouriertransf{\wavelet}}
\newcommand{\fourierFlippedWavelet}{\fouriertransf{\flippedWavelet}}

\newcommand{\cosfunInpimg}{\cosfun_\inpimg}

\newcommand{\multwavelet}{\boldsymbol{\wavelet}}

\newcommand{\inpfunInterp}{\inpfun_\inpimg}

\newcommand{\inpfunTranslInterp}{{\inpfun_{\translInpimg}}}
\newcommand{\lowfreqfunInterpSubsampl}{\interpshannon{\lowfreqimg}{\subsamplinterv}}
\newcommand{\lowfreqfunTranslInterpSubsampl}{\interpshannon{\translSubvecLowfreqimg}{\subsamplinterv}}

\newcommand{\waveletInterp}{\wavelet_\complexWeightimg}

\newcommand{\flippedWaveletInterp}{\flippedWavelet_\complexWeightimg}

\newcommand{\fourierWaveletInterp}{\fourierWavelet_\complexWeightimg}

\newcommand{\phaseInterp}{\phase_\inpimg}
\newcommand{\complexPhaseInterp}{\complexPhase_\inpimg}

\newcommand{\shanScalingfunSamplintervVectorindex}{\shanScalingfunSamplinterv_{\vectorindex}}
\newcommand{\shanScalingfunSubsamplintervVectorindex}{\shanScalingfunParam{\subsamplinterv}_{\vectorindex}}

\newcommand{\waveletOnedMultires}[2]{\iter{\waveletOned}{#1}_{#2}}
\newcommand{\waveletOnedFBMultires}[3]{\waveletOned^{[#1](#2)}_{#3}}
\newcommand{\waveletOnedRealMultires}[2]{\waveletOnedFBMultires{\realFB}{#1}{#2}}
\newcommand{\waveletOnedImagMultires}[2]{\waveletOnedFBMultires{\imagFB}{#1}{#2}}

\newcommand{\waveletNEMultiresZero}[2]{\wavelet^{\topright(#1)}_{#2}}
\newcommand{\waveletSEMultiresZero}[2]{\wavelet^{\bottomright(#1)}_{#2}}
\newcommand{\waveletSWMultiresZero}[2]{\wavelet^{\bottomleft(#1)}_{#2}}
\newcommand{\waveletNWMultiresZero}[2]{\wavelet^{\topleft(#1)}_{#2}}

\newcommand{\waveletNEMultires}[3]{\wavelet^{\topright(#1)}_{#2,\, #3}}
\newcommand{\waveletSEMultires}[3]{\wavelet^{\bottomright(#1)}_{#2,\, #3}}
\newcommand{\waveletSWMultires}[3]{\wavelet^{\bottomleft(#1)}_{#2,\, #3}}
\newcommand{\waveletNWMultires}[3]{\wavelet^{\topleft(#1)}_{#2,\, #3}}

\newcommand{\flippedWaveletNEMultiresZero}[2]{\flippedWavelet^{\topright(#1)}_{#2}}

\newcommand{\waveletMultiresComplexFamily}[2]{
    \bigl(
        \waveletNEMultires{#1}{#2}{\vectorindex},\,\allowbreak
        \waveletSEMultires{#1}{#2}{\vectorindex},\,\allowbreak
        \waveletSWMultires{#1}{#2}{\vectorindex},\,\allowbreak
        \waveletNWMultires{#1}{#2}{\vectorindex}
    \bigr)_{\vectorindexInZtwo}
}

\newcommand{\multwaveletFinal}[1]{\iter{\multwavelet}{#1}}
\newcommand{\complexMultwaveletFinal}[1]{\multwaveletFinal{#1}_{\mathC}}

\newcommand{\fourierWaveletOnedMultires}[2]{\iter{\fourierWaveletOned}{#1}_{#2}}

\newcommand{\fourierWaveletOnedFBMultires}[3]{\fourierWaveletOned^{[#1](#2)}_{#3}}
\newcommand{\fourierWaveletOnedRealMultires}[2]{\fourierWaveletOnedFBMultires{\realFB}{#1}{#2}}

\newcommand{\inpfunAstWavelet}{\inpfun \ast \flippedWavelet}
\newcommand{\inpfunAstRealWavelet}{\inpfun \ast \Real\flippedWavelet}
\newcommand{\inpfunAstWaveletInterp}{\inpfunInterp \ast \flippedWaveletInterp}

\newcommand{\inpfunAstWaveletTranslInterp}{\inpfunTranslInterp \ast \flippedWaveletInterp}
\newcommand{\inpfunAstRealWaveletInterp}{\inpfunInterp \ast \Real\flippedWaveletInterp}

\newcommand{\deltaop}{\delta}
\newcommand{\scatterop}{U}

\newcommand{\transl}{\MT}

\newcommand{\hilberttransf}{\MH}

\newcommand{\rmaxop}{\scatterop^{\idxmax}}
\newcommand{\rmaxopNEDepth}{\scatterop^{\idxmax\topright}}
\newcommand{\rmaxopSEDepth}{\scatterop^{\idxmax\bottomright}}

\newcommand{\cmodop}{\scatterop^{\idxmod}}
\newcommand{\cmodopNEDepth}{\scatterop^{\idxmod\topright}}
\newcommand{\cmodopSEDepth}{\scatterop^{\idxmod\bottomright}}

\newcommand{\rmaxopWavelet}{\rmaxop_{\spatialWindowsize}}
\newcommand{\rmaxopSubGrid}{\rmaxop_{\sub,\, \gridhalfsize}}
\newcommand{\rmaxopSubSelect}{\rmaxop_{\sub,\, \selectOutchannel}}
\newcommand{\rmaxopSubGridSelect}{\rmaxop_{\sub,\, \gridhalfsize,\, \selectOutchannel}}
\newcommand{\rmaxopSubGridWeight}{\rmaxopSubGrid\!\left[\complexWeightimg\right]}
\newcommand{\rmaxopSubGridSelectWeight}{\rmaxopSubGridSelect\!\left[\complexWeightmultimg\right]}
\newcommand{\rmaxopWinWavelet}{\rmaxop_{\spatialWindowsize}\!\left[\wavelet\right]}
\newcommand{\rmaxopNEDepthSelect}{\rmaxopNEDepth_{\selectOutchannel}}
\newcommand{\rmaxopSEDepthSelect}{\rmaxopSEDepth_{\selectOutchannel}}

\newcommand{\cmodopSub}{\cmodop_{\sub}}
\newcommand{\cmodopSubSelect}{\cmodop_{\sub,\, \selectOutchannel}}
\newcommand{\cmodopSubWeight}{\cmodopSub\!\left[\complexWeightimg\right]}
\newcommand{\cmodopSubSelectWeight}{\cmodopSubSelect\!\left[\complexWeightmultimg\right]}
\newcommand{\cmodopWavelet}{\cmodop\!\left[\wavelet\right]}
\newcommand{\cmodopNEDepthSelect}{\cmodopNEDepth_{\selectOutchannel}}
\newcommand{\cmodopSEDepthSelect}{\cmodopSEDepth_{\selectOutchannel}}

\newcommand{\deltaopSubGrid}{\deltaop_{\sub,\, \gridhalfsize}}

\newcommand{\translSamplintervVec}{\transl_{\samplinterv\translvecDiscrete}}
\newcommand{\translSubsamplintervSubvec}{\transl_{\subsamplinterv\translvecDiscrete'}}
\newcommand{\translSubvec}{\transl_{\translvecDiscrete'}}

\newcommand{\shiftinvCmod}{\alpha}
\newcommand{\proxyfunRmaxCmod}{\gamma}
\newcommand{\lowerboundDiffRmaxCmod}{\beta}
\newcommand{\angularMeasure}{\vartheta}
\newcommand{\cosUnitcircle}{g}
\newcommand{\probaMeasure}{\mu}
\newcommand{\characfun}{\mathone}

\newcommand{\maxCosUnitcircle}{\cosUnitcircle_{\idxmax}}

\newcommand{\selectInchannelInrange}{\selectInchannel \in \setof{\inchannels}}

\newcommand{\selectOutchannelInrange}{\selectOutchannel \in \setof{\outchannels}}

\newcommand{\selectOutchannelInrangeWPT}{\selectOutchannel \in \setof{4^\depth}}
\newcommand{\selectOutchannelInbigrangeWPT}{\selectOutchannel \in \bigsetof{4^\depth}}

\newcommand{\selectFBInrange}{\selectFB \in \setofFBs}

\newcommand{\sumof}[2]{\sum_{#1 = 0}^{#2 - 1}}

\newcommand{\sumoverInchannels}{\sumof{\selectInchannel}{\inchannels}}

\newcommand{\sumoverEvalpointsGrid}{\sumof{\selectEvalpoint}{\nevalpointsGrid}}

\newcommand{\sumoverVectorindices}{\sum_{\vectorindexInZtwo}}
\newcommand{\sumoverVectorindicesbis}{\sum_{\vectorindexbisInZtwo}}

\newcommand{\sumoverArbitrarygrid}{\sum_{\norminfty{\vectorindexbis} \leq \boundVectorindex}}

\newcommand{\vectorindexInZtwo}{\vectorindex \in \mathZ^2}
\newcommand{\vectorindexbisInZtwo}{\vectorindexbis \in \mathZ^2}
\newcommand{\vectorindexInPoolinggrid}{\vectorindexbis \in \poolinggrid}
\newcommand{\pointInSetofEvalpointsGrid}{\selectEvalpoint \in \setof{\nevalpointsGrid}}

\newcommand{\maxVectorindexInPoolinggrid}{\max_{\norminfty{\vectorindexbis} \leq \gridhalfsize}}

\newcommand{\maxSearchvecInSpatialwindow}{\max_{\norminfty{\searchvec} \leq \spatialWindowsize}}

\newcommand{\realFB}{0}
\newcommand{\imagFB}{1}

\newcommand{\qmfLow}{\rmh}
\newcommand{\qmfHigh}{\rmg}

\newcommand{\qmfLowFB}[1]{\qmfLow^{[#1]}}

\newcommand{\qmfHighFB}[1]{\qmfHigh^{[#1]}}

\newcommand{\qmfLowReal}{\qmfLowFB{\realFB}}
\newcommand{\qmfHighReal}{\qmfHighFB{\realFB}}

\newcommand{\qmfLowImag}{\qmfLowFB{\imagFB}}
\newcommand{\qmfHighImag}{\qmfHighFB{\imagFB}}

\newcommand{\fourierQmfLow}{\fouriertransf{\qmfLow}}

\newcommand{\fourierQmfLowFB}[1]{\fourierQmfLow^{[#1]}}

\newcommand{\fourierQmfLowReal}{\fourierQmfLowFB{\realFB}}
\newcommand{\fourierQmfLowImag}{\fourierQmfLowFB{\imagFB}}

\newcommand{\inpimg}{\RMX}
\newcommand{\inpimgbis}{\inpimg_1}
\newcommand{\outimg}{\RMY}

\newcommand{\weightimg}{\RMV}
\newcommand{\complexWeightimg}{\RMW}

\newcommand{\wptimg}{\RMD}

\newcommand{\idimg}{\RMI}

\newcommand{\twodqmfHigh}{\RMG}

\newcommand{\inpmultimg}{\BBX}

\newcommand{\complexWeightmultimg}{\BBW}
\newcommand{\dtmultimg}{\BBD}
\newcommand{\twodmultqmf}{\BBG}

\newcommand{\colormiximg}{\inpimg^{\idxlum}}
\newcommand{\colormiximgSelect}{\colormiximg_\selectOutchannel}
\newcommand{\rmaximg}{\outimg^{\idxmax}}
\newcommand{\cmodimg}{\outimg^{\idxmod}}

\newcommand{\rmaximgNEMultires}[1]{\outimg^{\idxmax\topright}_{#1}}

\newcommand{\cmodimgNEMultires}[1]{\outimg^{\idxmod\topright}_{#1}}

\newcommand{\rmaximgbisNEMultires}[1]{\outimg^{\idxmax'\topright}_{#1}}

\newcommand{\cmodimgbisNEMultires}[1]{\outimg^{\idxmod'\topright}_{#1}}

\newcommand{\rmaximgNEMultiresDepthSelect}{\rmaximgNEMultires{\selectExample\selectOutchannel}}

\newcommand{\cmodimgNEMultiresDepthSelect}{\cmodimgNEMultires{\selectExample\selectOutchannel}}

\newcommand{\rmaximgbisNEMultiresDepthSelect}{\rmaximgbisNEMultires{\selectExample\selectOutchannel}}

\newcommand{\cmodimgbisNEMultiresDepthSelect}{\cmodimgbisNEMultires{\selectExample\selectOutchannel}}

\newcommand{\wptimgMultires}[2]{\iter{\wptimg}{#1}_{#2}}

\newcommand{\wptimgNEMultires}[2]{\wptimg^{\topright(#1)}_{#2}}
\newcommand{\wptimgSEMultires}[2]{\wptimg^{\bottomright(#1)}_{#2}}
\newcommand{\wptimgSWMultires}[2]{\wptimg^{\bottomleft(#1)}_{#2}}
\newcommand{\wptimgNWMultires}[2]{\wptimg^{\topleft(#1)}_{#2}}

\newcommand{\wptimgFBMultires}[3]{\wptimg^{[#1](#2)}_{#3}}

\newcommand{\lowfreqimg}{\inpimg_0}

\newcommand{\inpimgSelect}{\inpimg_\selectInchannel}

\newcommand{\rmaximgSelect}{\rmaximg_\selectOutchannel}
\newcommand{\cmodimgSelect}{\cmodimg_\selectOutchannel}

\newcommand{\twodmultqmfFB}[1]{\twodmultqmf^{[#1]}}
\newcommand{\inpimgFB}[1]{\inpimg^{[#1]}}

\newcommand{\weightimgMultires}[1]{\iter{\weightimg}{#1}}

\newcommand{\complexWeightimgNEMultires}[1]{\complexWeightimg^{\topright (#1)}}
\newcommand{\complexWeightimgSEMultires}[1]{\complexWeightimg^{\bottomright (#1)}}
\newcommand{\complexWeightimgSWMultires}[1]{\complexWeightimg^{\bottomleft (#1)}}
\newcommand{\complexWeightimgNWMultires}[1]{\complexWeightimg^{\topleft (#1)}}

\newcommand{\complexWeightmultimgNEMultires}[1]{\complexWeightmultimg^{\topright (#1)}}
\newcommand{\complexWeightmultimgSEMultires}[1]{\complexWeightmultimg^{\bottomright (#1)}}

\newcommand{\complexWeightimgSelect}{\complexWeightimg_{\selectOutInchannel}}

\newcommand{\avgComplexWeightimg}{\avgseq{\complexWeightimg}}

\newcommand{\avgComplexWeightimgSelect}{\avgComplexWeightimg_{\selectOutchannel}}

\newcommand{\flippedWeightimg}{\flipped{\weightimg}}
\newcommand{\flippedComplexWeightimg}{\flipped{\complexWeightimg}}
\newcommand{\flippedComplexWeightimgSelect}{
    \flippedComplexWeightimg_{\selectOutInchannel}
}

\newcommand{\flippedWeightimgMultires}[1]{\iter{\flippedWeightimg}{#1}}

\newcommand{\flippedComplexWeightimgNEMultires}[1]{\flippedComplexWeightimg^{\topright (#1)}}

\newcommand{\fourierInpimg}{\fouriertransf{\inpimg}}

\newcommand{\fourierWeightimg}{\fouriertransf{\weightimg}}
\newcommand{\fourierComplexWeightimg}{\fouriertransf{\complexWeightimg}}

\newcommand{\fourierComplexWeightimgNEMultires}[1]{\fourierComplexWeightimg^{\topright(#1)}}

\newcommand{\weightimgFBMultires}[3]{\weightimg^{[#1](#2)}_{#3}}
\newcommand{\flippedWeightimgFBMultires}[3]{\flippedWeightimg^{[#1](#2)}_{#3}}

\newcommand{\detailCmodimg}{\cmodopSub\inpimg}

\newcommand{\detailRmaximg}{\rmaxopSubGrid\inpimg}

\newcommand{\wptimgMultiresSelectbis}{\wptimgMultires{\selectDepth}{\selectOutchannel}}

\newcommand{\translInpimg}{\transl_{\translvecDiscrete}\inpimg}
\newcommand{\translSubvecLowfreqimg}{\translSubvec\lowfreqimg}
\newcommand{\translInpmultimg}{\transl_{\translvecDiscrete}\inpmultimg}

\newcommand{\inpimgAstComplexWeightimg}{\inpimg \ast \flippedComplexWeightimg}
\newcommand{\inpimgAstComplexWeightimgSub}{(\inpimgAstComplexWeightimg) \downarrow \sub}

\newcommand{\translInpimgAstComplexWeightimg}{\translInpimg \ast \flippedComplexWeightimg}

\newcommand{\translInpimgAstComplexWeightimgSub}{(\translInpimgAstComplexWeightimg) \downarrow \sub}

\newcommand{\stochInpfun}{\SFF}
\newcommand{\stochMagnitude}{\SFM}
\newcommand{\stochComplexPhase}{\SFZ}
\newcommand{\stochCosfun}{\SFG}
\newcommand{\stochPhase}{\SFH}
\newcommand{\stochDiffRmaxCmod}{\SFP}
\newcommand{\stochOneminusCosmax}{\SFQ}
\newcommand{\stochDiffShiftRmax}{\SFR}

\newcommand{\normtwoStochMagnitude}{\avgseq{\SFS}}
\newcommand{\normtwoStochDeltaOutimgCmod}{\avgseq{\SFDelta}}

\newcommand{\avgStochDiffRmaxCmod}{\avgseq{\stochDiffRmaxCmod}}
\newcommand{\avgStochOneminusCosmax}{\avgseq{\stochOneminusCosmax}}
\newcommand{\avgStochDiffShiftRmax}{\avgseq{\stochDiffShiftRmax}}

\newcommand{\randMagnitudeMulteval}{\boldsymbol{\stochMagnitude}}

\newcommand{\stochInpfunInterp}{\stochInpfun_\inpimg}
\newcommand{\stochLowfreqfunInterp}{\stochInpfun_{0,\, \inpimg}}
\newcommand{\stochHighfreqfunInterp}{\stochInpfun_{1,\, \inpimg}}
\newcommand{\stochMagnitudeInterp}{\stochMagnitude_\inpimg}
\newcommand{\stochPhaseInterp}{\stochPhase_\inpimg}
\newcommand{\stochComplexPhaseInterp}{\stochComplexPhase_\inpimg}
\newcommand{\stochOneminusCosmaxInterp}{\stochOneminusCosmax_\inpimg}
\newcommand{\stochCosfunInterpEvalpoint}{\stochCosfun_{\inpimg,\, \vectorindexbis}}
\newcommand{\stochCosmaxInterp}{\stochCosfun^{\idxmax}_\inpimg}
\newcommand{\stochMagnitudeTranslInterp}{\stochMagnitude_{\translInpimg}}
\newcommand{\stochComplexPhaseTranslInterp}{\stochComplexPhase_{\translInpimg}}

\newcommand{\stochInpfunAstWaveletInterp}{\stochInpfunInterp \ast \flippedWaveletInterp}

\newcommand{\evalEnergyOutimgCmod}{\sigma}
\newcommand{\evalDiff}{\rho}

\newcommand{\evalDiffRmaxCmodNESelect}{\evalDiff^{\topright}_{\selectExample\selectOutchannel}}

\newcommand{\evalDiffRmaxCmodNESelectSquared}{\evalDiff^{\topright 2}_{\selectExample\selectOutchannel}}

\newcommand{\evalDiffShiftRmaxNESelect}{\evalDiff^{\idxmax\topright}_{\selectExample\selectOutchannel}}

\newcommand{\evalDiffShiftCModNESelect}{\evalDiff^{\idxmod\topright}_{\selectExample\selectOutchannel}}

\newcommand{\avgEvalDiff}{\avgseq{\evalDiff}}

\newcommand{\avgEvalDiffRmaxCmodNESelectSquared}{\avgEvalDiff^{\topright 2}_{\selectOutchannel}}
\newcommand{\avgEvalDiffRmaxCmodSESelectSquared}{\avgEvalDiff^{\bottomright 2}_{\selectOutchannel}}
\newcommand{\avgEvalDiffRmaxCmodSWSelectSquared}{\avgEvalDiff^{\bottomleft 2}_{\selectOutchannel}}
\newcommand{\avgEvalDiffRmaxCmodNWSelectSquared}{\avgEvalDiff^{\topleft 2}_{\selectOutchannel}}

\newcommand{\avgEvalDiffShiftRmaxNESelect}{\avgEvalDiff^{\idxmax\topright}_{\selectOutchannel}}
\newcommand{\avgEvalDiffShiftCmodNESelect}{\avgEvalDiff^{\idxmod\topright}_{\selectOutchannel}}

\newcommand{\avgStochDiffRmaxCmodInpimg}{\avgStochDiffRmaxCmod_{\inpimg}}
\newcommand{\avgStochDiffRmaxCmodInpimgbis}{\avgStochDiffRmaxCmod_{\inpimg'}}
\newcommand{\avgStochDiffRmaxCmodTranslInpimg}{\avgStochDiffRmaxCmod_{\translInpimg}}
\newcommand{\avgStochDiffMultRmaxCmod}{\avgStochDiffRmaxCmod_{\inpmultimg,\, \selectOutchannel}}
\newcommand{\avgStochOneminusCosmaxInpimg}{\avgStochOneminusCosmax_{\inpimg}}
\newcommand{\avgStochDiffShiftRmaxInpimg}{\avgStochDiffShiftRmax_{\inpimg,\, \translvecDiscrete}}
\newcommand{\avgStochDiffShiftMultRmaxInpimg}{\avgStochDiffShiftRmax_{\inpmultimg,\, \translvecDiscrete,\, \selectOutchannel}}
\newcommand{\avgStochDiffShiftMultRmax}{\avgStochDiffShiftRmax_{\inpmultimg,\, \translvecDiscrete,\, \selectOutchannel}}

\newcommand{\normtwoStochMagnitudeInpimg}{\normtwoStochMagnitude_{\inpimg}}
\newcommand{\normtwoStochMagnitudeInpimgBounded}{\normtwoStochMagnitude_{\inpimg,\, \boundVectorindex}}
\newcommand{\normtwoStochDeltaOutimgCmodInpimg}{\normtwoStochDeltaOutimgCmod_{\inpimg}}

\newcommand{\ExpvalDeltaOutimgCmodCondMagnitude}{\Bigcondexpval{\normtwoStochDeltaOutimgCmodInpimg^2}{\normtwoStochMagnitudeInpimg^2 = \evalEnergyOutimgCmod}}
\newcommand{\ExpvalDiffRmaxCmod}{\Expval\bigl[\avgStochDiffRmaxCmodInpimg^2\bigr]}
\newcommand{\ExpvalDiffMultRmaxCmod}{\Expval\bigl[\avgStochDiffMultRmaxCmod^2\bigr]}
\newcommand{\ExpvalOneminusCosmax}{\Expval\bigl[\avgStochOneminusCosmaxInpimg^2\bigr]}
\newcommand{\ExpvalDiffRmaxCmodCondMagnitude}{\Bigcondexpval{\avgStochOneminusCosmaxInpimg^2}{\normtwoStochMagnitudeInpimg^2 = \evalEnergyOutimgCmod}}
\newcommand{\ExpvalDiffShiftRmax}{\Expval\bigl[\avgStochDiffShiftRmaxInpimg\bigr]}
\newcommand{\ExpvalDiffShiftMultRmax}{\Expval\bigl[\avgStochDiffShiftMultRmaxInpimg\bigr]}

\newcommand{\ExpvalCosmaxInterp}[1]{\Expval\left[\stochCosmaxInterp(\spatialvec)^{#1}\right]}

\newcommand{\evalShiftinvCmod}{{\shiftinvCmod(\supportsizeDiscrete\translvecDiscrete)}}
\newcommand{\evalLowerboundDiffRmaxCmod}{\lowerboundDiffRmaxCmod_\gridhalfsize(\sub\supportsizeDiscrete)}
\newcommand{\evalProxyfunRmaxCmod}{\proxyfunRmaxCmod_\gridhalfsize(\sub \freqvecMpipi)}
\newcommand{\evalProxyfunRmaxCmodSelect}{\proxyfunRmaxCmod_\gridhalfsize(\sub \freqvecMpipiSelect)}
\newcommand{\evalCosfun}{\cosfun(\spatialvec, \searchvec)}
\newcommand{\evalCosfunInpimg}{\cosfunInpimg\bigl(\spatialvec,\, \searchvecSelect\bigr)}
\newcommand{\evalCosfunInpimgGrid}{\cosfunInpimg\bigl(\spatialvecSelect,\, \searchvecSelect\bigr)}

\newcommand{\expiFreqvecTranslvec}{\rme^{i\innerprod{\freqvec}{\translvecContinuous}}}

\newcommand{\complexLtwo}{L^2_{\mathC}}
\newcommand{\realLtwo}{L^2_{\mathR}}
\newcommand{\complexltwo}{l^2_{\mathC}}
\newcommand{\realltwo}{l^2_{\mathR}}

\newcommand{\complexLtwoRsq}{\complexLtwo(\mathR^2)}

\newcommand{\complexLtwoMpipisq}{\complexLtwo(\mpipi^2)}
\newcommand{\realLtwoRsq}{\realLtwo(\mathR^2)}
\newcommand{\realLtwoR}{\realLtwo(\mathR)}

\newcommand{\realltwoZ}{\realltwo(\mathZ)}
\newcommand{\complexltwoZsq}{\complexltwo(\mathZ^2)}
\newcommand{\realltwoZsq}{\realltwo(\mathZ^2)}

\newcommand{\complexltwoZsqpower}[1]{\bigl(\complexltwoZsq\bigr)^{#1}}
\newcommand{\realltwoZsqpower}[1]{\bigl(\realltwoZsq\bigr)^{#1}}

\newcommand{\Gaborfilter}[2]{\subspace\bigl(#1,\, #2\bigr)}
\newcommand{\DiscreteGaborfilter}[2]{\discreteSubspace\bigl(#1,\, #2\bigr)}
\newcommand{\GaborfilterGen}{\Gaborfilter{\freqvec}{\supportsizeContinuous}}
\newcommand{\DiscreteGaborfilterGen}{\DiscreteGaborfilter{\freqvecMpipi}{\supportsizeDiscrete}}

\newcommand{\GaborfilterDt}{\Gaborfilter{\freqvecMpipiMultires{\depth}_\selectOutchannel}{\supportsizeDiscrete_\depth}}
\newcommand{\DiscreteGaborfilterDt}{\DiscreteGaborfilter{\freqvecMpipiMultires{\depth}_\selectOutchannel}{\supportsizeDiscrete_\depth}}

\newcommand{\arc}[2]{\left[#1,\,#2\right]_{\unitcircle}}
\newcommand{\bigarc}[2]{\bigl[#1,\,#2\bigr]_{\unitcircle}}

\newcommand{\complexPhaseshift}{\complexPhase_{\vectorindexbis}}

\newcommand{\funcgridhalfsize}[2]{\iter{#1_{#2}}{\gridhalfsize}}
\newcommand{\orderedComplexPhaseshift}[1]{\funcgridhalfsize{\complexPhase}{#1}}
\newcommand{\orderedComplexPhaseshiftSelect}{\orderedComplexPhaseshift{\selectEvalpoint}}
\newcommand{\orderedComplexPhaseshiftSelectNext}{\orderedComplexPhaseshift{\selectEvalpoint + 1}}
\newcommand{\orderedComplexPhaseshiftSelectOther}{\orderedComplexPhaseshift{\selectEvalpoint'}}
\newcommand{\halfOrderedComplexPhaseshift}[1]{\funcgridhalfsize{\overline{\complexPhase}}{#1}}
\newcommand{\centeredHalfOrderedComplexPhaseshift}[1]{\funcgridhalfsize{{}\widetilde{\complexPhase}}{#1}}
\newcommand{\halfOrderedComplexPhaseshiftSelect}{\halfOrderedComplexPhaseshift{\selectEvalpoint}}
\newcommand{\centeredHalfOrderedComplexPhaseshiftSelect}{\centeredHalfOrderedComplexPhaseshift{\selectEvalpoint}}
\newcommand{\orderedPhase}[1]{\funcgridhalfsize{\phase}{#1}}
\newcommand{\orderedPhaseSelect}{\orderedPhase{\selectEvalpoint}}
\newcommand{\halfOrderedPhase}[1]{\funcgridhalfsize{\overline\phase}{#1}}
\newcommand{\diffOrderedPhase}[1]{\funcgridhalfsize{\diffPhase}{#1}}
\newcommand{\diffOrderedPhaseSelect}{\diffOrderedPhase{\selectEvalpoint}}
\newcommand{\evalbisDiffOrderedPhaseSelect}{\diffOrderedPhaseSelect\!(\freqvecMpipibis)}
\newcommand{\orderedArc}[1]{\funcgridhalfsize{\measurablesetUnitcircle}{#1}}
\newcommand{\halfOrderedArc}[1]{\funcgridhalfsize{\overline\measurablesetUnitcircle}{#1}}

\newcommand{\evalComplexPhaseshift}{\complexPhaseshift(\sub\freqvecMpipi)}
\newcommand{\evalbisComplexPhaseshift}{\complexPhaseshift(\freqvecMpipibis)}
\newcommand{\evalOrderedComplexPhaseshift}{\orderedComplexPhaseshiftSelect\!(\sub\freqvecMpipi)}

\newcommand{\evalbisOrderedComplexPhaseshift}{\orderedComplexPhaseshiftSelect\!(\freqvecMpipibis)}

\newcommand{\setofEvalbisComplexPhaseshift}{\{\evalbisComplexPhaseshift\}_{\vectorindexInPoolinggrid}}
\newcommand{\setofOrderedComplexPhaseshift}{\bigl(
    \orderedComplexPhaseshiftSelect
\bigr)_{\pointInSetofEvalpointsGrid}}
\newcommand{\setofEvalOrderedComplexPhaseshift}{\bigl(
    \evalOrderedComplexPhaseshift
\bigr)_{\pointInSetofEvalpointsGrid}}
\newcommand{\setofEvalbisOrderedComplexPhaseshift}{\bigl(
    \evalbisOrderedComplexPhaseshift
\bigr)_{\pointInSetofEvalpointsGrid}}

\newcommand{\arcPointToPoint}{\arc{\pointunitcircle}{\pointunitcircle'}}
\newcommand{\arcSelect}{\orderedArc{\selectEvalpoint}}
\newcommand{\halfarcSelect}{\halfOrderedArc{\selectEvalpoint}}

\newcommand{\evalbisArcSelect}{\arcSelect\!(\freqvecMpipibis)}

\newcommand{\angleArc}{\angle(\pointunitcircle^\ast \pointunitcircle')}

\title{On the Shift Invariance of Max Pooling Feature Maps in Convolutional Neural Networks}

\author{\name Hubert Leterme \email hubert.leterme@ensicaen.fr \\
      \addr Université Caen Normandie, ENSICAEN, CNRS, Normandie Univ, GREYC UMR 6072, F-14000 Caen, France \\
      Univ. Grenoble Alpes, CNRS, Inria, Grenoble INP, LJK, 38000 Grenoble, France
      \AND
      \name Kévin Polisano \email kevin.polisano@univ-grenoble-alpes.fr \\
      \addr Univ. Grenoble Alpes, CNRS, Grenoble INP, LJK, 38000 Grenoble, France\\
      \AND
      \name Valérie Perrier \email valerie.perrier@univ-grenoble-alpes.fr \\
      \addr Univ. Grenoble Alpes, CNRS, Grenoble INP, LJK, 38000 Grenoble, France
      \AND
      \name Karteek Alahari \email karteek.alahari@inria.fr \\ Univ. Grenoble Alpes, CNRS, Inria, Grenoble INP, LJK, 38000 Grenoble, France
      \addr}

\def\month{MM}  
\def\year{YYYY} 
\def\openreview{\url{https://openreview.net/forum?id=XXXX}} 

\begin{document}

\maketitle

\begin{abstract}
  	This paper focuses on improving the mathematical interpretability of convolutional neural networks (CNNs) in the context of image classification.
	Specifically, we tackle the instability issue arising in their first layer, which tends to learn parameters that closely resemble oriented band-pass filters when trained on datasets like ImageNet. Subsampled convolutions with such Gabor-like filters are prone to aliasing, causing sensitivity to small input shifts.
	In this context, we establish conditions under which the max pooling operator approximates a complex modulus, which is nearly shift invariant. We then derive a measure of shift invariance for subsampled convolutions followed by max pooling. In particular, we highlight the crucial role played by the filter's frequency and orientation in achieving stability. We experimentally validate our theory by considering a deterministic feature extractor based on the dual-tree complex wavelet packet transform, a particular case of discrete Gabor-like decomposition.
\end{abstract}

\section{Introduction}
\label{sec:intro}

Understanding the mathematical properties of deep convolutional neural networks (CNNs) \citep{LeCun2015} remains a challenging issue today. On the other hand, wavelet and multi-resolution analysis are built upon a well-established mathematical framework. They have proven to be efficient for tasks such as signal compression and denoising \citep{Vetterli2001}, and have been widely used as feature extractors for signal, image and texture classification \citep{Laine1993, Pittner1999, Yen2000, Huang2008}.
There is a broad literature revealing strong connections between these two paradigms, as discussed in \cref{subsec:intro_motivation,subsec:intro_relatedwork}.
Inspired by this line of research, the present paper extends existing knowledge about CNN properties. Specifically, we assess the shift invariance of max pooling feature maps through both theoretical and empirical approaches in the context of image classification, by leveraging the properties of oriented band-pass filters.

\subsection{Motivations and Main Contributions}
\label{subsec:intro_motivation}

CNNs process input images through convolutions and nonlinear pooling operations, transforming them into high-level feature vectors that are subsequently used for the task at hand. In image classification, the feature vectors are fed into a linear classifier. To achieve high classification accuracy, a convolutional network must preserve discriminative image features while reducing intra-class variability \citep{LeCun1998,Bruna2013}. An important and often desired property of CNNs is their ability to remain invariant to small input transformations, such as translations, rotations, distortions, or scaling \citep{Liao2010,Bruna2013,Sifre2013,Bietti2017,Wiatowski2018,Cahill2024}.

In particular, an image in which the main subject is slightly shifted from its original position should retain its initial classification. The absence of this property, known as \emph{translation invariance} or \emph{shift invariance}, can negatively impact the model’s predictive performance. Consequently, a reliable model must learn this property through extensive training on large datasets. The main focus of this paper is therefore to assess whether shift invariance is inherently guaranteed by the model's architecture. Since perfect invariance is rarely achieved, we also the term \emph{stability} to refer to this behavior.

More specifically, we focus on commonly observed phenomenon in CNNs when trained on image datasets: many convolution kernels in the first layer resemble band-pass oriented waveforms \citep{Yosinski2014,Rai2020}, referred to as \emph{Gabor-like filters}.
Whether the features extracted by these filters remain stable under translations has been partly addressed by \citet{Azulay2019, Zhang2019}, who highlight that strided convolution and pooling operations can significantly diverge from shift invariance, due to aliasing effects when subsampling high-frequency signals. In response, \citet{Zhang2019, Zou2023} introduced antialiasing methods based on low-pass filtering, improving both stability and predictive performance---albeit at the cost of some loss of information.

In the current paper, we show that, under specific conditions that we establish, the max pooling operator can partially restore shift invariance.
We unveil a connection between the output of the first max pooling layer and the modulus of complex Gabor-like coefficients, which is known to be nearly shift invariant. This work offers a promising direction for improving shift invariance in CNNs while preserving high-frequency information---unlike the previously-mentioned approaches.

Before proceeding further, we emphasize that our study is not limited to purely convolutional architectures. In recent years, self-attention mechanisms have gained significant interest in computer vision due to their ability to model complex, long-range dependencies in image representations. Notably, the \emph{vision transformer} (ViT) was introduced by \citet{Dosovitskiy2021} by adapting the transformer architecture, initially developed for natural language processing (NLP) \citep{Vaswani2017}, to computer vision tasks. Unlike CNNs, the ViT operates without any convolutional layers. Instead, input images are partitioned into fixed-size patches, which serve as inputs to the first self-attention module. However, more recent research has explored hybrid architectures that integrate self-attention with convolutional components \citep{Wu2021,Yuan2021,Hassani2022,Li2023a,Yin2024}. This approach allows for reducing the amount of labeled data required, while achieving faster training and improving generalizability. In particular, the first layers of a CNN can be used as a ``convolutional token embedding,'' replacing the naive patch extraction used in the original ViT. The theoretical framework presented in this paper also applies to such hybrid architectures, providing a better understanding of the invariance properties of the inputs to self-attention modules.

\subsection{Related Work}
\label{subsec:intro_relatedwork}

Analyzing the invariance properties of CNNs is critical as it enables to identify their shortcomings and provides an opportunity to enhance their performance.
In recent years, several works focused on this topic.

\subsubsection{Wavelet Scattering Networks}

Most notably, \citet{Bruna2013} developed a family CNN-like architectures, named \emph{wavelet scattering networks} (ScatterNets), based on a succession of complex convolutions with wavelet filters followed by nonlinear modulus pooling. They produce translation-invariant image representations which are stable to deformation and preserve high-frequency information \citep{Mallat2012,Mallat2016,Czaja2024}. A variation has been proposed by \citet{Sifre2013} to include rotational invariance.
ScatterNets achieve strong performance on handwritten digits and texture datasets, but do not scale well to more complex ones. To overcome this, \citet{Oyallon2017, Oyallon2018} introduced hybrid ScatterNets, where the scattering coefficients are fed into a standard CNN architecture, showing that the network complexity can be reduced while keeping competitive performance.
Derived models include ScatterNets built upon the dual-tree complex wavelet transform \citep{Singh2017}, learnable and parametric ScatterNets \citep{Cotter2019,Gauthier2022}, geometric ScatterNets operating on Riemanian manifolds \citep{Perlmutter2020}, and graph ScatterNets \citep{Gama2019,Zou2020a}.
Also worth mentioning, \citet{Czaja2019, Czaja2020} studied ScatterNets based on uniform covering frames, i.e., frames splitting the frequency domain into windows of roughly equal size, much like \dtcwpt frames (as used in the present paper).
Other works by \citet{Zarka2020,Zarka2021} proposed to sparsify wavelet scattering coefficients by learning a dictionary matrix, to learn $\obo$ convolutions between feature maps of scattering coefficients and to apply soft thresholding to reduce within-class variability.

ScatterNets are specifically designed to meet some desired properties. As deep learning architectures with well-established mathematical properties, they are sometimes used as explanatory models for standard, freely-trained networks. However, whether their properties are transferable to a broader class of models is unclear, because the former rely on complex-valued convolutions whereas more conventional architectures exclusively employ real-valued kernels. Moreover, the modulus operator is used as an activation and pooling layer in ScatterNets, whereas standard CNNs implement pointwise nonlinear operators such as ReLU and spatial pooling layers such as max pooling. This limitation has been pointed out by \citet{Tygert2016} as an argument in favor of complex-valued CNNs.
In this context, our work seeks evidence that properties established for complex-valued networks are---to some extent---embedded in standard architectures.

\subsubsection{Invariance Studies in CNNs}

\Citet{Wiatowski2018} considered a wide variety of feature extractors involving convolutions, Lipschitz-continuous non-linearities and pooling operators. The paper shows that outputs become more translation invariant with increasing network depth.
Additionally, \citet{Cahill2024} designed a family of operators called \emph{max filters}, which encompass a wide variety of operators including, in specific cases, the max pooling operator. Stability with respect to diffeomorphisms were established, following the ideas developed for scattering networks.
However, these results do not fully extend to the discrete framework, because subsampled convolutions with band-pass real-valued filters can introduce aliasing artifacts, resulting in instability to translations \citep{Azulay2019,Zhang2019}. The current paper specifically addresses this issue.

Another line of work is focused on modeling and studying CNNs from the point of view of convolutional kernel networks \citep{Bietti2019a,Bietti2019b,Scetbon2020,Bietti2022,Chen2023}. These authors showed that certain classes of CNNs are contained in the reproducing kernel Hilbert space (RKHS) of a multilayer convolutional kernel representation. As such, stability metrics are estimated, based on the RKHS norm which is difficult to control in practice. Kernel representations do not seem to suffer from aliasing effects; this can be explained by the Gaussian pooling layers that have been employed instead of max pooling: by discarding high-frequency information, shift invariance is preserved.

Finally, some papers studied stability of CNNs in a broader sense, measured in terms of Lipschitz continuity \citep{Szegedy2014,Balan2018,Virmaux2018,Perez2020,Zou2020,Gupta2022,Zuhlke2024}. However, the Lipschitz bounds, which have been obtained theoretically, are generally several orders of magnitude higher than empirical results.
This discrepancy may be due to the fact that these bounds were obtained for generic situations and represent overly conservative worst-case scenarios, rather than typical real-world situations.
Furthermore, the specific case of convolutions with band-pass Gabor-like filters have been overlooked, except for \citet{Perez2020}.

In summary, we have identified the following blind spots in the literature, regarding the topic of studying shift invariance in CNNs.
\begin{itemize}
   \item The effect of the max pooling operator on network stability under small input shifts has not been investigated, particularly when used in combination with Gabor-like convolutions.
   \item While the shift invariance of CNNs tends to increase with network depth in the continuous framework, in the discrete case, the presence of subsampled convolutions with oriented band-pass filters can lead to aliasing artifacts. To our knowledge, the literature lacks theoretical studies that take these aliasing effects into account.
   \item Although extensive studies have been conducted on complex-valued convolutions followed by modulus, a link is missing to extend these results to standard CNNs, which implement real-valued convolutions and spatial pooling operators.
\end{itemize}
All these points have been tackled in the present paper, from both theoretical and empirical perspectives.

\subsection{Paper Outline}
\label{subsec:intro_contributions}

In what follows, $\realltwoZsq$ and $\complexltwoZsq$ represent the discrete spaces of square-summable two-dimensional sequences with values in $\mathR$ and $\mathC$, respectively.
Let $\complexWeightimg \in \complexltwoZsq$ denote a two-dimensional band-pass, oriented and analytic \emph{Gabor-like} filter, for which a formal definition will be provided in \eqref{eq:gaborfilt_discrete}. We first consider an operator, referred to as \emph{real-max-pooling} (\rmax), which computes the subsampled cross-correlation between an input image $\inpimg \in \realltwoZsq$ and the real part of $\complexWeightimg$; then calculates the maximum value over a sliding discrete grid:
\begin{equation}
	\rmaxopSubGridWeight: \inpimg \mapsto \maxpool_\gridhalfsize\left(
		\bigl(
			\inpimg \ast \flipped{\Real\complexWeightimg}
		\bigr) \downarrow \sub
	\right),
\label{eq:rmax}
\end{equation}
where $\sub \in \nonzeroMathN$ denotes a subsampling factor, $\flipped\weightimg$ denotes the ``flipped'' sequence for any given $\weightimg \in \realltwoZsq$ or $\complexltwoZsq$, satisfying, for any $\vectorindexInZtwo$,
\begin{equation}
	\flipped\weightimg[\vectorindex] := \weightimg[-\vectorindex],
\label{eq:flippedseq}
\end{equation}
and $\ast$, $\downarrow$ respectively refer to the convolution and subsampling operations, defined by 
\begin{equation}
  (\inpimg \ast \flipped\weightimg)[\vectorindex] := \sumoverVectorindicesbis \inpimg[\vectorindexbis] \, \flipped\weightimg[\vectorindex - \vectorindexbis]
  \qqand
  (\outimg \downarrow \sub)[\vectorindex] := \outimg[\sub\vectorindex].
\label{eq:conv2d}
\end{equation}
In the above expression, $\maxpool_\gridhalfsize$ selects the maximum value over a sliding grid of size $(2\gridhalfsize+1) \times (2\gridhalfsize+1)$, with a subsampling factor of $2$. More formally, for any $\outimg \in \realltwoZsq$ and any $\vectorindexInZtwo$,
\begin{equation}
	\maxpool_\gridhalfsize(\outimg)[\vectorindex] := \maxVectorindexInPoolinggrid \outimg[2\vectorindex + \vectorindexbis].
\label{eq:maxpool}
\end{equation}
On the other hand, we consider an operator, referred to as \emph{complex-modulus} (\cmod), computing the modulus of subsampled cross-correlation between $\inpimg$ and $\complexWeightimg$:
\begin{equation}
	\cmodopSubWeight: \inpimg \mapsto \left|
		(\inpimg \ast \flipped\complexWeightimg) \downarrow (2\sub)
	\right|.
\label{eq:cmod}
\end{equation}

First, we show that, under the Gabor hypothesis, \cmod is stable with respect to small input shifts (\cref{sec:invariance_cmod}). We then establish conditions on the filter's frequency and orientation under which \cmod and \rmax produce comparable outputs (\cref{sec:cmod_rmax}):
\begin{equation}
	\cmodopSubWeight(\inpimg) \approx \rmaxopSubGridWeight(\inpimg).
\end{equation}
We deduce a measure of shift invariance for \rmax operators, which benefits from the stability of \cmod (\cref{sec:invariance_rmax}).
Next, we extend our results to multichannel operators (\ie, applied on RGB input images), such as implemented in conventional CNN architectures (\cref{sec:multichannel}). Our framework therefore provides a theoretical grounding to study these networks.

\begin{remark}
	In the above definitions, cross-correlations are computed with a subsampling factor which is twice larger for \cmod, compared to \rmax. However, since max pooling is also computed with subsampling, both operators have the same subsampling factor of $2\sub$.
\end{remark}

Finally, in \cref{sec:dtcwpt}, we assess our theoretical findings on a deterministic setting based on the dual-tree complex wavelet packet transform (\dtcwpt), a particular case of discrete Gabor-like decomposition with perfect reconstruction properties \citep{Bayram2008}. \dtcwpt spawns a set of convolution kernels which tile the Fourier domain into square regions of identical size. Such kernels possess characteristics that are comparable to those found in the first convolution layer of CNNs after training with image datasets such as ImageNet \citep{Russakovsky2015}. More specifically, given an input image, we compute the mean square error between the outputs of \cmod and \rmax, for each wavelet packet filter. We then observe that shift invariance, when measured on \rmax feature maps, is nearly achieved when they remain close to \cmod outputs. We therefore establish a domain of validity for shift invariance of the \rmax operator.

This work builds upon an idea sketched by \citet[pp.~190--191]{Waldspurger2015}, which suggests a potential connection between the combinations ``real wavelet transform $\to$ max pooling'' on the one hand and ``complex wavelet transform $\to$ modulus'' on the other hand. Building on this idea, we investigated whether the invariance properties of complex moduli could be captured by the max pooling operator. However, as shown in this paper, these principles do not fully extend to the discrete framework. To address this limitation, we adopted a Bayesian point of view.

\section{Shift Invariance of \cmod Outputs}
\label{sec:invariance_cmod}

The primary goal of this paper is to theoretically establish conditions for near-shift invariance at the output of the first max pooling layer. In this section, we start by proving shift invariance of \cmod operators. Then, in \cref{sec:cmod_rmax}, we establish conditions under which \rmax and \cmod produce closely related outputs. Finally, in \cref{sec:invariance_rmax}, we derive a probabilistic measure of shift invariance for \rmax.

\subsection{Notations}
\label{subsec:invariance_cmod_notations}

The complex conjugate of any number $\pointunitcircle \in \mathC$ is denoted by $\pointunitcircle^\ast$. For any $\normdegree \in \nonzeroMathRplus \cup \{\infty\}$, $\spatialvec \in \mathR^2$ and $\spatialWindowsize \in \mathR_+$, we denote by $\ball{\normdegree}(\spatialvec,\, \spatialWindowsize) \subset \mathR^2$ the closed $l^\normdegree$-ball with center $\spatialvec$ and radius $\spatialWindowsize$. When $\spatialvec = \bzero$, we write $\ball{\normdegree}(\spatialWindowsize)$.

\paragraph{Continuous Framework}

Considering a measurable subset $\probaspace$ of $\mathR^2$, we denote by $\complexLtwo(\probaspace)$ the Hilbert space of square-integrable functions $\inpfun: \probaspace \to \mathC$.
Whenever we talk about equality in $L^\normdegree_\mathC(\probaspace)$ or inclusion in $\probaspace$, it shall be understood as ``almost everywhere with respect to the Lebesgue measure.'' Additionally, we denote by $\realLtwo(\probaspace) \subset \complexLtwo(\probaspace)$ the subspace of real-valued functions. For any $\inpfun \in \complexLtwoRsq$, $\flipped{\inpfun}$ denotes its flipped version: $\flipped{\inpfun}(\spatialvec) := \inpfun(-\spatialvec)$.

The 2D Fourier transform of any $\inpfun \in \complexLtwoRsq$ is denoted by $\fourierInpfun \in \complexLtwoRsq$, such that
\begin{equation}
	\forall \freqvec \in \mathR^2,\, \fourierInpfun(\freqvec) := \iint_{\mathR^2} \inpfun(\spatialvec) \rme^{-i\innerprod{\freqvec}{\spatialvec}} \,\rmd^2\spatialvec.
\end{equation}
For any $\supportsizeContinuous > 0$ and $\freqvec \in \mathR^2$, we denote by $\GaborfilterGen \subset \complexLtwoRsq$ the set of functions whose Fourier transform is supported in a square region of size $\supportsizeContinuous \times \supportsizeContinuous$ centered in $\freqvec$:
\begin{equation}
	\GaborfilterGen := \set{\wavelet \in \complexLtwoRsq}{\supp\fourierWavelet \subset \linftyBall(\freqvec,\, \supportsizeContinuous/2)}.
\label{eq:gaborfilt_continuous}
\end{equation}
$\freqvec$ and $\supportsizeContinuous$ are respectively referred to as \emph{characteristic frequency} and \emph{bandwidth}.
Finally, for any $\translvecContinuous \in \mathR^2$, we consider the translation operator, denoted by $\transl_{\translvecContinuous}$, defined by
\begin{equation}
	\transl_{\translvecContinuous} \inpfun: \spatialvec \mapsto \inpfun(\spatialvec - \translvecContinuous).
\end{equation}

\paragraph{Discrete Framework}

We denote by $\complexltwoZsq$ the space of 2D complex-valued square\hyp{summable} sequences, represented by straight capital letters.
Indexing is made between square brackets: $\forall \inpimg \in \complexltwoZsq,\, \forall \vectorindexInZtwo,\, \inpimg[\vectorindex] \in \mathC$, and we denote by $\realltwoZsq \subset \complexltwoZsq$ the subset of real-valued sequences. For any $\weightimg \in \complexltwoZsq$, $\flippedWeightimg$ denotes its ``flipped'' version as defined in \eqref{eq:flippedseq}. The convolution and subsampling operators, respectively denoted by $\ast$ and $\downarrow$, are defined in \eqref{eq:conv2d}.
2D images, feature maps and convolution kernels are considered as elements of $\complexltwoZsq$. Additionally, multichannel arrays of 2D sequences are denoted by bold straight capital letters, for instance: $\inpmultimg := \left(\inpimg_\selectInchannel\right)_{\selectInchannel \in \setof{\inchannels}}$. Note that indexing starts at $0$ to comply with practical implementations.

The 2D discrete-time Fourier transform of any $\inpimg \in \allowbreak \complexltwoZsq$, denoted by $\fouriertransf{\inpimg} \in \complexLtwoMpipisq$, is defined by
\begin{equation}
	\forall \freqvecMpipi \in \interval{-\pi}{\pi}^2,\, \fouriertransf{\inpimg}(\freqvecMpipi) := \sumoverVectorindices \inpimg[\vectorindex] \rme^{-i\innerprod{\freqvecMpipi}{\vectorindex}}.
\end{equation}
For any $\supportsizeDiscrete \in \zeroexcltwopi$ and $\freqvecMpipi \in \linftyBall(\pi)$, we denote by $\DiscreteGaborfilterGen \subset \complexltwoZsq$ the set of 2D sequences whose Fourier transform is supported in a square region of size $\supportsizeDiscrete \times \supportsizeDiscrete$ centered in $\freqvecMpipi$:
\begin{equation}
	\DiscreteGaborfilterGen := \set{\complexWeightimg \in \complexltwoZsq}{\supp\fourierComplexWeightimg \subset \linftyBall(\freqvecMpipi,\, \supportsizeDiscrete/2)}.
\label{eq:gaborfilt_discrete}
\end{equation}
As in the continuous framework, $\freqvecMpipi$ and $\supportsizeDiscrete$ are respectively referred to as \emph{characteristic frequency} and \emph{bandwidth}. The elements of $\DiscreteGaborfilterGen$ are designated as \emph{Gabor-like filters}.

\begin{remark}
	\label{remark:quotientspace}
	The support $\linftyBall(\freqvecMpipi,\, \supportsizeDiscrete/2)$ actually lives in the quotient space $\mpipi^2 / (2\pi\allowbreak\mathZ^2)$. Consequently, when $\freqvecMpipi$ is close to an edge, a fraction of this region is located at the far end of the frequency domain. From now on, the choice of $\freqvecMpipi$ and $\supportsizeDiscrete$ is implicitly assumed to avoid such a situation.
\end{remark}

\subsection{Intuition}
\label{subsec:invariance_intuition}

\begin{figure}
	\centering
	\includegraphics[width=.32\textwidth]{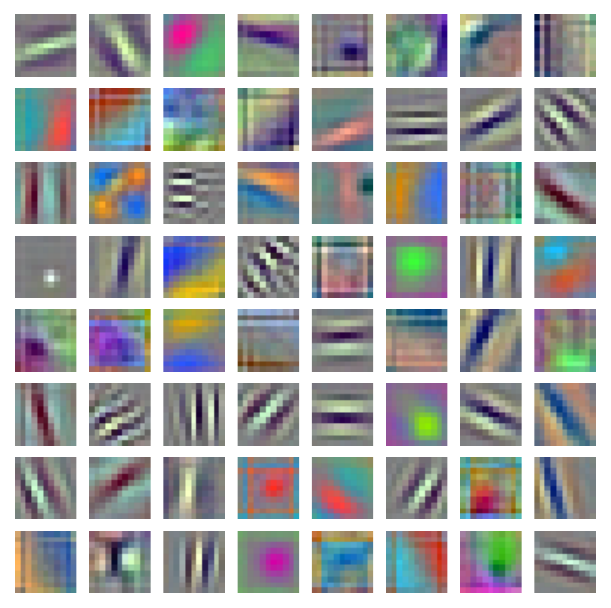}
   \hspace{10pt}
	\includegraphics[width=.32\textwidth]{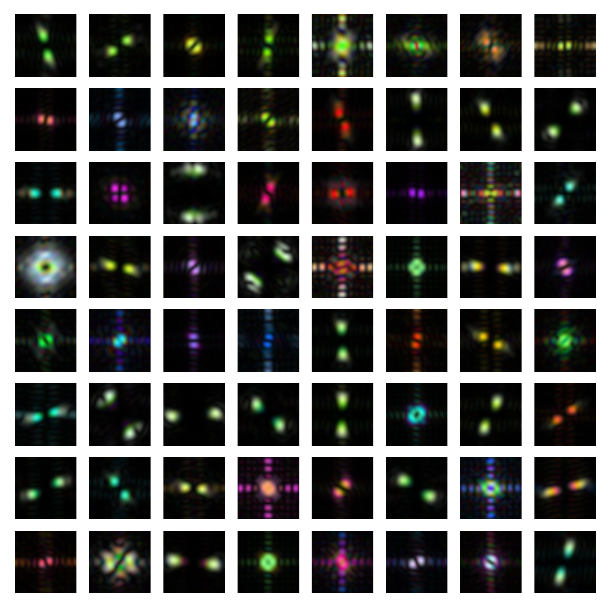}
	\caption[
      AlexNet Convolution Kernels
   ]{
		Spatial (left) and Fourier (right) representations of convolution kernels in the first layer of AlexNet, after training with ImageNet ILSVRC 2012-2017 \citep{Russakovsky2015}. Each kernel connects the $3$ RGB input channels to one of the $64$ output channels.
	}
	\label{fig:alexnetkernels}
\end{figure}

In many CNNs for computer vision, input images are first transformed through subsampled (or strided) convolutions. For instance, in AlexNet, convolution kernels are of size $11 \times 11$ and the subsampling factor is equal to $4$. \Cref{fig:alexnetkernels} displays the corresponding kernels after training with ImageNet. This linear transform is generally followed by rectified linear unit (ReLU) and max pooling.

We can observe that many kernels display oscillating patterns with well-defined orientations (Gabor-like filters). We denote by $\weightimg \in \realltwoZsq$ one of these ``well-behaved'' filters. Its Fourier spectrum roughly consists in two bright spots which are symmetric with respect to the origin.%
\footnote{
	Actually, the Fourier transform of any real-valued sequence is centrally symmetric: $\fourierWeightimg(-\freqvecMpipibis) = {\fourierWeightimg(\freqvecMpipibis)}^\ast$. The specificity of well-oriented filters lies in the concentration of their power spectrum around two precise locations.
} Next, we consider a complex-valued companion $\complexWeightimg \in \complexltwoZsq$ such that
\begin{equation}
	\fourierComplexWeightimg(\freqvecMpipibis) := \bigl(
		1 + \sign\!\innerprod{\freqvecMpipibis}{\unitvec}
	\bigr) \cdot \fourierWeightimg(\freqvecMpipibis) \qquad \forall \freqvecMpipibis \in \mpipi^2,
\label{eq:hilberttransform}
\end{equation}
where $\unitvec$ denotes a unit vector orthogonal to the filter's orientation.

We can show that $\weightimg$ is the real part of $\complexWeightimg$, and that $\complexWeightimg = \weightimg + i \hilberttransf(\weightimg)$, where $\hilberttransf$ denotes the two-dimensional Hilbert transform as introduced by \citet{Havlicek1997}.
It satisfies
\begin{equation}
	\fouriertransf{\hilberttransf(\weightimg)}(\freqvecMpipibis) := -i \sign\!\innerprod{\freqvecMpipibis}{\unitvec} \fourierWeightimg(\freqvecMpipibis).
\end{equation}
As a consequence, $\fourierComplexWeightimg$ is equal to $2\fourierWeightimg$ on one half of the Fourier domain, and $0$ on the other half. Therefore, only one bright spot remains in the spectrum.
We refer the reader to \cref{fig:convkernels_example} for visual example of complex-valued Gabor-like filter.
It turns out that such complex filters with high frequency resolution produce stable signal representations, as we will see in \cref{sec:invariance_cmod}. In the subsequent sections, we then wonder whether this property is kept when considering the max pooling of real-valued convolutions.

\begin{figure}
  \centering
  \input{inputs/convkernels_example}
  \vspace{-5pt}
  \caption[
      Example of Gabor-like Filter
  ]{
      (a), (b): Real and imaginary parts of a Gabor-like filter $\complexWeightimg$ as defined in \eqref{eq:hilberttransform}. (c), (d): Magnitude spectra (modulus of the Fourier transform) of $\weightimg$ and $\complexWeightimg$, respectively.
  }
  \vspace{-0pt}
  \label{fig:convkernels_example}
\end{figure}

In what follows, $\complexWeightimg$ will be referred to as a discrete Gabor-like filter, and the coefficients resulting from the convolution with $\complexWeightimg$ will be referred to as discrete Gabor-like coefficients.
The aim of this section is to show that, under the Gabor hypothesis on the convolution kernels $\complexWeightimg \in \complexltwoZsq$, \cmod is nearly shift-invariant. To clarify, we establish that
\begin{equation}
	\cmodopSubWeight(\inpimg) \approx \cmodopSubWeight(\translInpimg),
\end{equation}
for ``small'' translation vectors $\translvecDiscrete \in \mathR^2$, where a formal definition of the translation operator will be defined in \eqref{eq:transl}. This result is hinted by \citet{Kingsbury1998} but not formally proven.

\subsection{Continuous Framework}
\label{subsec:invariance_cmod_continuous}

We introduce several results regarding functions defined on the continuous space $\mathR^2$. Near-shift invariance on discrete 2D sequences will then be derived from these results by taking advantage of sampling theorems. \Cref{lemma:lowfreqfun} below is adapted from \citet[pp.~190--191]{Waldspurger2015}.

\begin{lemma}
	\label{lemma:lowfreqfun}
	Given $\supportsizeContinuous > 0$ and $\freqvec \in \mathR^2$, let $\wavelet \in \GaborfilterGen$ denote a complex-valued filter such as defined in \eqref{eq:gaborfilt_continuous}. Next, for any real-valued function $\inpfun \in \realLtwoRsq$, we consider the complex-valued function $\lowfreqfun \in \complexLtwoRsq$ defined by
	\begin{equation}
		\lowfreqfun: \spatialvec \mapsto (\inpfunAstWavelet)(\spatialvec) \, \rme^{i\innerprod{\freqvec}{\spatialvec}}.
	\label{eq:lowfreqfun}
	\end{equation}
	Then $\lowfreqfun$ is low-frequency. Specifically,
	\begin{equation}
		\supp \fourierLowfreqfun \subset \fourierWindowGaborContinuous.
	\label{eq:suppLowfreqfun}
	\end{equation}
\end{lemma}

\begin{proof}
	See \cref{subsec:appendix_proofs_lemma_lowfreqfun}.
\end{proof}

On the other hand, the following proposition provides a shift invariance bound for low-frequency functions such as introduced above.

\begin{proposition}
	\label{prop:shiftinvariance_cont}
	For any $\lowfreqfun \in \realLtwoRsq$ such that $\supp \fourierLowfreqfun \subset \fourierWindowGaborContinuous$, and any $\translvecContinuous \in \mathR^2$,
	\begin{equation}
		\normLtwo{\transl_{\translvecContinuous} \lowfreqfun - \lowfreqfun} \leq \shiftinvCmod(\supportsizeContinuous\translvecContinuous) \normLtwo{\lowfreqfun},
	\label{eq:shiftinvariance_continuous}
	\end{equation}
	where we have defined
	\begin{equation}
		\shiftinvCmod: \btau \mapsto \frac{\normone{\btau}}{2}.
	\label{eq:shiftinvariance_continuous_multfac}
	\end{equation}
\end{proposition}

\begin{proof}
	See \cref{subsec:appendix_proofs_prop_shiftinvariance_cont}.
\end{proof}

\subsection{Adaptation to Discrete 2D Sequences}
\label{subsec:invariance_cmod_discrete2continuous}

Given $\supportsizeDiscrete \in \zeroexcltwopi$ and $\freqvecMpipi \in \linftyBall(\pi)$, let $\complexWeightimg \in \DiscreteGaborfilterGen$ denote a discrete Gabor-like filter such as defined in \eqref{eq:gaborfilt_discrete}. For any image $\inpimg \in \complexltwoZsq$ with finite support and any subsampling factor $\sub \in \nonzeroMathN$, we express $\inpimgAstComplexWeightimgSub$ using the continuous framework introduced above, and derive an invariance formula.

For any sampling interval $\samplinterv \in \nonzeroMathRplus$, let $\shanScalingfunSamplinterv \in \realLtwoRsq$ denote the Shannon scaling function parameterized by $\samplinterv$, such that
\begin{equation}
	\fouriertransf{\shanScalingfunSamplinterv} := \samplinterv \mathone_{\fourierWindowSampling}.
\label{eq:fourierShanScalingfun}
\end{equation}
This 2D function is a tensor product of scaled and normalized sinc functions. For any $\vectorindexInZtwo$, we denote by $\shanScalingfunSamplintervVectorindex$ a shifted version of $\shanScalingfunSamplinterv$, satisfying
\begin{equation}
	\shanScalingfunSamplintervVectorindex(\spatialvec) := \shanScalingfunSamplinterv(\spatialvec - \samplinterv\vectorindex).
\end{equation}
Then, $\bigl\{
	\shanScalingfunSamplintervVectorindex
\bigr\}_{\vectorindexInZtwo}$ is an orthonormal basis of
\begin{equation}
	\shannonspaceSamplinterv := \bigset{\inpfun \in \complexLtwoRsq}{\supp\fourierInpfun \subset \fourierWindowSampling}.
\label{eq:shannonspace}
\end{equation}
Then, using the notation introduced in \eqref{eq:gaborfilt_continuous}, we have $\shannonspaceSamplinterv = \MV(\bzero, \, 2\pi/\samplinterv)$.

We now consider the following lemma.

\begin{lemma}
	Let $\samplinterv > 0$. For any $\inpfun \in \shannonspaceSamplinterv$ and any $\freqvecbis \in \fourierWindowSampling$, we have
	\begin{equation}
		\fourierInpfun(\freqvecbis) = \samplinterv \, \fourierInpimg(\samplinterv\freqvecbis),
	\label{eq:fouriersampling_1}
	\end{equation}
	where $\inpimg \in \complexltwoZsq$ is a uniform sampling of $\inpfun$, defined such that $\inpimg[\vectorindex] := \samplinterv \, \inpfun(\samplinterv\vectorindex)$, for any $\vectorindex \in \mathZ^2$. Moreover, we have the following norm equality:
	\begin{equation}
		\normLtwo{\inpfun} = \normtwo{\inpimg}.
	\label{eq:fouriersampling_2}
	\end{equation}
	\label{lemma:fouriersampling}
\end{lemma}

\begin{proof}
	See \cref{subsec:appendix_proofs_lemma_fouriersampling}.
\end{proof}

We then get the following proposition, which draws a bond between the discrete and continuous frameworks.

\begin{proposition}
	Let $\inpimg \in \realltwoZsq$ denote an input image with finite support, and $\complexWeightimg \in \DiscreteGaborfilterGen$. Considering a sampling interval $\samplinterv \in \nonzeroMathRplus$, we define $\inpfunInterp$ and $\waveletInterp \in \shannonspaceSamplinterv$ such that
	\begin{equation}
		\inpfunInterp := \sumoverVectorindices \inpimg[\vectorindex] \,\shanScalingfunSamplintervVectorindex \qqand \waveletInterp := \sumoverVectorindices \complexWeightimg[\vectorindex] \,\shanScalingfunSamplintervVectorindex.
	\label{eq:interp}
	\end{equation}
	Then,
	\begin{equation}
		\waveletInterp \in \Gaborfilter{\freqvecMpipi / \samplinterv}{\supportsizeDiscrete / \samplinterv}.
	\label{eq:gaborfilter}
	\end{equation}
	Moreover, for all $\vectorindex \in \mathZ$,
	\begin{equation}
		\inpimg[\vectorindex] = \samplinterv \, \inpfunInterp(\samplinterv\vectorindex);\qquad \complexWeightimg[\vectorindex] = \samplinterv \, \waveletInterp(\samplinterv\vectorindex),
	\label{eq:shannon}
	\end{equation}
	and, for a given subsampling factor $\sub \in \nonzeroMathN$,
	\begin{equation}
		\left(
			\inpimgAstComplexWeightimgSub
		\right)[\vectorindex] = \bigl(
			\inpfunAstWaveletInterp
		\bigr)\left(
			\sub\samplinterv\vectorindex
		\right).
	\label{eq:discrete2continuous}
	\end{equation}
\label{prop:discrete2continuous}
\end{proposition}

\begin{proof}
	See \cref{subsec:appendix_proofs_prop_discrete2continuous}.
\end{proof}

\Cref{prop:discrete2continuous} introduces a latent subspace of $\realLtwoRsq$ from which input images are uniformly sampled. This allows us to define, for any $\translvecDiscrete \in \mathR^2$, a translation operator $\transl_{\translvecDiscrete}$ on discrete sequences, even if $\translvecDiscrete$ has non-integer values:
\begin{equation}
	\translInpimg[\vectorindex] := \samplinterv \, \transl_{\samplinterv\translvecDiscrete} \inpfunInterp(\samplinterv\vectorindex),
\label{eq:transl}
\end{equation}
where $\inpfunInterp$ is defined in \eqref{eq:interp}. We can indeed show that this definition is independent from the choice of sampling interval $\samplinterv > 0$. Moreover, given $\inpimg \in \realltwoZsq$, we have
\begin{gather}
	\forall \vectorindexbisInZtwo,\, \transl_{\vectorindexbis}\inpimg[\vectorindex] = \inpimg[\vectorindex - \vectorindexbis];
\label{eq:integer_transl} \\
	\forall \translvecDiscrete,\, \translvecDiscreteBis \in \mathR^2,\, \transl_{\translvecDiscrete}(\transl_{\translvecDiscreteBis}\inpimg) = \transl_{\translvecDiscrete + \translvecDiscreteBis}\inpimg,
\label{eq:comp_transl}
\end{gather}
which shows that $\transl_{\translvecDiscrete}$ corresponds to the intuitive idea of a translation operator. Expressions \eqref{eq:integer_transl} and \eqref{eq:comp_transl} are direct consequence of the following lemma, which bonds the shift operator in the discrete and continuous frameworks.

\begin{lemma}
	\label{lemma:commut_interp_transl}
	For any $\inpimg \in \realltwoZsq$ and any $\translvecDiscrete \in \mathR^2$,
	\begin{equation}
		\inpfunTranslInterp = \translSamplintervVec\inpfunInterp.
	\label{eq:commut_interp_transl}
	\end{equation}
\end{lemma}

\begin{proof}
	See \cref{subsec:appendix_proofs_lemma_commut_interp_transl}.
\end{proof}

We now consider the following corollary to \cref{prop:discrete2continuous}.

\begin{corollary}
	\label{cor:discrete2continuous}
	For any shift vector $\translvecDiscrete \in \mathR^2$, we have
	\begin{equation}
		\left(
			\translInpimgAstComplexWeightimgSub
		\right)[\vectorindex] = \bigl(
			\transl_{\samplinterv\translvecDiscrete}\inpfunAstWaveletInterp
		\bigr)\left(
			\sub\samplinterv\vectorindex
		\right).
	\label{eq:discrete2continuous_shift}
	\end{equation}
\end{corollary}

\begin{proof}
	Applying \eqref{eq:discrete2continuous} in \cref{prop:discrete2continuous} with $\inpimg \leftarrow \translInpimg$, we get
	\begin{equation}
		\left(
			\translInpimgAstComplexWeightimgSub
		\right)[\vectorindex] = \bigl(
			\inpfunAstWaveletTranslInterp
		\bigr)\left(\sub\samplinterv \vectorindex\right),
	\label{eq:discrete2continuousTransl}
	\end{equation}
	and \cref{lemma:commut_interp_transl} concludes the proof.
\end{proof}

\subsection{Shift Invariance in the Discrete Framework}

We consider the \cmod operator defined in \eqref{eq:cmod}.
For the sake of conciseness, in what follows we will write $\cmodopSub$ instead of $\cmodopSubWeight$, when no ambiguity is possible.
First, we state the following lemma.

\begin{lemma}
	For any input image $\inpimg \in \realltwoZsq$ with finite support, and any Gabor-like filter $\complexWeightimg \in \DiscreteGaborfilterGen$, we consider the low-frequency function
	\begin{equation}
		\lowfreqfun: \spatialvec \mapsto (\inpfunAstWaveletInterp)(\spatialvec) \, \rme^{i\innerprod{\freqvecMpipi / \samplinterv}{\spatialvec}},
	\label{eq:lowfreqfun_interp}
	\end{equation}
	with $\inpfunInterp$ and $\waveletInterp$ satisfying \eqref{eq:interp}. If $\supportsizeDiscrete \leq \pi / \sub$, then
	\begin{equation}
		\lowfreqfun \in \shannonspaceSubsamplinterv.
	\label{eq:suppLowfreqfunSubsamp}
	\end{equation}
	Moreover, for any $\translvecContinuous \in \mathR^2$,
	\begin{equation}
		\sumoverVectorindices \Bigl|
			\transl_{\translvecContinuous}\lowfreqfun(\subsamplinterv\vectorindex) - \lowfreqfun(\subsamplinterv\vectorindex)
		\Bigr|^2 = \frac1{\subsamplinterv^2} \normLtwo{\transl_{\translvecContinuous}\lowfreqfun - \lowfreqfun}^2,
	\label{eq:normequality_1}
	\end{equation}
	where we have denoted $\subsamplinterv := 2\sub \samplinterv$. Finally,
	\begin{equation}
		\bignormtwo{\cmodopSub\inpimg} = \frac1{\subsamplinterv} \normLtwo{\lowfreqfun}.
	\label{eq:normequality_2}
	\end{equation}
\label{lemma:normequality}
\end{lemma}

\begin{proof}
	See \cref{subsec:appendix_proofs_lemma_normequality}.
\end{proof}

We are now ready to state the main result about shift invariance of \cmod outputs.

\begin{theorem}[Shift invariance of \cmod]
	\label{th:shiftinvariance_cmod}
	Let $\complexWeightimg \in \DiscreteGaborfilterGen$ denote a discrete Gabor-like filter and $\sub \in \nonzeroMathN$ denote a subsampling factor. Then, under the following condition:
	\begin{equation}
		\supportsizeDiscrete \leq \pi / \sub,
	\label{eq:condition_bandwidth}
	\end{equation}
	we have, for any input image $\inpimg \in \realltwoZsq$ with finite support and any translation vector $\translvecDiscrete \in \mathR^2$,
	\begin{equation}
		\bignormtwo{\cmodopSub (\translInpimg) - \cmodopSub \inpimg} \leq \evalShiftinvCmod \, \bignormtwo{\cmodopSub \inpimg},
	\label{eq:shiftinvariance_cmod}
	\end{equation}
	where $\shiftinvCmod$ has been defined in \eqref{eq:shiftinvariance_continuous_multfac}.
\end{theorem}

\begin{proof}
	See \cref{subsec:appendix_proofs_th_shiftinvariance_cmod}.
\end{proof}

Interestingly, the reference value used in \cref{th:shiftinvariance_cmod}, \ie, $\bignormtwo{\cmodopSub \inpimg}$, is fully shift-invariant, as stated in the following proposition.

\begin{proposition}
	\label{prop:shiftinvariance_normcmod}
	Let $\complexWeightimg \in \DiscreteGaborfilterGen$ and $\sub \in \nonzeroMathN$. Under condition \eqref{eq:condition_bandwidth}, we have, for any $\inpimg \in \realltwoZsq$ and any $\translvecDiscrete \in \mathR^2$,
	\begin{equation}
		\bignormtwo{\cmodopSub(\translInpimg)} = \bignormtwo{\cmodopSub \inpimg}.
	\label{eq:shiftinvariance_normcgmod}
	\end{equation}
\end{proposition}

\begin{proof}
	See \cref{subsec:appendix_proofs_prop_shiftinvariance_normcmod}.
\end{proof}

\section{From \cmod to \rmax}
\label{sec:cmod_rmax}

\cmod operators are found in ScatterNets and complex-valued convolutional networks \citep{Tygert2016}. However, they are absent from conventional, freely-trained CNN architectures. Therefore, \cref{th:shiftinvariance_cmod} cannot be applied as is. Instead, the first convolution layer contains real-valued kernels, and is generally followed by ReLU and max pooling. As shown in \cref{sec:multichannel}, this process can be described with \rmax operators, such as defined in \eqref{eq:rmax}.

As explained in \cref{subsec:intro_motivation}, an important number of trained convolution kernels exhibit oscillating patterns with well-defined frequencies and orientations. To elaborate, let $\weightimg \in \realltwoZsq$ denote such a trained kernel, and consider $\complexWeightimg \in \complexltwoZsq$ as the complex-valued companion of $\weightimg$ satisfying \eqref{eq:hilberttransform}. Then, $\complexWeightimg$ has its energy concentrated in a small region of the Fourier domain. We thus state the hypotheses that $\complexWeightimg \in \DiscreteGaborfilterGen$ \eqref{eq:gaborfilt_discrete} for a certain value of $\freqvecMpipi \in \mpipi^2$ and $\supportsizeDiscrete \in \zeroexcltwopi$. For the sake of conciseness, from now on we write $\rmaxopSubGrid$ instead of $\rmaxopSubGridWeight$, when no ambiguity is possible.
In what follows, we establish conditions on $\complexWeightimg$ under which \cmod \eqref{eq:cmod} and \rmax \eqref{eq:rmax} operators produce comparable outputs. The final goal, achieved in \cref{sec:invariance_rmax}, is to provide a shift invariance bound for \rmax.

To give an intuition about why \rmax may act as a proxy for \cmod, we place ourselves in the continuous framework. Consider the real-valued wavelet transform output $\Real\inpfunbis := \inpfunAstRealWavelet$, employed in \rmax, as the real part of the complex-valued wavelet transform output $\inpfunbis := \inpfunAstWavelet$, used in \cmod. At a given location $\spatialvec \in \mathR^2$, the corresponding imaginary part may carry a large amount of information, which somehow needs to be retrieved. The key idea is that, if $\wavelet$ is sufficiently localized in the Fourier domain, then only the phase of $\inpfunbis$ significantly varies in the vicinity of $\spatialvec$, whereas its magnitude remains nearly constant. Therefore, finding the maximum value of $\Real\inpfunbis$ within a local neighborhood around $\spatialvec$ is nearly equivalent to shifting the phase of $\inpfunbis(\spatialvec)$ towards $0$. The resulting value then approximates $|\inpfunbis(\spatialvec)|$. To put it differently, max pooling
pushes energy towards lower frequencies, in a similar way as the modulus does for complex-valued transforms \citep{Bruna2013}. This result is hinted in \cref{subsec:cmod_rmax_continuous}.

Regretfully, things do not work so smoothly in the discrete case. At first glance, this is surprising because Shannon's sampling theorem allows to cast discrete problems into the continuous framework, as done in \cref{subsec:invariance_cmod_discrete2continuous}. However, as explained in \cref{subsec:cmod_rmax_discrete}, max pooling operates over a discrete grid instead of a continuous window. Consequently, in some situations, the maximum value may fall far away from any zero-phase coefficient. Taking into account this behavior, we adopt a probabilistic point of view, as detailed in \cref{subsec:cmod_rmax_probaframework}. Then, we provide in \cref{subsec:cmod_rmax_expectedmse} an upper bound for the expected gap between \cmod and \rmax outputs.

\subsection{Continuous Framework}
\label{subsec:cmod_rmax_continuous}

This section, inspired from \citet[pp.~190--191]{Waldspurger2015}, provides an intuition about resemblance between \rmax and \cmod in the continuous framework. As will be highlighted in \cref{subsec:cmod_rmax_discrete}, adaptation to discrete 2D sequences is not straightforward and will require a probabilistic approach.

We consider an input function $\inpfun \in \realLtwoRsq$ and a band-pass filter $\wavelet \in \GaborfilterGen$. Let us also consider
\begin{equation}
	\cosfun: (\spatialvec,\, \searchvec) \mapsto \cos\bigl(\innerprod{\freqvec}{\searchvec} - \phase(\spatialvec)\bigr),
\label{eq:cos}
\end{equation}
where $\phase: \mathR^2 \to \zerotwopiexcl$ denotes the phase of $\inpfunAstWavelet$. \Cref{lemma:lowfreqfun} introduced low-frequency functions $\lowfreqfun$, with slow variations. In a nutshell, since $\supp\lowfreqfun \subset \fourierWindowGaborContinuous$, we can write
\begin{equation}
	\normtwo{\searchvec} \ll \wavelength_{\lowfreqfun} \implies \lowfreqfun(\spatialvec + \searchvec) \approx \lowfreqfun(\spatialvec),
\label{eq:approx_lowfreqfun}
\end{equation}
where we have defined $\wavelength_{\lowfreqfun} := 2\pi / \supportsizeContinuous$.
Therefore, according to \cref{prop:approx_periodicfun} below, we get the following approximation of $\inpfunAstRealWavelet$ in a neighborhood around any point $\spatialvec \in \mathR^2$:
\begin{equation}
	\normtwo{\searchvec} \ll \wavelength_{\lowfreqfun} \implies (\inpfunAstRealWavelet)(\spatialvec + \searchvec) \approx \bigl|(\inpfunAstWavelet)(\spatialvec)\bigr| \, \evalCosfun.
\label{eq:approx_conv}
\end{equation}

\begin{proposition}
	For any $\searchvec \in \mathR^2$,
	\begin{equation}
		\left|
			(\inpfunAstRealWavelet)(\spatialvec + \searchvec) - \bigl|
				(\inpfunAstWavelet)(\spatialvec)
			\bigr| \, \evalCosfun
		\right| \leq \bigl|
			\lowfreqfun(\spatialvec + \searchvec) - \lowfreqfun(\spatialvec)
		\bigr|.
	\label{eq:approx_periodicfun}
	\end{equation}
\label{prop:approx_periodicfun}
\end{proposition}

\begin{proof}
	See \cref{subsec:appendix_proofs_prop_approx_periodicfun}.
\end{proof}

On the one hand, we consider a continuous equivalent of the \cmod operator $\cmodopSubWeight$ as introduced in \eqref{eq:cmod}. Such an operator, denoted by $\cmodopWavelet$, is defined, for any $\inpfun \in \realLtwoRsq$, by
\begin{equation}
	\cmodopWavelet (\inpfun): \spatialvec \mapsto \left|(\inpfunAstWavelet)(\spatialvec)\right|.
\label{eq:cmod_continuous}
\end{equation}
On the other hand, we consider the continuous counterpart of \rmax as introduced in \eqref{eq:rmax}. It is defined as the maximum value of $\inpfunAstRealWavelet$ over a sliding spatial window of size $\spatialWindowsize > 0$. This is possible because $\inpfun$ and $\Real\flippedWavelet$ both belong to $\realLtwoRsq$, and therefore $\inpfunAstRealWavelet$ is continuous. Such an operator, denoted by $\rmaxopWinWavelet$, is defined, for any $\inpfun \in \realLtwoRsq$, by
\begin{equation}
	\rmaxopWinWavelet(\inpfun): \spatialvec \mapsto \maxSearchvecInSpatialwindow (\inpfunAstRealWavelet)(\spatialvec + \searchvec).
\label{eq:rmax_continuous}
\end{equation}
For the sake of conciseness, the parameter between square brackets is ignored from now on.
If $\spatialWindowsize \ll \wavelength_{\lowfreqfun}$, then \eqref{eq:approx_conv} is valid for any $\searchvec \in \spatialWindow$. Then, using \eqref{eq:cmod_continuous} and \eqref{eq:rmax_continuous}, we get
\begin{equation}
	\spatialWindowsize \ll \wavelength_{\lowfreqfun} \implies \rmaxopWavelet \inpfun(\spatialvec)
	\approx \cmodop \inpfun(\spatialvec) \maxSearchvecInSpatialwindow \evalCosfun.
\label{eq:approx_rmax_continuous}
\end{equation}
Using the periodicity of $\cosfun$, we can show that, if $\spatialWindowsize \geq \frac{\pi}{\normtwo{\freqvec}}$, then $\searchvec \mapsto \evalCosfun$ necessarily reaches its maximum value (\ie, $1$) on $\spatialWindow$. We therefore get
\begin{equation}
	\frac{\pi}{\normtwo{\freqvec}} \leq \spatialWindowsize \ll \frac{2\pi}{\supportsizeContinuous} \implies \rmaxopWavelet \inpfun(\spatialvec)
	\approx \cmodop \inpfun(\spatialvec).
\label{eq:equiv_cmodrmax_continuous}
\end{equation}

\subsection{Adaptation to Discrete 2D Sequences}
\label{subsec:cmod_rmax_discrete}

As in \cref{subsec:invariance_cmod_discrete2continuous}, we consider an input image $\inpimg \in \realltwoZsq$, a complex, analytic convolution kernel $\complexWeightimg \in \DiscreteGaborfilterGen$, a subsampling factor $\sub \in \nonzeroMathN$ and an integer $\gridhalfsize \in \nonzeroMathN$, referred to as a \emph{half-size}, such that max pooling operates on a grid of size $(2\gridhalfsize + 1) \times (2\gridhalfsize + 1)$. We seek a relationship between
\begin{equation}
	\rmaximg := \rmaxopSubGridWeight(\inpimg) \qqand \cmodimg := \cmodopSubWeight(\inpimg),
\label{eq:outimg_rmax_cmod}
\end{equation}
where $\rmaxopSubGridWeight$ (\rmax) and $\cmodopSubWeight$ (\cmod) have been respectively defined in \eqref{eq:rmax} and \eqref{eq:cmod}. As before, in what follows we omit the parameter between square brackets.

\begin{figure}
	\centering
	\includegraphics[width=0.22\textwidth]{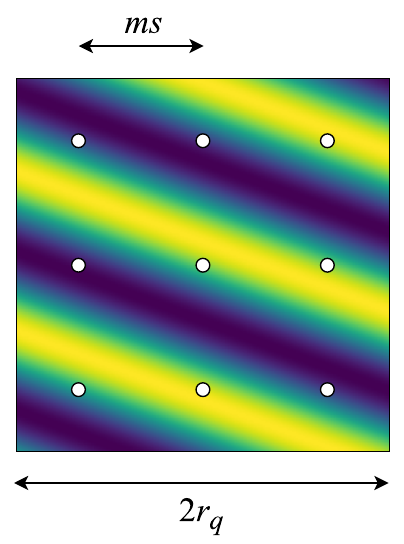}
	\hspace{20pt}
	\includegraphics[width=0.22\textwidth]{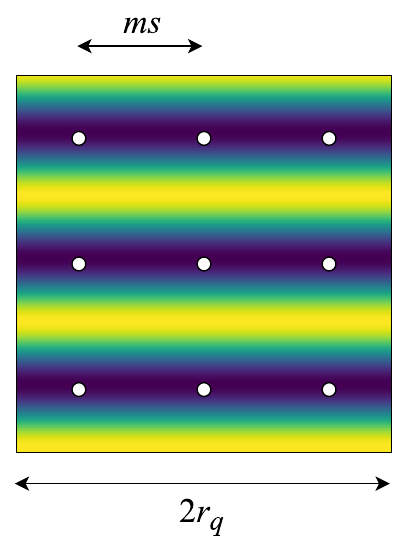}
	\hspace{20pt}
	\includegraphics[width=0.22\textwidth]{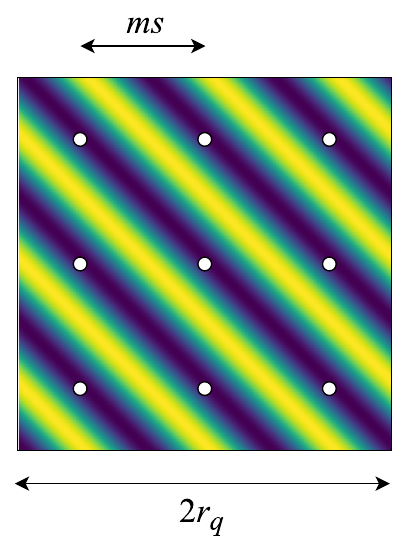}
	\caption[
		Max Pooling Grid
	]{
		Search for the maximum value of $\searchvec \mapsto \cosfunInpimg(\spatialvec,\, \searchvec)$ over a discrete grid of size $3 \times 3$, \ie, $\gridhalfsize = 1$. This figure displays $3$ examples with different frequencies $\freqvec := \freqvecMpipi / \samplinterv$ and phases $\phaseInterp(\spatialvec)$. Hopefully the result will be close to the true maximum (left), but there are some pathological cases in which all points in the grid fall into pits (middle and right).
	}
	\label{fig:discretegrid}
\end{figure}

We now use the sampling results from \cref{prop:discrete2continuous}. Let $\inpfunInterp$ and $\waveletInterp \in \shannonspaceSamplinterv$ denote the functions satisfying \eqref{eq:interp}. Recall that the continuous versions of \cmod and \rmax operators have been defined in \eqref{eq:cmod_continuous} and \eqref{eq:rmax_continuous}, respectively. On the one hand, we apply \eqref{eq:discrete2continuous} with $\sub \leftarrow 2\sub$ to $\cmodimg$. For any $\vectorindexInZtwo$,
\begin{align}
	\detailCmodimg[\vectorindex]
		&= (\inpfunAstWaveletInterp)(\spatialvecSelect)
\label{eq:discrete2continuous_cmod} \\
		&= \cmodop \inpfunInterp(\spatialvecSelect),
\label{eq:discrete_right}
\end{align}
with $\spatialvecSelect := 2\sub \samplinterv\vectorindex$.
On the other hand, we postulate that
\begin{equation}
	\detailRmaximg[\vectorindex] = \rmaxopWavelet \inpfunInterp(\spatialvecSelect)
\label{eq:discrete_left}
\end{equation}
for a certain value of $\spatialWindowsize \in \nonzeroMathRplus$. Then, \eqref{eq:equiv_cmodrmax_continuous} implies $\cmodimg \approx \rmaximg$.
However, as explained hereafter, \eqref{eq:discrete_left} is not satisfied, due to the discrete nature of the max pooling grid.
According to \eqref{eq:rmax} and \eqref{eq:maxpool}, we have
\begin{equation}
	\detailRmaximg[\vectorindex] = \maxVectorindexInPoolinggrid \Real\left(
		\bigl(
			\inpimg \ast \flippedComplexWeightimg
		\bigr) \downarrow \sub
	\right)[2\vectorindex + \vectorindexbis].
\end{equation}
Therefore, according to \eqref{eq:discrete2continuous} in \cref{prop:discrete2continuous}, we get
\begin{equation}
\begin{split}
	\detailRmaximg[\vectorindex]
		&= \maxVectorindexInPoolinggrid(\inpfunAstRealWaveletInterp)\left(\spatialvecSelect + \searchvecSelect\right),
\end{split}
\label{eq:discrete2continuous_rmax}
\end{equation}
with
\begin{equation}
	\spatialvecSelect := 2\sub \samplinterv\vectorindex \qqand \searchvecSelect := \sub \samplinterv \vectorindexbis.
\label{eq:2dpoints_continuous}
\end{equation}
By considering $\spatialWindowsizeDiscretegrid := \sub \samplinterv \left(\gridhalfsize+\frac12\right)$, we get a variant of \eqref{eq:discrete_left} in which the maximum is evaluated on a discrete grid of $(2\gridhalfsize+1)^2$ elements, instead of the continuous region $\spatialWindowDiscretegrid$, as defined in \eqref{eq:rmax_continuous} with $\spatialWindowsize \leftarrow \spatialWindowsizeDiscretegrid$. As a consequence, \eqref{eq:approx_rmax_continuous} is replaced in the discrete framework by
\begin{equation}
	\gridhalfsize \ll 2\pi / (\sub\supportsizeDiscrete) \quad\implies\quad
	\detailRmaximg[\vectorindex] \approx
	\detailCmodimg[\vectorindex] \, \maxVectorindexInPoolinggrid \evalCosfunInpimgGrid,
\label{eq:discrete_left0}
\end{equation}
where we have introduced, similarly to \eqref{eq:cos},
\begin{equation}
	\cosfunInpimg: (\spatialvec,\, \searchvec) \mapsto \cos\bigl(
		\innerprod{\freqvec}{\searchvec} - \phaseInterp(\spatialvec)
	\bigr),
\label{eq:cos_discrete}
\end{equation}
with
\begin{equation}
	\freqvec := \freqvecMpipi / \samplinterv \qqand \phaseInterp := \angle\left(
		\inpfunAstWaveletInterp
	\right),
\label{eq:freqvec_complexPhase}
\end{equation}
where $\angle: \mathC \to \zerotwopiexcl$ denotes the phase operator.
Unlike the continuous case, even if the window size $\spatialWindowsizeDiscretegrid$ is large enough, the existence of $\vectorindexInPoolinggrid$ such that $\evalCosfunInpimgGrid = 1$ is not guaranteed, as illustrated in \cref{fig:discretegrid} with $\gridhalfsize = 1$. Instead, we can only seek a probabilistic estimation of the normalized mean squared error between $\rmaximg$ and $\cmodimg$.

Approximation \eqref{eq:discrete_left0} implies
\begin{equation}
	\gridhalfsize \ll 2\pi / (\sub\supportsizeDiscrete) \implies \bignormtwo{\detailCmodimg - \detailRmaximg} \approx \normtwo{\deltaopSubGrid\inpimg},
\label{eq:approxdiffenergy}
\end{equation}
where $\deltaopSubGrid\inpimg \in \realltwoZsq$ is defined such that, for any $\vectorindex \in \mathZ^2$,
\begin{equation}
	\deltaopSubGrid\inpimg[\vectorindex] := \detailCmodimg[\vectorindex] \, \left(1 - \maxVectorindexInPoolinggrid \evalCosfunInpimgGrid\right).
\label{eq:delta_outimg_cmod}
\end{equation}
Expression \eqref{eq:approxdiffenergy} suggests that the difference between the left and right terms can be bounded by a quantity which only depends on the product $\sub \supportsizeDiscrete$ (subsampling factor $\times$ frequency localization) and the grid half-size $\gridhalfsize$. In what follows, we establish a bound characterizing this approximation, which will be provided in \cref{prop:diffmodpool}.

For the sake of conciseness, we introduce the following notations:
\begin{align}
	\exactfunInterp: (\spatialvec,\, \searchvec)
		&\mapsto (\inpfunAstRealWaveletInterp)(\spatialvec + \searchvec); \\
	\approxfunInterp: (\spatialvec,\, \searchvec)
		&\mapsto \bigl|
			(\inpfunAstWaveletInterp)(\spatialvec)
		\bigr| \, \cosfunInpimg(\spatialvec, \searchvec).
\end{align}
We now consider, for any $\vectorindexInZtwo$, the vectors $\searchvecMaxSelect$ and $\searchvecApproxMaxSelect \in \sub\samplinterv\poolinggrid$ achieving the maximum value of $\exactfunInterp(\spatialvecSelect,\, \searchvecSelect)$ and $\approxfunInterp(\spatialvecSelect,\, \searchvecSelect)$ over the max pooling grid, respectively. They satisfy
\begin{align}
	\evalMaxExactfunInterp
		&:= \evalExactfunInterp = \maxVectorindexInPoolinggrid\exactfunInterp(\spatialvecSelect,\, \searchvecSelect); \\
	\evalMaxApproxfunInterp
		&:= \evalApproxfunInterp = \maxVectorindexInPoolinggrid\approxfunInterp(\spatialvecSelect,\, \searchvecSelect).
\end{align}
Then, according to \eqref{eq:discrete2continuous_cmod} and \eqref{eq:discrete2continuous_rmax}, we get, for any $\vectorindexInZtwo$,
\begin{align}
	\evalMaxExactfunInterp
		&= \detailRmaximg[\vectorindex];
\label{eq:evalMaxExactfunInterp} \\
	\evalMaxApproxfunInterp
		&= \detailCmodimg[\vectorindex] \, \maxVectorindexInPoolinggrid \evalCosfunInpimgGrid,
\label{eq:evalMaxApproxfunInterp}
\end{align}
and \eqref{eq:discrete_left0} becomes
\begin{equation}
	\gridhalfsize \ll 2\pi / (\sub\supportsizeDiscrete) \quad\implies\quad
	\evalMaxExactfunInterp \approx \evalMaxApproxfunInterp.
\label{eq:approxfun}
\end{equation}

\begin{remark}
	Expression~\eqref{eq:approx_conv} implies that, if $\gridhalfsize \ll 2\pi / (\sub\supportsizeDiscrete)$, then $\exactfunInterp(\spatialvecSelect,\, \searchvecSelect) \approx \approxfunInterp(\spatialvecSelect,\, \allowbreak\searchvecSelect)$ for all $\vectorindexbis \in \poolinggrid$. However, this property does not guarantee that $\exactfunInterp$ and $\approxfunInterp$ reach their maximum in the same exact location; \ie, that $\searchvecMaxSelect = \searchvecApproxMaxSelect$.
\end{remark}

The following lemma provides a bound for approximation \eqref{eq:approxfun}.
\begin{lemma}
	For any $\spatialvec \in \mathR^2$,
	\begin{equation}
		\left|
			\evalMaxExactfunInterp - \evalMaxApproxfunInterp
		\right| \leq
		\max\limits_{
			\searchvec \in \left\{
				\searchvecMaxSelect,\, \searchvecApproxMaxSelect
			\right\}
		}\Bigl|
			\lowfreqfun(
				\spatialvecSelect + \searchvec
			) - \lowfreqfun(\spatialvecSelect)
		\Bigr|.
	\label{eq:true_versus_false_maxvalues}
	\end{equation}
\label{lemma:true_versus_false_maxvalues}
\end{lemma}

\begin{proof}
	See \cref{subsec:appendix_proofs_lemma_true_versus_false_maxvalues}.
\end{proof}

Before stating \cref{prop:diffmodpool}, we consider the following hypothesis:
\begin{hypothesis}
	There exists $\searchvec_0 \in \mathR^2$ with $\normtwo{\searchvec_0} = \sqrt{2}\gridhalfsize\sub\samplinterv$, such that
	\begin{equation}
		\sumoverVectorindices\,\max\limits_{
			\searchvec \in \left\{
				\searchvecMaxSelect,\, \searchvecApproxMaxSelect
			\right\}
		}\Bigl|
			\lowfreqfun(
				\spatialvecSelect + \searchvec
			) - \lowfreqfun(\spatialvecSelect)
		\Bigr|^2
		\leq
		\sumoverVectorindices\Bigl|
			\lowfreqfun(
				\spatialvecSelect + \searchvec_0
			) - \lowfreqfun(\spatialvecSelect)
		\Bigr|^2.
	\end{equation}
\label{hyp:bound_sumofdiffs}
\end{hypothesis}
The underlying idea is explained as follows. The absolute difference between $\lowfreqfun(\spatialvecSelect + \searchvec)$ and $\lowfreqfun(\spatialvecSelect)$ is more likely to increase with the norm of $\searchvec$. For any given $\vectorindexInZtwo$, we have, by construction, $\normtwo{\searchvecMaxSelect} \leq \sqrt{2}\gridhalfsize\sub\samplinterv$ and $\normtwo{\searchvecApproxMaxSelect} \leq \sqrt{2}\gridhalfsize\sub\samplinterv$. Therefore, we can expect to observe
\begin{equation}
	\max\limits_{
			\searchvec \in \left\{
				\searchvecMaxSelect,\, \searchvecApproxMaxSelect
			\right\}
		}\Bigl|
			\lowfreqfun(
				\spatialvecSelect + \searchvec
			) - \lowfreqfun(\spatialvecSelect)
		\Bigr|^2 \leq \Bigl|
			\lowfreqfun(
				\spatialvecSelect + \searchvec_0
			) - \lowfreqfun(\spatialvecSelect)
		\Bigr|^2.
\end{equation}
While this might occasionally not be true, \cref{hyp:bound_sumofdiffs} postulates that, when summing over all the datapoints, the inequality holds.

We now formally state the result characterizing approximation \eqref{eq:approxdiffenergy}.

\begin{proposition}
	We assume that condition \eqref{eq:condition_bandwidth} is satisfied: $\supportsizeDiscrete \leq \pi / \sub$. Then, under \cref{hyp:bound_sumofdiffs},
	\begin{equation}
		\bignormtwo{\detailCmodimg - \detailRmaximg} \leq \bignormtwo{\deltaopSubGrid\inpimg} + \evalLowerboundDiffRmaxCmod \, \bignormtwo{\detailCmodimg},
	\label{eq:diffmodpool}
	\end{equation}
	where $\lowerboundDiffRmaxCmod_\gridhalfsize: \mathR_+ \to \mathR_+$ is defined by
	\begin{equation}
		\lowerboundDiffRmaxCmod_\gridhalfsize: \supportsizeDiscrete' \mapsto \gridhalfsize\supportsizeDiscrete'.
	\label{eq:diffmodpool_multfactor}
	\end{equation}
\label{prop:diffmodpool}
\end{proposition}

\begin{proof}
	See \cref{subsec:appendix_proofs_prop_diffmodpool}.
\end{proof}


We now seek a probabilistic estimation of $\bignormtwo{\deltaopSubGrid\inpimg}$. For this purpose, we first reformulate the problem using the unit circle $\unitcircle \subset \mathC$, before introducing a probabilistic framework in \cref{subsec:cmod_rmax_probaframework}.

\subsection{Notations on the Unit Circle}
\label{subsec:cmod_rmax_unitcircle}

In what follows, for any $\pointunitcircle \in \nonzeroMathC$, we denote by $\angle \pointunitcircle \in \zerotwopiexcl$ the argument of $\pointunitcircle$. For any $\pointunitcircle$, $\pointunitcircle' \in \unitcircle$, the angle between $\pointunitcircle$ and $\pointunitcircle'$ is given by $\angleArc$. We then denote by $\arcPointToPoint \subset \unitcircle$ the arc on the unit circle going from $\pointunitcircle$ to $\pointunitcircle'$ counterclockwise:
\begin{equation}
	\arcPointToPoint := \set{\pointunitcircle'' \in \unitcircle}{\angle(\pointunitcircle^\ast \pointunitcircle'') \leq \angleArc}.
\label{eq:arc}
\end{equation}

We remind readers that $\spatialvecSelect$ and $\searchvecSelect \in \mathR^2$ have been defined in \eqref{eq:2dpoints_continuous}. By using the relation $\cos\alpha = \Real(\rme^{i\alpha})$, \eqref{eq:cos_discrete} becomes, for any $\vectorindex \in \mathZ^2$ and any $\vectorindexInPoolinggrid$,
\begin{equation}
	\evalCosfunInpimgGrid = \Real\bigl(\complexPhaseInterp^\ast(\spatialvecSelect) \, \evalComplexPhaseshift\bigr),
\label{eq:cos_discretegrid}
\end{equation}
where we have defined the following functions with outputs on the unit circle:
\begin{equation}
	\complexPhaseInterp
		: \spatialvec \mapsto \rme^{i\,\phaseInterp(\spatialvec)} 
	\qqand
	\complexPhaseshift
		: \freqvecMpipibis \mapsto = \rme^{i\innerprod{\freqvecMpipibis}{\vectorindexbis}},
\label{eq:projunitcircle}
\end{equation}
where $\phaseInterp$ denotes the phase of $\inpfunAstWaveletInterp$ as introduced in \eqref{eq:freqvec_complexPhase}. On the one hand, $\complexPhaseInterp(\spatialvecSelect)$ is the phase (represented on the unit circle $\unitcircle$) of the complex wavelet transform $\inpfunAstWaveletInterp$ at location $\spatialvecSelect$. On the other hand, $\evalComplexPhaseshift$ approximates the phase shift between any two evaluations of $\inpfunAstWaveletInterp$ at locations $\spatialvec,\, \spatialvec'$ such that $\spatialvec' - \spatialvec = \searchvecSelect$. This however is only true if we assume that $\waveletInterp$ exhibits slow amplitude variations. Then, $\evalCosfunInpimgGrid$ approximates the cosine of the phase of $\inpfunAstWaveletInterp$ at location $\spatialvecSelect + \searchvecSelect$.

According to \eqref{eq:discrete_left0}, $\maxVectorindexInPoolinggrid \evalCosfunInpimgGrid$ approximates the ratio between \rmax and \cmod outputs at discrete location $\vectorindex \in \mathZ^2$. The intuition behind this is that max pooling seeks a point in a discrete grid around $\spatialvecSelect$ where the phase of $\inpfunAstWaveletInterp$ is the closest to $1$, thereby maximizing the amount of energy on the real part of the signal. Assuming slow amplitude variations of $\waveletInterp$, the result therefore approximates the modulus of the complex coefficients.

To get an estimation of $\deltaopSubGrid\inpimg[\vectorindex]$ \eqref{eq:delta_outimg_cmod}, we will exploit the following property. If the phases $\evalComplexPhaseshift$ for $\vectorindexInPoolinggrid$ are well distributed on the unit circle, then the values of $\evalCosfunInpimgGrid$ are evenly spread out on $\interval{-1}{1}$. Therefore, its maximum value is more likely to be close to $1$, and \eqref{eq:delta_outimg_cmod} becomes
\begin{equation}
	\deltaopSubGrid\inpimg[\vectorindex] \ll \detailCmodimg[\vectorindex] \qquad \forall \vectorindexInZtwo.
\end{equation}

Let $\nevalpointsGrid := (2\gridhalfsize+1)^2$ denote the number of evaluation points for the max pooling operator. For any $\freqvecMpipibis \in \mathR^2$, we consider a sequence of values on $\unitcircle$, denoted by $\setofEvalbisOrderedComplexPhaseshift$, obtained by sorting $\setofEvalbisComplexPhaseshift$ \eqref{eq:projunitcircle} in ascending order of their arguments:
\begin{equation}
	0 = \orderedPhase{0}(\freqvecMpipibis) \leq \dots \leq \orderedPhase{\nevalpointsGrid - 1}(\freqvecMpipibis) < 2\pi,
\label{eq:phaseshift}
\end{equation}
where $\orderedPhase{\selectEvalpoint}(\freqvecMpipibis)$ denotes the phase of $\evalbisOrderedComplexPhaseshift$. In addition, we close the loop with $\orderedPhase{\nevalpointsGrid}(\freqvecMpipibis) := 2\pi$ and $\orderedComplexPhaseshift{\nevalpointsGrid}(\freqvecMpipibis) := 1$. Then, we split $\unitcircle$ into $\nevalpointsGrid$ arcs delimited by $\evalbisOrderedComplexPhaseshift$:
\begin{equation}
	\evalbisArcSelect := \begin{cases}
		\arc{\orderedComplexPhaseshiftSelect\!(\freqvecMpipibis)}{\orderedComplexPhaseshiftSelectNext(\freqvecMpipibis)} & \mbox{if $\orderedPhase{\selectEvalpoint+1}(\freqvecMpipibis) - \orderedPhase{\selectEvalpoint}(\freqvecMpipibis) < 2\pi;$} \\
		\unitcircle & \mbox{otherwise.}
	\end{cases}
\label{eq:arcunitcircle}
\end{equation}
Finally, for any $\selectEvalpoint \in \setof{\nevalpointsGrid}$, we denote by
\begin{equation}
	\diffOrderedPhaseSelect : \freqvecMpipibis \mapsto \orderedPhase{\selectEvalpoint+1}(\freqvecMpipibis) - \orderedPhase{\selectEvalpoint}(\freqvecMpipibis)
\label{eq:arclength}
\end{equation}
the function computing the angular measure of arc $\evalbisArcSelect$, for any $\freqvecMpipibis \in \mathR^2$.

\subsection{Probabilistic Framework}
\label{subsec:cmod_rmax_probaframework}

From now on, input $\inpimg$ is considered as a discrete 2D stochastic process. In order to ``randomize'' $\inpfunInterp$ introduced in \eqref{eq:interp}, we define a continuous stochastic process from $\inpimg$, denoted by $\stochInpfunInterp$, such that
\begin{equation}
	\forall \spatialvec \in \mathR^2,\, \stochInpfunInterp(\spatialvec) := \sumoverVectorindices \inpimg[\vectorindex] \,\shanScalingfunSamplintervVectorindex(\spatialvec).
\label{eq:interp_stoch}
\end{equation}
Next, we consider the following stochastic processes, which are parameterized by $\inpimg$:
\begin{equation}
	\stochMagnitudeInterp := |\stochInpfunAstWaveletInterp|; \qquad \stochPhaseInterp := \angle (\stochInpfunAstWaveletInterp);
	\qquad \stochComplexPhaseInterp := \rme^{i\stochPhaseInterp},
\label{eq:magnitude_phase_stoch}
\end{equation}
and, for any $\vectorindexInPoolinggrid$,
\begin{align}
	\stochCosfunInterpEvalpoint := \Real\bigl(
		\stochComplexPhaseInterp^\ast \, \evalComplexPhaseshift
	\bigr); \qquad \stochCosmaxInterp := \maxVectorindexInPoolinggrid \stochCosfunInterpEvalpoint,
\label{eq:cos_stoch}
\end{align}
where the deterministic function $\complexPhaseshift$ has been defined in \eqref{eq:projunitcircle}.

\begin{remark}
\label{remark:degeneratedPhase}
	By continuous extension, $\stochPhaseInterp(\spatialvec)$ and $\stochComplexPhaseInterp(\spatialvec)$ are uniquely defined at $\spatialvec$ such that $\stochMagnitudeInterp(\spatialvec) = 0$.
\end{remark}

For any $\spatialvec \in \mathR^2$, $\inpfunInterp(\spatialvec)$ \eqref{eq:interp} and $\phaseInterp(\spatialvec)$ \eqref{eq:freqvec_complexPhase} are respectively drawn from $\stochInpfunInterp(\spatialvec)$ and $\stochPhaseInterp(\spatialvec)$. Then, $\complexPhaseInterp(\spatialvec)$ \eqref{eq:projunitcircle} is a realization of $\stochComplexPhaseInterp(\spatialvec)$. Consequently, according to \eqref{eq:cos_discretegrid}, $\evalCosfunInpimg$ is a realization of $\stochCosfunInterpEvalpoint(\spatialvec)$. Furthermore, according to the definition of \cmod in \eqref{eq:cmod} and $\spatialvecSelect$ in \eqref{eq:2dpoints_continuous}, \cref{prop:discrete2continuous} with $\sub \leftarrow 2\sub$ implies that
\begin{equation}
	\stochMagnitudeInterp(\spatialvecSelect) = \cmodopSub\inpimg[\vectorindex].
\label{eq:modulus_complexconv_stoch}
\end{equation}

We remind that $\freqvecMpipi \in \mpipi^2$ and $\supportsizeDiscrete \in \zeroexcltwopi$ respectively denote the center and size of the Fourier support of the complex kernel $\complexWeightimg \in \DiscreteGaborfilterGen$. To compute the expected discrepancy between $\rmaximg$ and $\cmodimg$ such as introduced in \eqref{eq:outimg_rmax_cmod}, we assume that
\begin{align}
	\normtwo{\freqvecMpipi}
		&\gg 2\pi / \imgsize;
\label{eq:condition_characfreq} \\
	\normtwo{\freqvecMpipi}
		&\gg \supportsizeDiscrete,
\label{eq:condition_characfreq_fourierwindow}
\end{align}
where $\imgsize \in \nonzeroMathN$ denotes the support size of input images. These assumptions exclude low-frequency filters from the scope of our study. We then state the following hypotheses, for which a justification is provided in \cref{sec:appendix_invariance_hyps}.

\begin{hypothesis}
	\label{hyp:uniformdist}
	For any $\spatialvec \in \mathR^2$, $\stochComplexPhaseInterp(\spatialvec)$ is uniformly distributed on $\unitcircle$.
\end{hypothesis}

\begin{hypothesis}
	\label{hyp:indep_phase_modulus}
	For any $\nevalpoints \in \nonzeroMathN$ and $\spatialvec,\, \spatialvecbis_0,\, \dots,\, \spatialvecbis_{\nevalpoints-1} \in \mathR^2$, the random variables $\stochMagnitudeInterp(\spatialvecbis_\selectEvalpoint)$ for $\selectEvalpoint \in \setof{\nevalpoints}$ are jointly independent of $\stochComplexPhaseInterp(\spatialvec)$.
\end{hypothesis}

\subsection{Expected Quadratic Error between \rmax and \cmod}
\label{subsec:cmod_rmax_expectedmse}

In this section, we propose to estimate the expected value of the stochastic quadratic error $\avgStochDiffRmaxCmodInpimg^2$, defined such that
\begin{equation}
	\avgStochDiffRmaxCmodInpimg := \bignormtwo{\detailCmodimg - \detailRmaximg} / \bignormtwo{\detailCmodimg}.
\label{eq:normdiff_cmodrmax_stoch}
\end{equation}
According to \eqref{eq:outimg_rmax_cmod}, this is an estimation of the relative error between $\cmodimg$ and $\rmaximg$.

First, let us reformulate $\deltaopSubGrid\inpimg$, introduced in \eqref{eq:delta_outimg_cmod}, using the probabilistic framework. According to \eqref{eq:cos_discretegrid} and \eqref{eq:cos_stoch}, we have, for any $\vectorindexInZtwo$,
\begin{equation}
	\deltaopSubGrid\inpimg[\vectorindex] := \detailCmodimg[\vectorindex] \, \bigl(
		1 - \stochCosmaxInterp(\spatialvecSelect)
	\bigr).
\label{eq:delta_outimg_cmod_stoch}
\end{equation}
We now consider the stochastic process
\begin{equation}
	\stochOneminusCosmaxInterp := 1 - \stochCosmaxInterp,
\label{eq:oneminuscos_stoch}
\end{equation}
and the random variable
\begin{equation}
	\avgStochOneminusCosmaxInpimg := \normtwo{\deltaopSubGrid\inpimg} / \bignormtwo{\detailCmodimg}.
\label{eq:normdelta_cmod_stoch}
\end{equation}
The next steps are as follows:
\begin{enumerate*}
	\item at the pixel level, show that $\Expval[\stochOneminusCosmaxInterp(\spatialvec)^2]$ depends on the subsampling factor $\sub$ and the filter frequency $\freqvecMpipi$, and remains close to zero with some exceptions;
	\item at the image level, show that the expected value of $\avgStochOneminusCosmaxInpimg^2$ is equal to the latter quantity;
	\item use \cref{prop:diffmodpool}, which implies that $\avgStochDiffRmaxCmodInpimg \approx \avgStochOneminusCosmaxInpimg$, to deduce an upper bound on the expected value of $\avgStochDiffRmaxCmodInpimg^2$.
\end{enumerate*}

The first point, which is established in \cref{prop:expval_oneminuscos} below, is a key result of this paper. It will be used to prove \cref{th:scdmoment_normdiff_cmodrmax}, which corresponds to the two remaining points.%

\begin{proposition}
	\label{prop:expval_oneminuscos}
	Assuming \cref{hyp:uniformdist}, the expected value of $\stochOneminusCosmaxInterp(\spatialvec)^2$ is independent from the choice of $\spatialvec \in \mathR^2$, and
	\begin{equation}
		\Expval\left[\stochOneminusCosmaxInterp(\spatialvec)^2\right] = \proxyfunRmaxCmod_\gridhalfsize(\sub\freqvecMpipi)^2,
	\label{eq:expval_oneminuscos}
	\end{equation}
	where we have defined
	\begin{equation}
		\proxyfunRmaxCmod_\gridhalfsize : \freqvecMpipibis \mapsto \sqrt{
			\frac32 + \frac1{4\pi}  \sumoverEvalpointsGrid \left(
				\sin\evalbisDiffOrderedPhaseSelect - 8 \sin\frac{\evalbisDiffOrderedPhaseSelect}{2}
			\right)
		},
	\label{eq:proxyfunction}
	\end{equation}
	with $\evalbisDiffOrderedPhaseSelect \in \zerotwopi$ \eqref{eq:arclength} being the length of arc $\evalbisArcSelect$.
\end{proposition}

\begin{proof}
	For the sake of readability, in this proof we omit the argument of functions $\complexPhaseshift$ \eqref{eq:projunitcircle}, $\orderedComplexPhaseshiftSelect$, $\orderedPhaseSelect$ \eqref{eq:phaseshift}, $\arcSelect$ \eqref{eq:arcunitcircle}, and $\diffOrderedPhaseSelect$ \eqref{eq:arclength}; we assume they are evaluated at $\freqvecMpipibis \leftarrow \sub\freqvecMpipi$.
	We consider the ``Lebesgue'' Borel \sigmaAlgebra on $\unitcircle$ generated by $\set{\arcPointToPoint}{\pointunitcircle,\, \pointunitcircle' \in \unitcircle} \cup \{\unitcircle\}$, on which we have defined the angular measure $\angularMeasure$ such that $\angularMeasure(\unitcircle) := 2\pi$, and
	\begin{equation}
		\forall \pointunitcircle,\, \pointunitcircle' \in \unitcircle,\, \angularMeasure\left(\arcPointToPoint\right) := \angleArc.
	\end{equation}

	For any $\normdegree \in \nonzeroMathN$, we compute the $\normdegree$-th moment of $\stochCosmaxInterp(\spatialvec)$ defined in \eqref{eq:cos_stoch}. By considering
	\begin{equation}
	\begin{split}
		\maxCosUnitcircle : \unitcircle &\to \interval{-1}{1} \\
		\pointunitcircle &\mapsto \maxVectorindexInPoolinggrid \Real\bigl(
			\pointunitcircle^\ast \complexPhaseshift
		\bigr),
	\end{split}
	\label{eq:maxCosUnitcircle}
	\end{equation}
	we get $\stochCosmaxInterp(\spatialvec) = \maxCosUnitcircle(\stochComplexPhaseInterp(\spatialvec))$. A visual representation of $\maxCosUnitcircle$ is provided in \cref{fig:phaserepartition}, for two different values of $\freqvecMpipi$.
  \begin{figure}[t]
		\centering
		\begin{subfigure}{0.42\textwidth}
			\centering
			\includegraphics[width=0.6\textwidth]{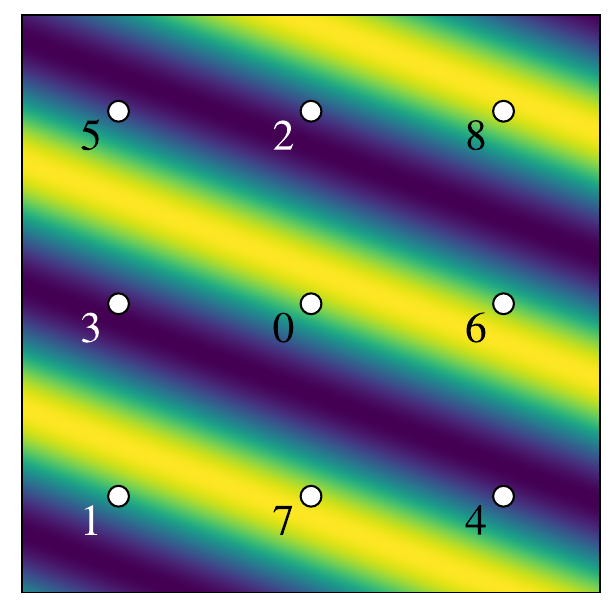}
			\vspace{10pt}
			\includegraphics[width=0.8\textwidth]{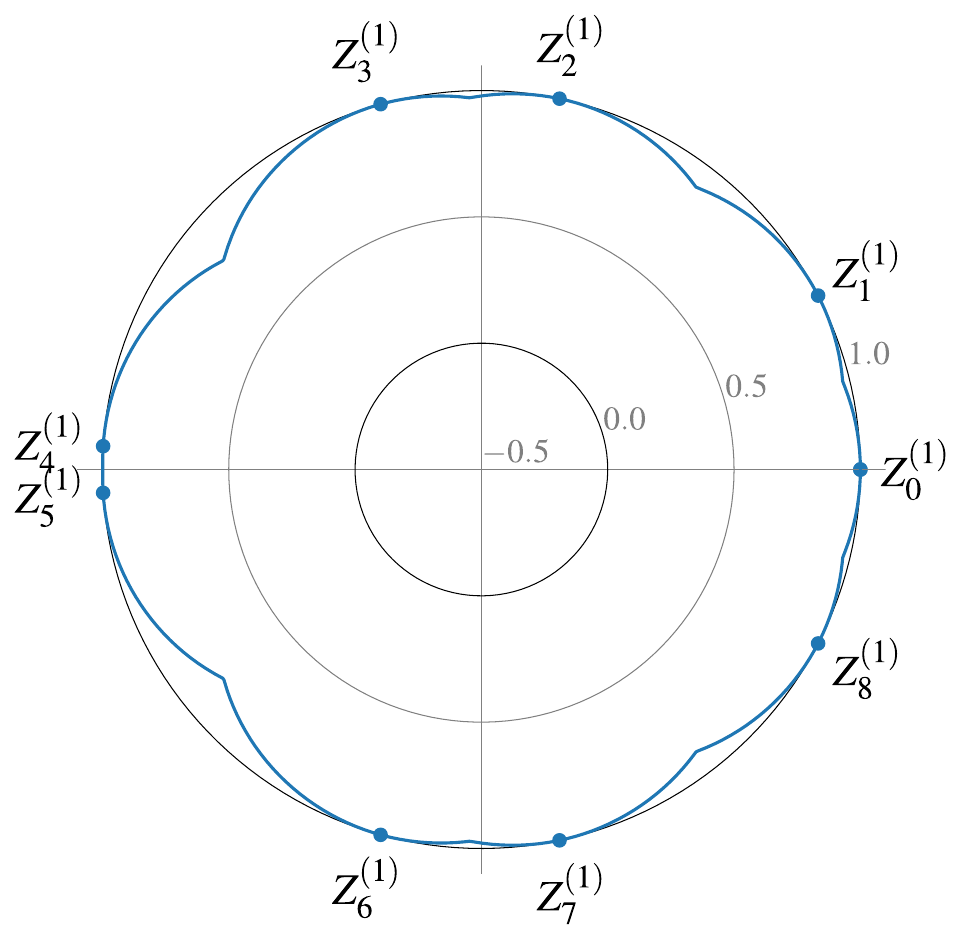}
			\caption{General case}
		\end{subfigure}
		\begin{subfigure}{0.42\textwidth}
			\centering
			\includegraphics[width=0.6\textwidth]{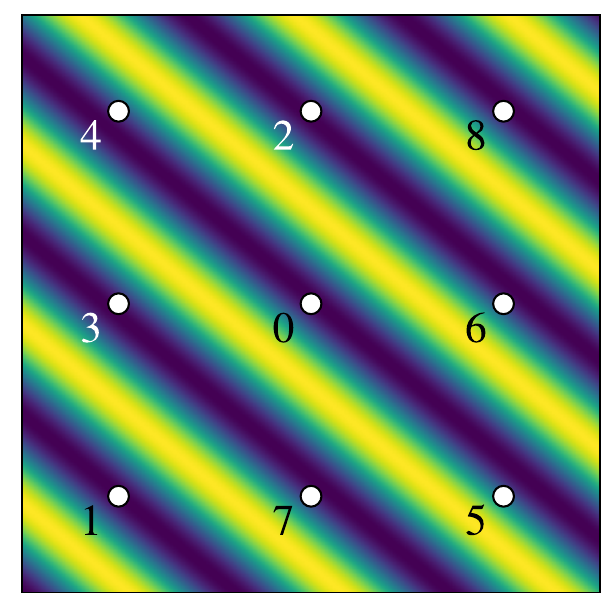}
			\includegraphics[trim={-30 -37 0 -18},clip,width=0.8\textwidth]{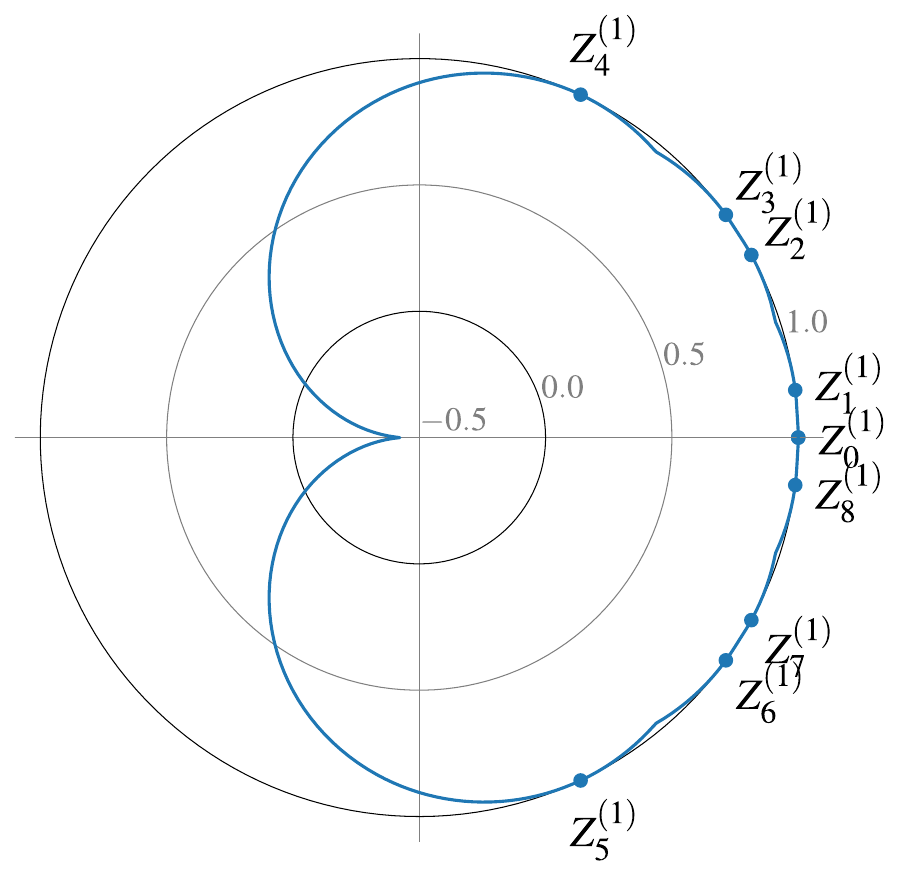}
			\caption{Pathological case}
		\end{subfigure}
		\caption[
			Polar Plots
		]{
			Top: 2D representation of $\searchvec \mapsto \cosfunInpimg(\spatialvecSelect,\, \searchvec)$ \eqref{eq:cos_discrete}, for two different values of $\freqvecMpipi \in \mathR^2$, $\gridhalfsize = 1$ and arbitrary values of $\sub \in \nonzeroMathN$ and $\samplinterv \in \nonzeroMathR$. Assuming the plots are centered around $\searchvec = \bzero$, each point materializes a location $\searchvecSelect$ in the max pooling grid, for $\vectorindexInPoolinggrid$. The desirable situation occurs when one of these locations falls near a ridge (bright areas), in which case the outputs produced by \rmax and \cmod are similar---see \eqref{eq:discrete_left0}. Each number $\selectEvalpoint \in \zeroto{8}$ represents the rank of $\complexPhaseshift \in \unitcircle$ \eqref{eq:projunitcircle}, when these values are sorted by ascending order of their arguments \eqref{eq:phaseshift}. If location $\searchvecSelect$ gets ranked $\selectEvalpoint$, then we have $\complexPhaseshift = \orderedComplexPhaseshiftSelect$. Bottom: polar representations of $\maxCosUnitcircle: \unitcircle \to \interval{-1}{1}$ \eqref{eq:maxCosUnitcircle}, corresponding to the same settings. The closer the curve is from the outer ring, the more likely some points $\searchvecSelect$ will fall near a ridge of $\cosfunInpimg$.
			(a) Case where the values $\complexPhaseshift$ are roughly evenly distributed on $\unitcircle$.
			(b) Case where these values are concentrated in a small portion of the unit circle.
			The most extreme cases occurs when $\complexPhaseshift = 1$ for any $\vectorindexbis$. \Cref{fig:discretegrid} (middle and right) depicts two such situations.
		}
	\label{fig:phaserepartition}
	\end{figure}
	According to \cref{hyp:uniformdist}, $\stochComplexPhaseInterp(\spatialvec)$ follows a uniform distribution on $\unitcircle$. Therefore,
	\begin{equation}
		\Expval\left[\stochCosmaxInterp(\spatialvec)^\normdegree\right] = \frac1{2\pi} \int_{\unitcircle} \maxCosUnitcircle(\pointunitcircle)^\normdegree \, \rmd\angularMeasure(\pointunitcircle),
	\end{equation}
	which proves that $\ExpvalCosmaxInterp{\normdegree}$ does not depend on $\spatialvec$. Let us split the unit circle $\unitcircle$ into the arcs $\orderedArc{0},\, \dots,\, \orderedArc{\nevalpointsGrid - 1}$ such as introduced in \eqref{eq:arcunitcircle}:
	\begin{equation}
		\ExpvalCosmaxInterp{\normdegree} = \frac1{2\pi} \sumoverEvalpointsGrid \int_{\arcSelect} \maxCosUnitcircle(\pointunitcircle)^\normdegree \, \rmd\angularMeasure(\pointunitcircle).
	\label{eq:moments_maxcos}
	\end{equation}
	Let $\selectEvalpoint \in \setof{\nevalpointsGrid}$. We show that
	\begin{equation}
		\forall \pointunitcircle \in \arcSelect,\, \maxCosUnitcircle(\pointunitcircle) = \max\pair{\Real\bigl(
			\pointunitcircle^\ast \orderedComplexPhaseshiftSelect
		\bigr)}{\Real\bigl(
			\pointunitcircle^\ast \orderedComplexPhaseshiftSelectNext
		\bigr)}.
	\label{eq:maxcosunitcircle_1}
	\end{equation}
	Let $\pointunitcircle \in \arcSelect$ and $\selectEvalpoint' \notin \tuple{\selectEvalpoint}{\selectEvalpoint + 1}$. We prove that
	\begin{equation}
		\Real\bigl(
			\pointunitcircle^\ast \orderedComplexPhaseshiftSelectOther
		\bigr) \leq \Real\bigl(
			\pointunitcircle^\ast \orderedComplexPhaseshiftSelect
		\bigr) \qqor
		\Real\bigl(
			\pointunitcircle^\ast \orderedComplexPhaseshiftSelectOther
		\bigr) \leq \Real\bigl(
			\pointunitcircle^\ast \orderedComplexPhaseshiftSelectNext
		\bigr).
	\label{eq:maxcosunitcircle_2}
	\end{equation}
	On the one hand, we assume that $\angle\bigl(
		\pointunitcircle^\ast \orderedComplexPhaseshiftSelectOther
	\bigr) \leq \pi$. By design of $\setofOrderedComplexPhaseshift$, we have
	\begin{equation}
		\orderedComplexPhaseshiftSelectNext \in \bigarc{\pointunitcircle}{\orderedComplexPhaseshiftSelectOther}.
	\end{equation}
	Therefore, by definition of arcs on the unit circle \eqref{eq:arc}, we get
	\begin{equation}
		\angle\bigl(
			\pointunitcircle^\ast \orderedComplexPhaseshiftSelectNext
		\bigr)
		\leq
		\angle\bigl(
			\pointunitcircle^\ast \orderedComplexPhaseshiftSelectOther
		\bigr).
	\end{equation}
	Then, since $\cos$ is non-increasing on $\zeropi$, we get
	\begin{equation}
		\cos\angle\bigl(
			\pointunitcircle^\ast \orderedComplexPhaseshiftSelectNext
		\bigr)
		\geq
		\cos\angle\bigl(
			\pointunitcircle^\ast \orderedComplexPhaseshiftSelectOther
		\bigr),
	\end{equation}
	which yields the right part of \eqref{eq:maxcosunitcircle_2}. On the other hand, if $\angle\bigl(
		\pointunitcircle^\ast \orderedComplexPhaseshiftSelectOther
	\bigr) \geq \pi$, a similar reasoning yields the left part of \eqref{eq:maxcosunitcircle_2}. Then, \eqref{eq:maxcosunitcircle_1} holds.

	Next, we show that, as observed in \cref{fig:phaserepartition}, $\maxCosUnitcircle$ is piecewise-symmetric with respect to the center value of each arc $\arcSelect$, denoted by
	\begin{equation}
		\halfOrderedComplexPhaseshiftSelect := \sqrt{\orderedComplexPhaseshiftSelect \orderedComplexPhaseshiftSelectNext}.
	\end{equation}
	Let $\pointunitcircle_1,\, \pointunitcircle_2 \in \arcSelect$ which are symmetric with respect to $\halfOrderedComplexPhaseshiftSelect$. Therefore, there exists $\pointunitcircle' \in \unitcircle$ such that $\pointunitcircle_1 = \halfOrderedComplexPhaseshiftSelect \pointunitcircle'$ and $\pointunitcircle_2 = \halfOrderedComplexPhaseshiftSelect {\pointunitcircle'}^\ast$. We now prove that
	\begin{equation}
		\maxCosUnitcircle(\pointunitcircle_1) = \maxCosUnitcircle(\pointunitcircle_2).
	\label{eq:symmetryMaxcosunitcircle_1}
	\end{equation}
	A simple calculation yields
	\begin{equation}
		\pointunitcircle_1^\ast \orderedComplexPhaseshiftSelectNext = {\pointunitcircle'}^\ast \centeredHalfOrderedComplexPhaseshiftSelect
		\qqand
		\pointunitcircle_2^\ast \orderedComplexPhaseshiftSelect = \bigl(
			{\pointunitcircle'}^\ast \centeredHalfOrderedComplexPhaseshiftSelect
		\bigr)^\ast,
	\end{equation}
	with
	\begin{equation}
		\centeredHalfOrderedComplexPhaseshiftSelect
			:= \bigl(
				{\orderedComplexPhaseshiftSelect}^\ast \, \halfOrderedComplexPhaseshiftSelect
			\bigr) = \bigl(
				{\halfOrderedComplexPhaseshiftSelect}^\ast \orderedComplexPhaseshiftSelectNext
			\bigr).
	\end{equation}
	Therefore,
	\begin{equation}
		\Real\bigl(
			\pointunitcircle_1^\ast \orderedComplexPhaseshiftSelectNext
		\bigr)
		=
		\Real\bigl(
			\pointunitcircle_2^\ast \orderedComplexPhaseshiftSelect
		\bigr).
	\label{eq:symmetryMaxcosunitcircle_2}
	\end{equation}
	Since $\pointunitcircle_1,\, \pointunitcircle_2$ both belong to $\arcSelect$, $\maxCosUnitcircle(\pointunitcircle_1)$ and $\maxCosUnitcircle(\pointunitcircle_2)$ satisfy \eqref{eq:maxcosunitcircle_1}. Then, by symmetry, \eqref{eq:symmetryMaxcosunitcircle_2} implies \eqref{eq:symmetryMaxcosunitcircle_1}.
	One can observe from \cref{fig:phaserepartition} that $\maxCosUnitcircle$ reaches its local minimum at the center of arc $\arcSelect$, \ie, $\halfOrderedComplexPhaseshiftSelect$. This corresponds to a point where $\maxCosUnitcircle$ is non-differentiable.

	We denote by $\halfarcSelect := \bigarc{\orderedComplexPhaseshift{\selectEvalpoint}}{\halfOrderedComplexPhaseshiftSelect}$ the first half of arc $\arcSelect$. Then,
	\begin{equation}
		\forall \pointunitcircle \in \halfarcSelect,\, \maxCosUnitcircle(\pointunitcircle) = \Real\bigl(
			\pointunitcircle^\ast \orderedComplexPhaseshift{\selectEvalpoint}
		\bigr).
	\end{equation}
	As a consequence, using symmetry, we get
	\begin{align*}
		\int_{\arcSelect} \maxCosUnitcircle(\pointunitcircle)^\normdegree \,\rmd\angularMeasure(\pointunitcircle)
			&= 2 \int_{\halfarcSelect} \maxCosUnitcircle(\pointunitcircle)^\normdegree \,\rmd\angularMeasure(\pointunitcircle) \\
			&= 2 \int_{\halfarcSelect} \Real\bigl(
				\pointunitcircle^\ast \orderedComplexPhaseshift{\selectEvalpoint}
			\bigr)^\normdegree \,\rmd\angularMeasure(\pointunitcircle).
	\end{align*}
	By using the change of variable formula \citep[p.~81]{Athreya2006} with $\pointunitcircle \leftarrow \rme^{i\angleval}$, we get
	\begin{equation}
		\int_{\arcSelect} \maxCosUnitcircle(\pointunitcircle)^\normdegree \,\rmd\angularMeasure(\pointunitcircle) = 2 \int_{\orderedPhase{\selectEvalpoint}}^{\halfOrderedPhase{\selectEvalpoint}} \cos^\normdegree\bigl(
			\angleval - \orderedPhase{\selectEvalpoint}
		\bigr) \,\rmd\angleval,
	\end{equation}
	where $\halfOrderedPhase{\selectEvalpoint} := \bigl(
		\orderedPhase{\selectEvalpoint} + \orderedPhase{\selectEvalpoint+1}
	\bigr) / 2$ denotes the argument of $\halfOrderedComplexPhaseshiftSelect$. Then, the change of variable $\angleval' \leftarrow \angleval - \orderedPhase{\selectEvalpoint}$ yields
	\begin{equation}
		\int_{\arcSelect} \maxCosUnitcircle(\pointunitcircle)^\normdegree \,\rmd\angularMeasure(\pointunitcircle) = 2 \int_0^{\diffOrderedPhaseSelect / 2} \cos^\normdegree \angleval' \,\rmd\angleval'.
	\label{eq:integralonarc_unitcircle}
	\end{equation}

	Next, we insert \eqref{eq:integralonarc_unitcircle} into \eqref{eq:moments_maxcos}, and compute $\Expval\left[\stochCosmaxInterp(\spatialvec)^\normdegree\right]$ for $\normdegree \leftarrow 1$ and $\normdegree \leftarrow 2$:
	\begin{align*}
		\ExpvalCosmaxInterp{}
			&= \frac1{\pi} \sumoverEvalpointsGrid \sin\frac{\diffOrderedPhaseSelect}{2}; \\
		\ExpvalCosmaxInterp{2}
			&= \frac12 + \frac1{4\pi} \sumoverEvalpointsGrid \sin\diffOrderedPhaseSelect.
	\end{align*}
	We recall that $\stochOneminusCosmaxInterp := 1 - \stochCosmaxInterp$. By linearity of $\Expval$, we get
	\begin{equation}
		\Expval\left[\stochOneminusCosmaxInterp(\spatialvec)^2\right] := \frac32 + \frac1{4\pi} \sumoverEvalpointsGrid \left(\sin\diffOrderedPhaseSelect - 8 \sin\frac{\diffOrderedPhaseSelect}{2}\right),
	\end{equation}
	which concludes the proof.
\end{proof}

\begin{figure}
	\centering
	\includegraphics[height=0.37\textwidth]{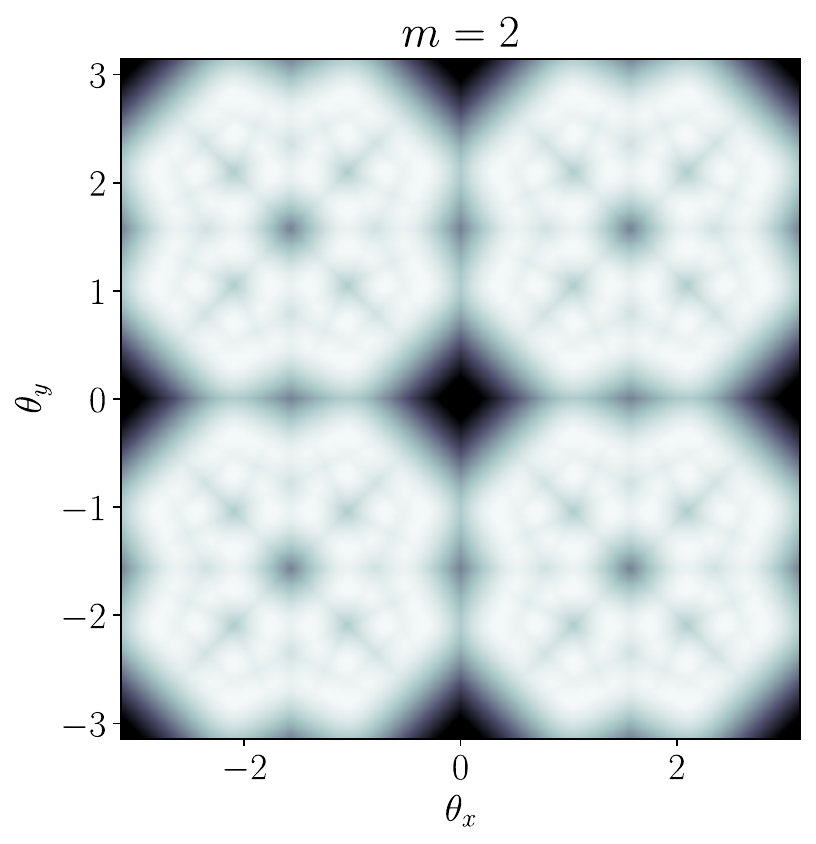}
	\hspace{20pt}
	\includegraphics[height=0.37\textwidth]{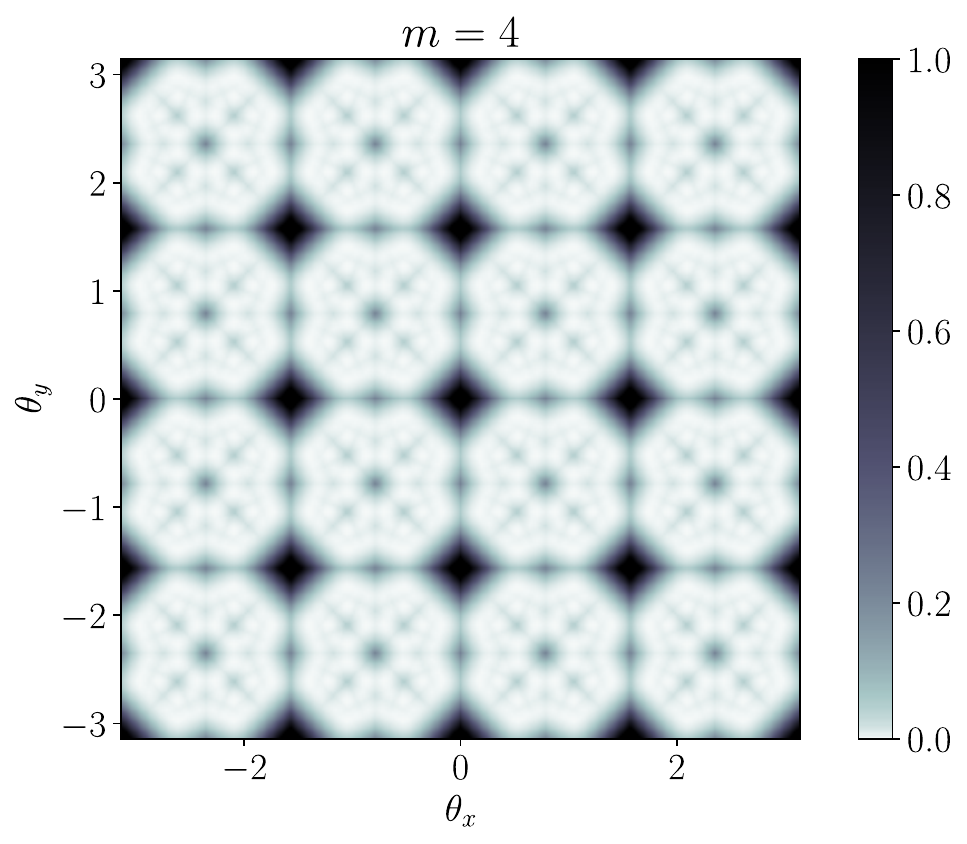}
	\vspace{-10pt}
	\caption[
		Expected Discrepancy between \rmax and \cmod
	]{
		$\proxyfunRmaxCmod(\sub\freqvecMpipi)^2$ as a function of the kernel characteristic frequency $\freqvecMpipi \in \mpipi^2$. According to \cref{th:scdmoment_normdiff_cmodrmax}, this quantity provides an approximate bound for the expected quadratic error between \rmax and \cmod outputs. The subsampling factor $\sub$ has been set to $2$ as in ResNet (left), and $4$ as in AlexNet (right). The bright regions correspond to frequencies for which the two outputs are expected to be similar. However, in the dark regions, pathological cases such as illustrated in \cref{fig:discretegrid} are more likely to occur.
	}
	\label{fig:proxyfunction}
\end{figure}

We consider an ideal scenario where $\setofEvalOrderedComplexPhaseshift$ are evenly spaced on $\unitcircle$. Then, an order $2$ Taylor expansion yields
\begin{equation}
	\proxyfunRmaxCmod_\gridhalfsize(\sub\freqvecMpipi) = \smallO(1/\gridhalfsize^2),
\label{eq:decay_proxyfunRmaxCmod}
\end{equation}
providing an order-two-polynomial decay rate for $\stochOneminusCosmaxInterp(\spatialvec)$, when the grid half-size $\gridhalfsize$ increases. \Cref{fig:proxyfunction} displays $\freqvecMpipi \mapsto \proxyfunRmaxCmod_\gridhalfsize(\sub\freqvecMpipi)^2$ for $\freqvecMpipi \in \mpipi^2$, with $\sub = 4$ and $\gridhalfsize = 1$ as in AlexNet. We notice that, for the major part of the Fourier domain, $\proxyfunRmaxCmod_\gridhalfsize$ remains close to $0$. However, we observe a regular pattern of dark regions, which correspond to pathological frequencies where the repartition of $\setofEvalOrderedComplexPhaseshift$ is unbalanced.

So far, we established a result at the pixel level. Before stating \cref{th:scdmoment_normdiff_cmodrmax}, which extends the result to the image level, we need the following intermediate statement.

\begin{proposition}
	\label{prop:indep}
	We consider the random variable
	\begin{equation}
		\normtwoStochMagnitudeInpimg := \bignormtwo{\detailCmodimg}.
	\label{eq:norm_outimg_cmod}
	\end{equation}
	Under \cref{hyp:indep_phase_modulus}, for any $\spatialvec \in \mathR^2$,
	\begin{itemize}
		\item $\stochComplexPhaseInterp(\spatialvec)$ is independent of $\normtwoStochMagnitudeInpimg$;
		\item $\stochComplexPhaseInterp(\spatialvec)$, $\stochMagnitudeInterp(\spatialvec)$ are conditionally independent given $\normtwoStochMagnitudeInpimg$.
	\end{itemize}
\end{proposition}

\begin{proof}
	See \cref{subsec:appendix_proofs_prop_indep}.
\end{proof}

Finally, \cref{prop:expval_oneminuscos,prop:indep} yield the following theorem. It provides an upper bound on the expected value of the normalized mean squared error $\avgStochDiffRmaxCmodInpimg^2$, such as defined in \eqref{eq:normdiff_cmodrmax_stoch}.

\begin{theorem}[MSE between \cmod and \rmax]
\label{th:scdmoment_normdiff_cmodrmax}
	Let $\complexWeightimg \in \DiscreteGaborfilterGen$ denote a discrete Gabor-like filter, $\sub \in \nonzeroMathN$ a subsampling factor and $\gridhalfsize \in \nonzeroMathN$ a grid half-size. We consider a stochastic process $\inpimg$ whose realizations are elements of $\realltwoZsq$. We assume that condition \eqref{eq:condition_bandwidth} is satisfied: $\supportsizeDiscrete \leq \pi / \sub$. Then, under \cref{hyp:bound_sumofdiffs,hyp:uniformdist,hyp:indep_phase_modulus},%
	\footnote{
		We can easily prove that these properties are independent from the choice of sampling interval $\samplinterv > 0$.
	}
	\begin{equation}
		\ExpvalDiffRmaxCmod \leq \bigl(
			\evalLowerboundDiffRmaxCmod + \evalProxyfunRmaxCmod
		\bigr)^2,
	\label{eq:scdmoment_normdiff_cmodrmax}
	\end{equation}
	where $\avgStochDiffRmaxCmodInpimg^2$ \eqref{eq:normdiff_cmodrmax_stoch} denotes the stochastic quadratic error between \cmod and \rmax outputs. We remind that $\lowerboundDiffRmaxCmod_\gridhalfsize$ and $\proxyfunRmaxCmod_\gridhalfsize$ have been introduced in \eqref{eq:diffmodpool_multfactor} and \eqref{eq:proxyfunction}, respectively.
\end{theorem}

\begin{proof}
	See \cref{subsec:appendix_proofs_th_scdmoment_normdiff_cmodrmax}.
\end{proof}

Let us analyze the bound obtained in \eqref{eq:scdmoment_normdiff_cmodrmax}. The first term, $\evalLowerboundDiffRmaxCmod$, accounts for the localized property of the convolution filter $\complexWeightimg$. This term decreases linearly with the product $\sub\supportsizeDiscrete$. In the limit case where $\supportsizeDiscrete = 0$ (infinite, nonlocal filter), we get $\evalLowerboundDiffRmaxCmod = 0$. Note that a smaller subsampling factor $\sub$ allows for a larger bandwidth $\supportsizeDiscrete$. Moreover, $\evalLowerboundDiffRmaxCmod$ increases linearly with the size of the max pooling grid, which is characterized by $\gridhalfsize$.
The second term, $\evalProxyfunRmaxCmod$, accounts for the discrete nature of the max pooling grid. It strongly depends on the characteristic frequency $\freqvecMpipi$, as illustrated in \cref{fig:proxyfunction}. According to \eqref{eq:decay_proxyfunRmaxCmod}, this term has a polynomial decay when $\gridhalfsize$ increases. However, increasing the size of the max pooling grid also results in increasing the term $\evalLowerboundDiffRmaxCmod$, as explained above. Therefore, a tradeoff must be found to get an optimal bound.

\section{Shift Invariance of \rmax Outputs}
\label{sec:invariance_rmax}

In this section, we present the main theoretical claim of this paper. Based on the previous results, we provide a probabilistic measure of shift invariance for \rmax operators. First, we state the following lemma.

\begin{lemma}
\label{lemma:hyps_tranls}
	If \cref{hyp:uniformdist,hyp:indep_phase_modulus} are satisfied, then they are also true with $\inpimg \leftarrow \translInpimg$, for any $\translvecDiscrete \in \mathR^2$.
\end{lemma}

\begin{proof}
	See \cref{subsec:appendix_proofs_lemma_hyps_tranls}.
\end{proof}

We are now ready to state the main result about shift invariance of \rmax outputs.

\begin{theorem}[Shift invariance of \rmax]
	\label{th:shiftinvariance_rmax}
	We assume that the requirements stated in \cref{th:scdmoment_normdiff_cmodrmax} are satisfied. Additionally, given a translation vector $\translvecDiscrete \in \mathR^2$, we consider the following random variable:
	\begin{equation}
		\avgStochDiffShiftRmaxInpimg := \bignormtwo{\rmaxopSubGrid(\translInpimg) - \detailRmaximg} / \bignormtwo{\detailCmodimg}.
	\label{eq:normdiff_rmax_transl_stoch}
	\end{equation}
	Then, under condition \eqref{eq:condition_bandwidth}, we have
	\begin{equation}
		\ExpvalDiffShiftRmax \leq 2 \, \bigl(
			\evalLowerboundDiffRmaxCmod + \evalProxyfunRmaxCmod
		\bigr) + \evalShiftinvCmod,
	\label{eq:expval_normdiff_rmax_transl}
	\end{equation}
	where $\shiftinvCmod$, $\lowerboundDiffRmaxCmod_\gridhalfsize$ and $\proxyfunRmaxCmod_\gridhalfsize$ are defined in \eqref{eq:shiftinvariance_continuous_multfac}, \eqref{eq:diffmodpool_multfactor} and \eqref{eq:proxyfunction}, respectively.
\end{theorem}

\begin{proof}
	See \cref{subsec:appendix_proofs_th_shiftinvariance_rmax}.
\end{proof}

In the bound established in \eqref{eq:expval_normdiff_rmax_transl}, the sum $\evalLowerboundDiffRmaxCmod + \evalProxyfunRmaxCmod$ accounts for the discrepancy between \rmax and \cmod outputs, as stated in \cref{th:scdmoment_normdiff_cmodrmax}, whereas the term $\evalShiftinvCmod$ characterizes the stability of \cmod outputs, as stated in \cref{th:shiftinvariance_cmod}.
If $\supportsizeDiscrete$ is sufficiently small, then $\evalShiftinvCmod$ and $\evalLowerboundDiffRmaxCmod$ become negligible with respect to $\evalProxyfunRmaxCmod$, and the bound can be approximated by $2 \, \evalProxyfunRmaxCmod$. \Cref{th:shiftinvariance_rmax} therefore provides a validity domain for shift invariance of \rmax operators, as illustrated in \cref{fig:proxyfunction} with $\gridhalfsize = 1$.

\begin{remark}
	\label{remark:normalization}
	The stochastic discrepancy introduced in \eqref{eq:normdiff_rmax_transl_stoch} is estimated relatively to the \cmod output. This choice is motivated by the perfect shift invariance of its norm, as shown in \cref{prop:shiftinvariance_normcmod}.
\end{remark}

\begin{remark}
	\label{remark:gridhalfsize}
	In practice, most of the time max pooling is performed on a grid of size $3 \times 3$; therefore $\gridhalfsize = 1$. For the sake of conciseness, we shall sometimes drop $\gridhalfsize$ in the notations, which implicitly means $\gridhalfsize = 1$.
\end{remark}

\section{Adaptation to Multichannel Convolution Operators}
\label{sec:multichannel}

In this section, we adapt \cref{th:shiftinvariance_cmod,th:scdmoment_normdiff_cmodrmax,th:shiftinvariance_rmax} to multichannel inputs (\eg, RGB images), employed in conventional CNNs such as AlexNet or ResNet.

First, we define multichannel \rmax and \cmod operators relatively to \eqref{eq:rmax} and \eqref{eq:cmod}.
We denote by $\inchannels$ and $\outchannels \in \nonzeroMathN$ the number of input and output channels, respectively. Additionally, we consider a \emph{multichannel convolution tensor}
\begin{equation}
	\complexWeightmultimg := (\complexWeightimgSelect)_{\selectOutchannelInrange,\, \selectInchannelInrange} \in \complexltwoZsqpower{\outchannels \times \inchannels}.
\end{equation}
Multichannel \rmax and \cmod operators take as input images, denoted by
\begin{equation}
	\inpmultimg := (\inpimgSelect)_{\selectInchannelInrange} \in \realltwoZsqpower{\inchannels}.
\label{eq:inpmultimg}
\end{equation}
They are defined, for any given output channel $\selectOutchannelInrange$, by
\begin{align}
	\rmaxopSubGridSelectWeight : \inpmultimg
	&\mapsto \maxpool_\gridhalfsize \left(
		\sumoverInchannels \left(
			\inpimgSelect \ast \Real\flippedComplexWeightimgSelect
		\right) \downarrow \sub
	\right);
\label{eq:multrmax} \\
	\cmodopSubSelectWeight: \inpmultimg
		&\mapsto \left|
			\sumoverInchannels(\inpimgSelect \ast \flippedComplexWeightimgSelect) \downarrow (2\sub)
		\right|,
\label{eq:multcmod}
\end{align}
where $\sub,\, \gridhalfsize \in \nonzeroMathN$ respectively denote a subsampling factor and the max pooling grid half-size.
Analogously to \eqref{eq:outimg_rmax_cmod} for single-channel inputs, we now consider
\begin{equation}
	\rmaximgSelect := \rmaxopSubGridSelectWeight(\inpmultimg) \qqand \cmodimgSelect := \cmodopSubSelectWeight(\inpmultimg).
\label{eq:outimg_multrmax_multcmod}
\end{equation}
Again, in what follows we omit the parameter between square brackets.
To apply \cref{th:shiftinvariance_cmod,th:scdmoment_normdiff_cmodrmax,th:shiftinvariance_rmax} to the current setting on the $\selectOutchannel$-th output channel, we need the following hypotheses.

\begin{hypothesis}[Monochorome filters]
\label{hyp:monochrome}
	Let
	\begin{equation}
		\avgComplexWeightimgSelect := \frac{1}{\inchannels} \sumoverInchannels \complexWeightimgSelect
	\label{eq:avgWeightimgSelect}
	\end{equation}
	denote the mean kernel of the $\selectOutchannel$-th output channel. Then, there exists $\colormixvec_\selectOutchannel \in \mathR^\inchannels$ such that
	\begin{equation}
		\forall \selectInchannel \in \setof{\inchannels},\, \complexWeightimgSelect = \colormixvalSelectOutIn \avgComplexWeightimgSelect.
	\label{eq:monochrome}
	\end{equation}
\end{hypothesis}

\begin{hypothesis}[Gabor-like filters]
	\label{hyp:cnngaborfilter}
	There exists a bandwidth $\supportsizeDiscrete > 0$ satisfying $\supportsizeDiscrete \leq \pi / \sub$ and a frequency vector $\freqvecMpipiSelect \in \mpipi^2$ such that
	\begin{equation}
		\avgComplexWeightimg_\selectOutchannel \in \DiscreteGaborfilter{\freqvecMpipiSelect}{\supportsizeDiscrete}.
	\label{eq:sharedfourierwindowsize}
	\end{equation}
\end{hypothesis}

Note that the bandwidth $\supportsizeDiscrete$ is not indexed by $\selectOutchannel$, because it shall later be assumed to be shared across the output channels.
Then, under \cref{hyp:monochrome}, $\rmaximgSelect$ and $\cmodimgSelect$ are the outputs of single-channel \rmax and \cmod operators, as introduced in \eqref{eq:rmax} and \eqref{eq:cmod}:
\begin{equation}
	\rmaximgSelect = \rmaxopSubGrid\bigl[
		\avgComplexWeightimg_{\selectOutchannel}
	\bigr]\bigl(
		\colormiximgSelect
	\bigr)
	\qqand
	\cmodimgSelect = \cmodopSub\bigl[
		\avgComplexWeightimg_{\selectOutchannel}
	\bigr]\bigl(
		\colormiximgSelect
	\bigr),
\label{eq:outimg_multrmax_multcmod_2}
\end{equation}
where $\colormiximgSelect \in \realltwoZsq$ (``luminance'' image) is defined as the following linear combination:
\begin{equation}
	\colormiximgSelect := \sumoverInchannels \colormixvalSelectOutIn \inpimgSelect.
\label{eq:colormiximgSelect}
\end{equation}
The results established for single-channel inputs can therefore be extended to multichannel operators. Specifically, we get the following corollaries to \cref{th:shiftinvariance_cmod,th:scdmoment_normdiff_cmodrmax,th:shiftinvariance_rmax}.

\begin{corollary}[Shift invariance of \cmod]
\label{cor:shiftinvariance_multcmod}
	For a given output channel $\selectOutchannelInrange$, we postulate \cref{hyp:monochrome,hyp:cnngaborfilter}. Then, for any input image $\inpmultimg \in \realltwoZsqpower{\inchannels}$ with finite support and any translation vector $\translvecDiscrete \in \mathR^2$,
	\begin{equation}
		\bignormtwo{\cmodopSubSelect (\translInpmultimg) - \cmodopSubSelect\inpmultimg} \leq \evalShiftinvCmod \, \bignormtwo{\cmodopSubSelect \inpmultimg},
	\label{eq:shiftinvariance_multcmod}
	\end{equation}
	where $\shiftinvCmod$ has been defined in \eqref{eq:shiftinvariance_continuous_multfac}.
\end{corollary}

\begin{corollary}[MSE between \cmod and \rmax]
\label{cor:scdmoment_normdiff_multcmodrmax}
	As in \cref{cor:shiftinvariance_multcmod}, we postulate \cref{hyp:monochrome,hyp:cnngaborfilter}. Again, we assume that condition \eqref{eq:condition_bandwidth} is satisfied: $\supportsizeDiscrete \leq \pi / \sub$. Additionally, we consider $\inpmultimg$ as a stack of $\inchannels$ discrete stochastic processes, and assume \cref{hyp:bound_sumofdiffs,hyp:uniformdist,hyp:indep_phase_modulus} with $\inpimg \leftarrow \colormiximgSelect$ and $\complexWeightimg \leftarrow \avgComplexWeightimgSelect$. Then,
	\begin{equation}
		\ExpvalDiffMultRmaxCmod \leq \bigl(
			\evalLowerboundDiffRmaxCmod + \evalProxyfunRmaxCmodSelect
		\bigr)^2,
	\label{eq:scdmoment_normdiff_multcmodrmax}
	\end{equation}
	where we have defined the following random variable:
	\begin{equation}
		\avgStochDiffMultRmaxCmod
			:= \bignormtwo{\cmodopSubSelect\inpmultimg - \rmaxopSubSelect\inpmultimg} / \bignormtwo{\cmodopSubSelect\inpmultimg}.
	\end{equation}
\end{corollary}

\begin{corollary}[Shift invariance of \rmax]
	\label{cor:shiftinvariance_multrmax}
	We assume that the requirements stated in \cref{cor:scdmoment_normdiff_multcmodrmax} are satisfied.
	Then, for any translation vector $\translvecDiscrete \in \mathR^2$,
	\begin{equation}
		\ExpvalDiffShiftMultRmax \leq 2 \, \bigl(
			\evalLowerboundDiffRmaxCmod + \evalProxyfunRmaxCmodSelect
		\bigr) + \evalShiftinvCmod,
	\label{eq:expval_normdiff_multrmax_transl}
	\end{equation}
	where we have defined the following random variable:
	\begin{equation}
		\avgStochDiffShiftMultRmax
			:= \bignormtwo{\rmaxopSubSelect(\translInpmultimg) - \rmaxopSubSelect\inpmultimg} / \bignormtwo{\cmodopSubSelect\inpmultimg}.
	\end{equation}
\end{corollary}

\begin{remark}
	In the above results, we used a translation operator on multichannel tensors, obtained by applying $\transl_{\translvecDiscrete}$, as defined in \eqref{eq:transl}, to each channel $\inpimgSelect$.
\end{remark}


\section[A Case Study Implementing \dtcwpt]{A Case Study Implementing the Dual-Tree Complex Wavelet Packet Transform}
\label{sec:dtcwpt}

In this section, we experimentally validate the results stated in \cref{th:shiftinvariance_cmod,th:scdmoment_normdiff_cmodrmax,th:shiftinvariance_rmax}. To this end, we consider a fully-deterministic scenario implementing the dual-tree complex wavelet packet transform (\dtcwpt), which exhibit characteristics akin to those observed in the initial convolution layer of freely-trained CNNs such as AlexNet or ResNet. In particular, as stated in \cref{subsec:dtcwpt_summary}, \dtcwpt achieves subsampled convolutions with oriented band-pass filters tiling the Fourier domain into overlapping square windows. As such, it provides a convenient framework to experimentally validate our theoretical findings in a controlled environment. Then, in \cref{subsec:dtcwpt_rmaxcmod}, we build \cmod and \rmax operators based on \dtcwpt convolution kernels.

\subsection{Main Properties}
\label{subsec:dtcwpt_summary}

In what follows, we outline the principal characteristics of \dtcwpt. A detailed description of the transform itself is provided in \cref{subsec:appendix_dtcwpt_background}, whereas the results presented hereafter are formally established in \cref{subsec:appendix_dtcwpt_conv,subsec:appendix_dtcwpt_properties}.

\begin{figure}[h]
	\centering
	\begin{subfigure}{\textwidth}
		\centering
		\includegraphics[width=0.35\textwidth]{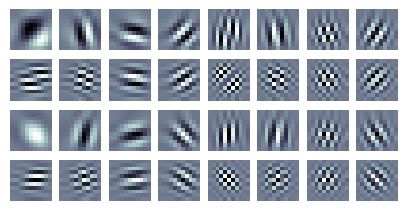}
		\caption{$\depth = 2$}
	\label{subfig:dtcwptkernels_J2}
	\end{subfigure}
	\begin{subfigure}{\textwidth}
		\centering
		\includegraphics[width=0.75\textwidth]{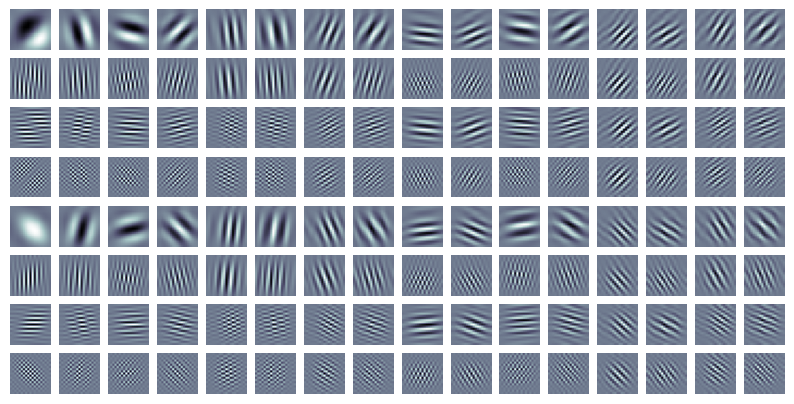}
		\caption{$\depth = 3$}
	\label{subfig:dtcwptkernels_J3}
	\end{subfigure}
	\vspace{-10pt}
	\caption[
		WPT and \dtcwpt Kernels
	]{
		Real part of the convolution kernels $\complexWeightmultimgNEMultires{\depth}$, $\complexWeightmultimgSEMultires{\depth}$, with $\depth = 2$ ($32$ filters, $\subDepth = 2$) and $\depth = 3$ ($128$ filters, $\subDepth = 4$), respectively. The kernels have been computed using Q-shift orthogonal QMFs of length $10$ \citep{Kingsbury2003}. The kernels have been respectively cropped to size $11 \times 11$ and $19 \times 19$, for the sake of legibility. Note that the filters displayed in (a) and (b) share similarities with those found in, respectively, ResNet ($\sub = 2$) and AlexNet ($\sub = 4$), after training with ImageNet.
	}
	\label{fig:dtcwptkernels}
\end{figure}

For a given decomposition depth $\depth \in \nonzeroMathN$, \dtcwpt achieves subsampled convolutions with $4 \times 4^\depth$ oriented band-pass filters that tile the Fourier domain into overlapping square windows of size
\begin{equation}
	\supportsizeDiscreteDepth := \pi / \subDepth, \qqwith \subDepth := 2^{\depth-1}.
\label{eq:bandwidth_dtcwpt}
\end{equation}
More specifically, considering an input image $\inpimg \in \realltwoZsq$, it produces a set of $4 \times 4^\depth$ output feature maps
\begin{equation}
	\iter{\dtmultimg}{\depth} := \bigl(
		\wptimgNEMultires{\depth}{\selectOutchannel},\, \wptimgSEMultires{\depth}{\selectOutchannel},\, \wptimgSWMultires{\depth}{\selectOutchannel},\, \wptimgNWMultires{\depth}{\selectOutchannel}
	\bigr)_{\selectOutchannelInrangeWPT},
\end{equation}
where each arrow points to the Fourier quadrant where the feature map's energy is concentrated. Moreover, as stated in \cref{prop:conv_dtcwpt}, for any $\selectOutchannelInbigrangeWPT$, there exists $\complexWeightimgNEMultires{\depth}_{\selectOutchannel} \in \complexltwoZsq$ such that
\begin{equation}
	\wptimgNEMultires{\depth}{\selectOutchannel} = \Bigl(
		\inpimg \ast {\flippedComplexWeightimgNEMultires{\depth}_{\selectOutchannel}}^\ast
	\Bigr) \downarrow 2^\depth.
\label{eq:conv_dtcwpt}
\end{equation}
An interesting property is that each kernel $\complexWeightimgNEMultires{\depth}_{\selectOutchannel}$ approximately satisfies
\begin{equation}
	\complexWeightimgNEMultires{\depth}_{\selectOutchannel} \in \DiscreteGaborfilterDt
\label{eq:dtkernel_gaborfilt_discrete}
\end{equation}
for a certain characteristic frequency $\freqvecMpipiMultires{\depth}_\selectOutchannel \in \zerotopi^2$. In other words, it approximately behaves as a Gabor-like filter in the discrete framework \eqref{eq:gaborfilt_discrete}. Moreover, each kernel corresponds to a different frequency, thereby covering the top-right quadrant of the Fourier domain. Similar results can be established for the other three Fourier quadrants. Graphical representations of $\complexWeightmultimgNEMultires{\depth} := \bigl(
	\complexWeightimgNEMultires{\depth}_{\selectOutchannel}
\bigr)_{\selectOutchannelInrangeWPT}$ and $\complexWeightmultimgSEMultires{\depth} := \bigl(
	\complexWeightimgSEMultires{\depth}_{\selectOutchannel}
\bigr)_{\selectOutchannelInrangeWPT}$ are provided in \cref{fig:dtcwptkernels} with $\depth=2$ (\cref{subfig:dtcwptkernels_J2}, $32$ filters) and $\depth=3$ (\cref{subfig:dtcwptkernels_J3}, $128$ filters).

The \rmax and \cmod operators implemented in our experiments respectively satisfy \eqref{eq:rmax} and \eqref{eq:cmod} with with $\complexWeightimg \leftarrow \complexWeightimgNEMultires{\depth}_{\selectOutchannel}$ or $\complexWeightimgSEMultires{\depth}_{\selectOutchannel}$, and $\sub \leftarrow \subDepth$.
Note that increasing the decomposition depth $\depth$, and therefore the subsampling factor $\subDepth$, results in a decreased Fourier support size $\supportsizeDiscreteDepth$, therefore matching the condition stated in \eqref{eq:condition_bandwidth} $\supportsizeDiscrete \leftarrow \supportsizeDiscreteDepth$ and $\sub \leftarrow \subDepth$.

\begin{remark}
\label{remark:only2quadrants}
	Because $\inpimg$ is real-valued, the feature maps $\wptimgSWMultires{\depth}{\selectOutchannel}$ and $\wptimgNWMultires{\depth}{\selectOutchannel}$ are the respective complex conjugates of $\wptimgNEMultires{\depth}{\selectOutchannel}$ and $\wptimgSEMultires{\depth}{\selectOutchannel}$, and thus do not need to be explicitly computed. Then, we can easily show that $\complexWeightimgSWMultires{\depth}_{\selectOutchannel}$ and $\complexWeightimgNWMultires{\depth}_{\selectOutchannel}$ are also the complex conjugates of $\complexWeightimgNEMultires{\depth}_{\selectOutchannel}$ and $\complexWeightimgSEMultires{\depth}_{\selectOutchannel}$, respectively.
\end{remark}

\subsection{\dtcwpt-Based \rmax and \cmod Operators}
\label{subsec:dtcwpt_rmaxcmod}

According to \eqref{eq:bandwidth_dtcwpt}, \eqref{eq:conv_dtcwpt}, and \eqref{eq:dtkernel_gaborfilt_discrete}, we can apply \cref{th:shiftinvariance_cmod,th:scdmoment_normdiff_cmodrmax,th:shiftinvariance_rmax} to the dual-tree framework. More precisely, for any output channel $\selectOutchannelInbigrangeWPT$, we consider the following \rmax and \cmod operators:
\begin{align}
	\rmaxopNEDepthSelect: \inpimg
		&\mapsto \maxpool\left(
			\Bigl(
				\inpimg \ast{\Real\flippedComplexWeightimgNEMultires{\depth}_{\selectOutchannel}}
			\Bigr) \downarrow 2^{\depth - 1}
		\right);
\label{eq:rmax_dt} \\
	\cmodopNEDepthSelect: \inpimg
		&\mapsto \Bigl|
			\Bigl(
				\inpimg \ast \flippedComplexWeightimgNEMultires{\depth}_{\selectOutchannel}
			\Bigr) \downarrow 2^\depth
		\Bigr|.
\label{eq:cmod_dt}
\end{align}
Using the notations introduced in \eqref{eq:cmod} and \eqref{eq:rmax}, we have
\begin{equation}
	\rmaxopNEDepthSelect = \rmaxop_{\subDepth}\bigl[
		\complexWeightimgNEMultires{\depth}_\selectOutchannel
	\bigr] \qqand
	\cmodopNEDepthSelect := \cmodop_{\subDepth}\bigl[
		\complexWeightimgNEMultires{\depth}_\selectOutchannel
	\bigr],
\label{eq:rmax_cmod_dtcwpt}
\end{equation}
where we have defined $\subDepth := 2^{\depth - 1}$. Note that, following \cref{remark:gridhalfsize}, we have omitted the grid half-size $\gridhalfsize$, which is equal to $1$ (max pooling operates on a grid of size $3 \times 3$). Furthermore, for the sake of brevity, we have omitted the depth $\depth$ in the above notations.

\begin{figure}
	\centering
	\begin{subfigure}{0.29\textwidth}
		\centering
		\includegraphics[height=0.27\textheight]{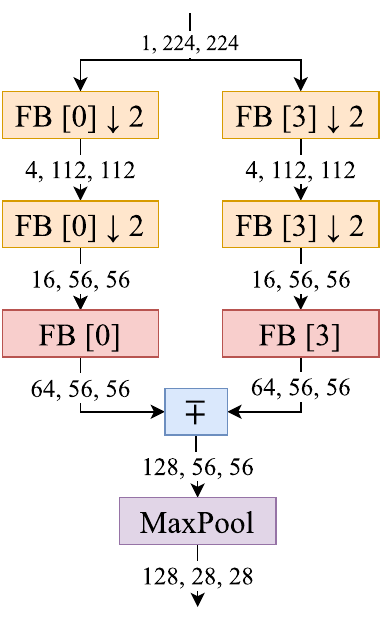}
		\caption{\rmax}
	\end{subfigure}
	\begin{subfigure}{0.7\textwidth}
		\centering
		\includegraphics[height=0.27\textheight]{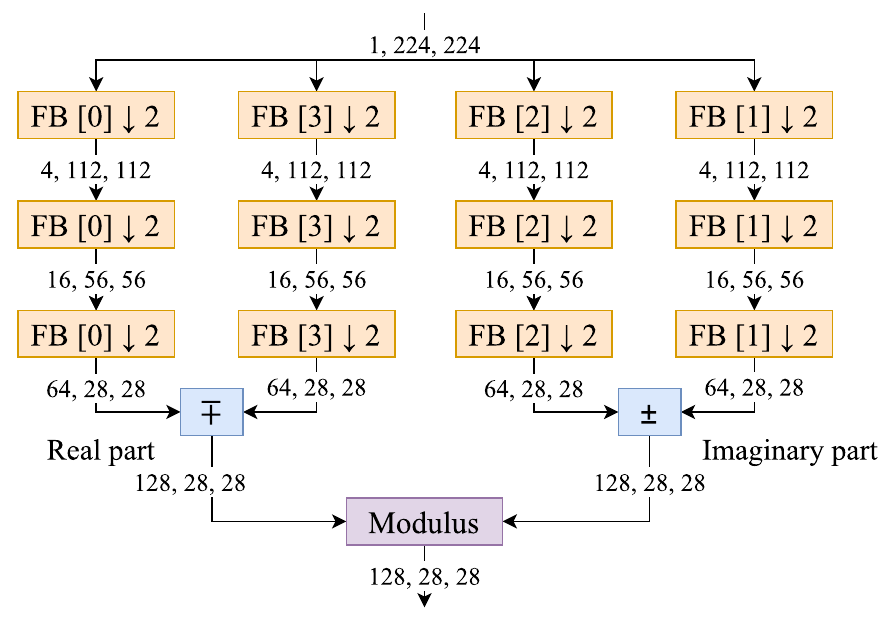}
		\caption{\cmod}
	\end{subfigure}
	\caption[
		Schematic Representations of \dtcwpt-based Operators
	]{
		Detailed illustration of the \rmax (a) and \cmod (b) operators based on \dtcwpt, with $\depth = 3$ decomposition stages. The numbers between modules correspond to the number of feature maps, height and width. The orange modules represent subsampled convolutions using one of the four 2D filter banks $\twodmultqmfFB{0-3}$, such as introduced in \eqref{eq:2dfb}. The FB index is indicated between square brackets. The \rmax model (a) only computes the real part of the dual-tree coefficients, and the last stage of decomposition is performed without subsampling (red modules). Additionally, the blue modules represent linear combinations of feature maps such as described in \eqref{eq:dtcwpt}.
	}
\label{fig:models}
\end{figure}

\begin{remark}
	\label{remark:stationary_dtcwpt}
	Both $\rmaxopNEDepthSelect$ and $\cmodopNEDepthSelect$ are implemented using \dtcwpt with $\depth$ decomposition stages. However, in \eqref{eq:rmax_dt}, the subsampling factor is equal to $2^{\depth - 1}$, instead of $2^\depth$, as stated in \eqref{eq:conv_dtcwpt}. In order to accommodate this property of \rmax operators, the last stage of \dtcwpt decomposition is carried out without subsampling, resulting in higher redundancy. This is similar to the concept of stationary wavelet transform as described by \citet{Nason1995}. Furthermore, only the real component of the wavelet feature maps is preserved. On the other hand, $\cmodopNEDepthSelect$ implements a fully-decimated wavelet packet transform, and keeps both real and imaginary parts. \Cref{fig:models} illustrates these technical details.
\end{remark}

\subsection{Experiments and Results}
\label{subsec:dtcwpt_experiments}

We implemented the \rmax and \cmod operators $\rmaxopNEDepthSelect$ and $\cmodopNEDepthSelect$, as introduced in \eqref{eq:rmax_dt} and \eqref{eq:cmod_dt}, with both $\depth = 2$ and $3$ stages of wavelet packet decomposition. To cover the whole frequency plane, we also implemented similar operators, denoted by $\rmaxopSEDepthSelect$ and $\cmodopSEDepthSelect$. They are associated with the convolution filters $\complexWeightimgSEMultires{\depth}_{\selectOutchannel}$, introduced in \cref{prop:conv_dtcwpt}, with energy being located in the bottom-right quadrant. However, as explained in \cref{remark:only2quadrants}, we did not need to deal with the two other quadrants (negative \xcoord-values).
Using the validation set of ImageNet-1K \citep{Russakovsky2015}, ($\nexamples := 50\, 000$ images), we measured the mean discrepancy between \rmax and \cmod outputs, and evaluated the shift invariance of both models. Dual-tree decompositions have been performed with Q-shift orthogonal filters of length $10$ \citep{Kingsbury2003}.

\begin{figure}
	\centering
	\includegraphics[height=0.32\textwidth]{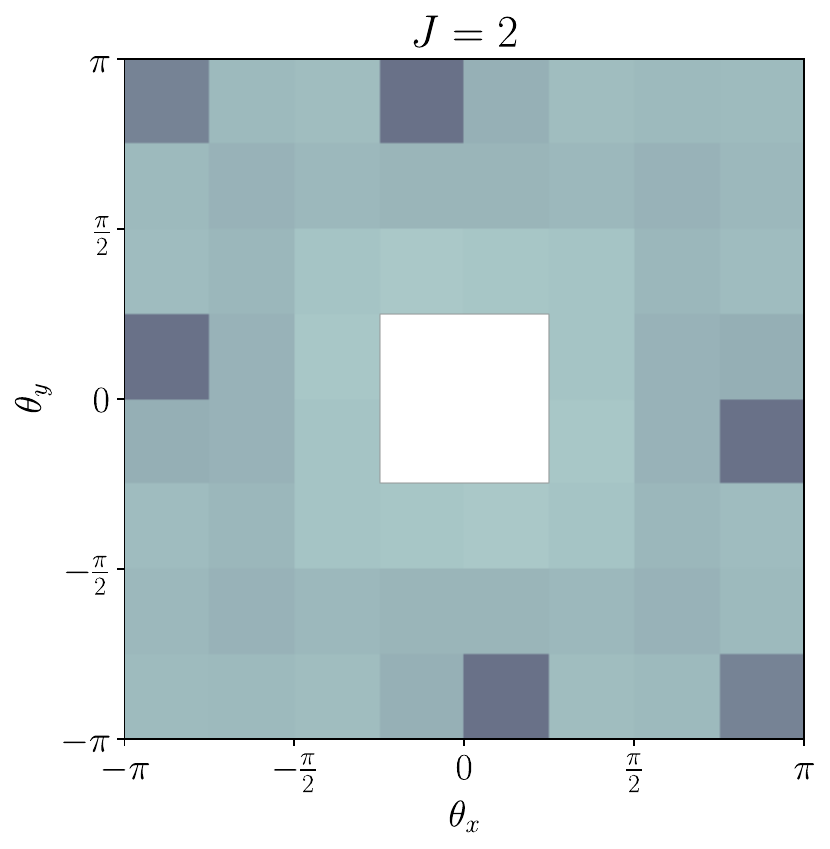}
	\hspace{20pt}
	\includegraphics[height=0.32\textwidth]{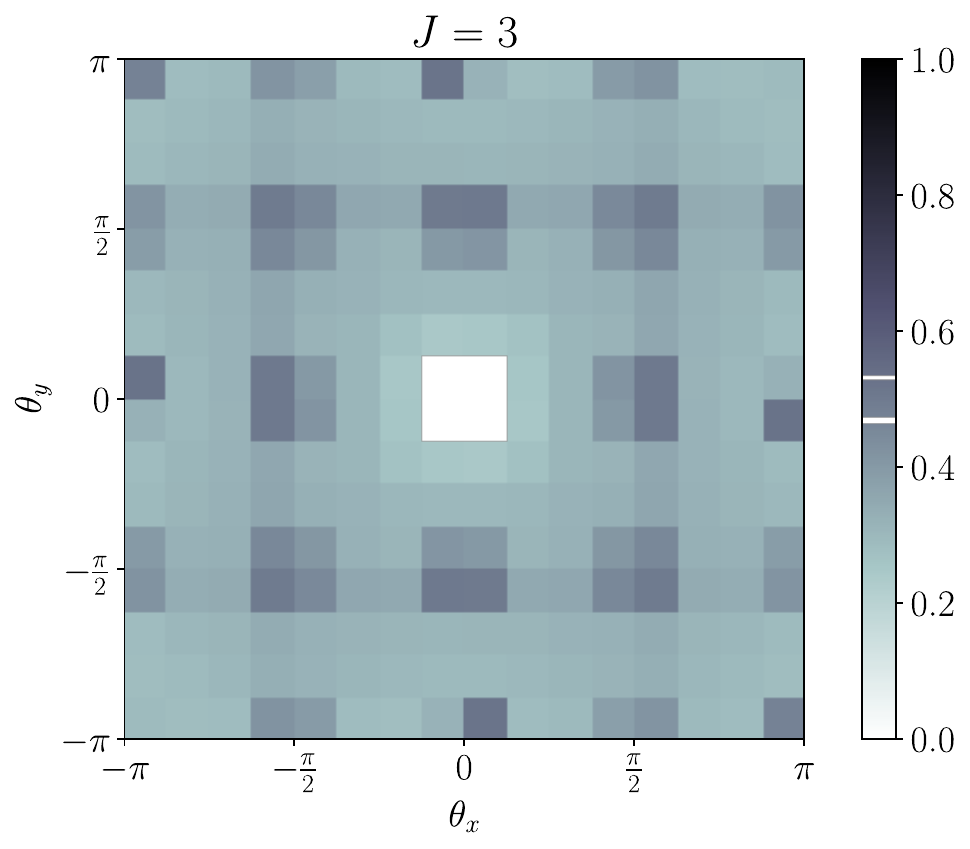}
	\vspace{-10pt}
	\caption[
		Experiments: Discrepancies between \rmax and \cmod
	]{Empirical estimates of the normalized mean squared error between \rmax and \cmod outputs, computed on ImageNet-1K (validation set). For each channel $\selectOutchannelInbigrangeWPT$, $\avgEvalDiffRmaxCmodNESelectSquared$ is plotted as a grayscale pixel centered in $\freqvecMpipiMultires{\depth}_\selectOutchannel$ such as introduced in \cref{eq:dtkernel_gaborfilt_discrete} (top-right quadrant). Similarly, $\avgEvalDiffRmaxCmodSESelectSquared$ is plotted in the bottom-right quadrant. Finally, the bottom- and top-left quadrants ($\avgEvalDiffRmaxCmodSWSelectSquared$ and $\avgEvalDiffRmaxCmodNWSelectSquared$) are simply obtained by symmetrizing the figures. Since the subsampling factor $\subDepth$ is equal to $2^{\depth-1}$, these experimental results can be compared with the left and right parts of \cref{fig:proxyfunction}. Note that the low-pass filters have been discarded because they are outside the scope of this study.}
\label{fig:scdmoment_normdiff_cmodrmax}
	\vspace{4pt}
	\begin{subfigure}{\textwidth}
		\centering
		\includegraphics[height=0.32\textwidth]{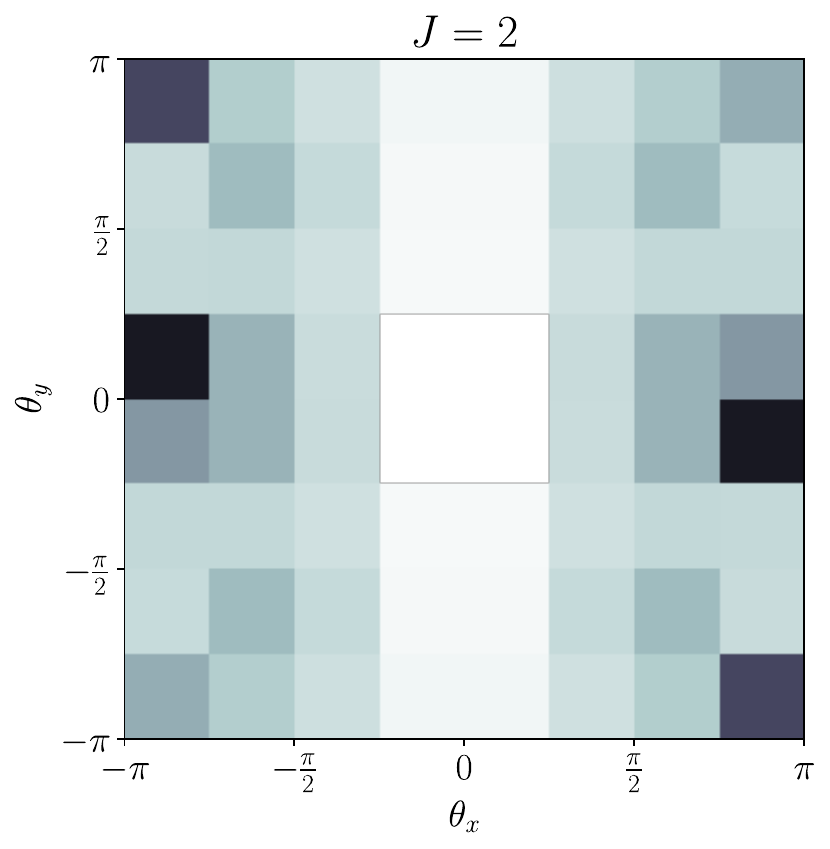}
		\hspace{20pt}
		\includegraphics[height=0.32\textwidth]{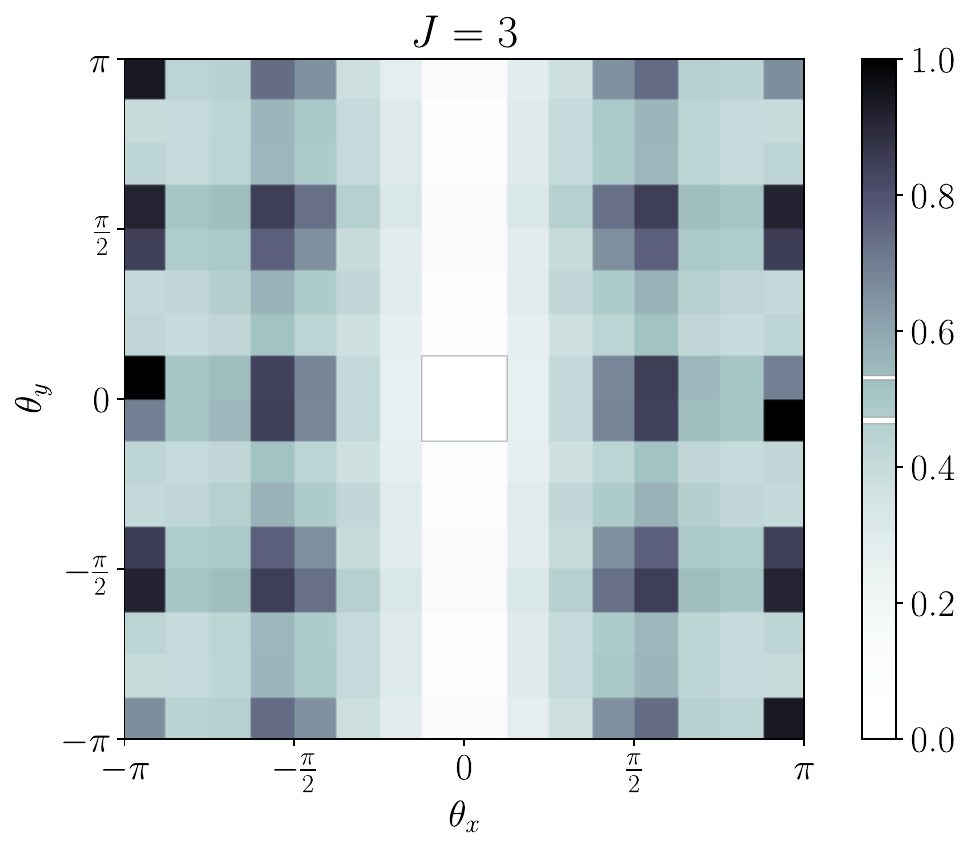}
		\vspace{-12pt}
		\caption{\rmax operators}
	\label{subfig:shiftinvariance_rmax}
	\end{subfigure}
	\vspace{-10pt} \\
	\begin{subfigure}{\textwidth}
		\centering
		\includegraphics[height=0.32\textwidth]{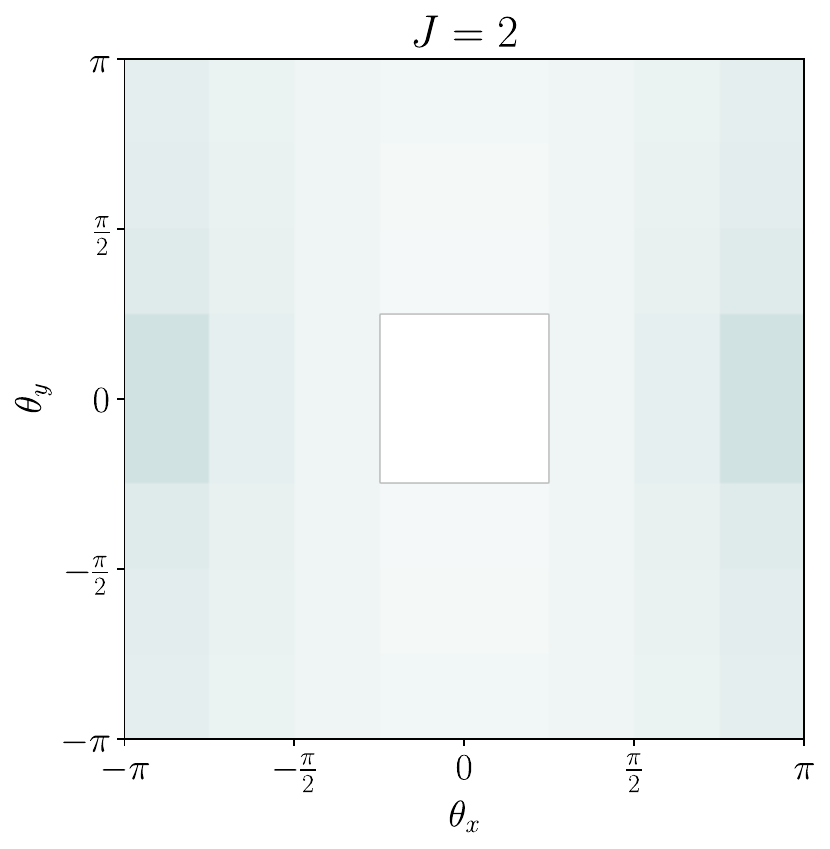}
		\hspace{20pt}
		\includegraphics[height=0.32\textwidth]{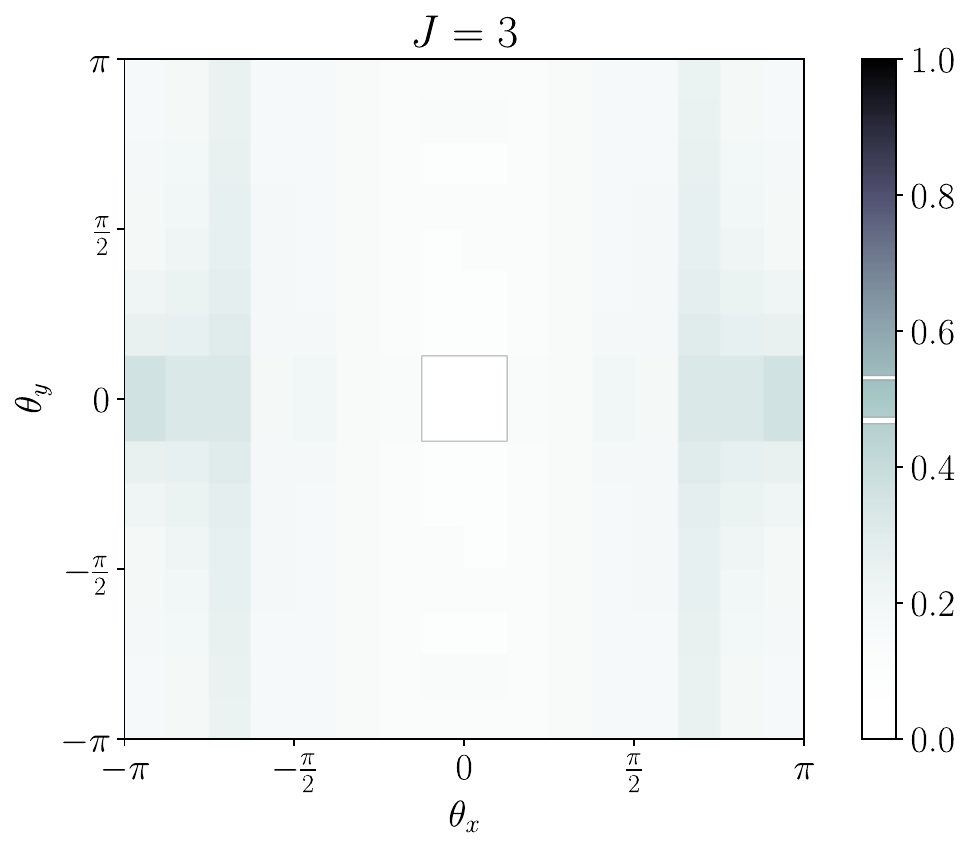}
		\vspace{-12pt}
		\caption{\cmod operators}
	\label{subfig:shiftinvariance_cmod}
	\end{subfigure}
	\caption[
		Experiments: Shift Invariance of \rmax and \cmod Operators
	]{Shift invariance of \rmax and \cmod outputs, computed on ImageNet 2012 (validation set). For each $\selectOutchannelInbigrangeWPT$, $\avgEvalDiffShiftRmaxNESelect$ (\cref{subfig:shiftinvariance_rmax}) and $\avgEvalDiffShiftCmodNESelect$ (\cref{subfig:shiftinvariance_cmod}) are plotted by applying the same procedure as in \cref{fig:scdmoment_normdiff_cmodrmax}.} 
	\label{fig:shiftinvariance}
\end{figure}

\subsubsection{MSE between \rmax and \cmod}

Each image $\selectExample \in \setof{\nexamples}$ in the dataset was converted to grayscale, from which a center crop of size $224 \times 224$ was extracted. We denote by $\inpimg_\selectExample \in \realltwoZsq$ the resulting input feature map. For any $\selectOutchannelInbigrangeWPT$, we denote by
\begin{equation}
	\rmaximgNEMultiresDepthSelect := \rmaxopNEDepthSelect(\inpimg_\selectExample)
	\qqand
	\cmodimgNEMultiresDepthSelect := \cmodopNEDepthSelect(\inpimg_\selectExample)
\end{equation}
the outputs of the $\selectOutchannel$-th \rmax and \cmod operators as defined in \eqref{eq:rmax_dt} and \eqref{eq:cmod_dt}, respectively. We adopt similar notations for the bottom-right Fourier quadrant.
Then, the normalized mean squared error between $\cmodimgNEMultiresDepthSelect$ and $\rmaximgNEMultiresDepthSelect$ was computed. It is defined by the square of
\begin{equation}
	\evalDiffRmaxCmodNESelect := \bignormtwo{\cmodimgNEMultiresDepthSelect - \rmaximgNEMultiresDepthSelect} / \bignormtwo{\cmodimgNEMultiresDepthSelect}.
\end{equation}
Finally, the  for each output channel $\selectOutchannel$, an empirical estimate for $\ExpvalDiffRmaxCmod$, introduced in \eqref{eq:normdiff_cmodrmax_stoch}, was obtained by averaging $\evalDiffRmaxCmodNESelectSquared$ over the whole dataset. We denote by $\avgEvalDiffRmaxCmodNESelectSquared$ the corresponding quantity.

Since $\rmaxopNEDepthSelect$ and $\cmodopNEDepthSelect$ are parameterized by $\complexWeightimgNEMultires{\depth}_{\selectOutchannel}$, it follows that $\avgEvalDiffRmaxCmodNESelectSquared$ depends on the filter's characteristic frequency $\freqvecMpipiMultires{\depth}_\selectOutchannel$ \eqref{eq:dtkernel_gaborfilt_discrete}. According to \cref{prop:supp_dtcwpt}, these frequencies form a regular grid in the top-right quadrant of Fourier domain. This provides a visual representation of $\avgEvalDiffRmaxCmodNESelectSquared$, as shown in \cref{fig:scdmoment_normdiff_cmodrmax}. This figure also displays $\avgEvalDiffRmaxCmodSESelectSquared$, corresponding to the bottom-right quadrant. The half-plane of negative \xcoord-values has simply been symmetrized, following \cref{remark:only2quadrants}.
We can observe a regular pattern of dark spots. More precisely, high discrepancies between max pooling and modulus seem to occur when the energy of $\complexWeightimgNEMultires{\depth}_\selectOutchannel$ or $\complexWeightimgSEMultires{\depth}_\selectOutchannel$ overlaps a dark region of \cref{fig:proxyfunction}. This result corroborates \cref{th:shiftinvariance_rmax}, which states that high discrepancies are expected for certain pathological frequencies, due to the search for a maximum value over a discrete grid.

\subsubsection{Shift invariance}

For each input image previously converted to grayscale, two crops of size $224 \times 224$ were extracted, such that the corresponding sequences $\inpimg_\selectExample$ and $\inpimg'_\selectExample$ are shifted by one pixel along the \xcoord-axis.
From these inputs, the following quantity was then computed:
\begin{equation}
	\evalDiffShiftRmaxNESelect := \bignormtwo{\rmaximgbisNEMultiresDepthSelect - \rmaximgNEMultiresDepthSelect} \,/\, \bignormtwo{\cmodimgNEMultiresDepthSelect},
\end{equation}
where $\rmaximgbisNEMultiresDepthSelect$ satisfies \eqref{eq:rmax_cmod_dtcwpt} with $\inpimg_\selectExample \leftarrow \inpimg'_\selectExample$.
Finally, for each output channel $\selectOutchannelInbigrangeWPT$, an empirical estimate for $\ExpvalDiffShiftRmax$, satisfying \eqref{eq:normdiff_rmax_transl_stoch} with $\translvecDiscrete = (1,\, 0)^\top$, was obtained by averaging $\evalDiffShiftRmaxNESelect$ over the whole dataset. We denote by $\avgEvalDiffShiftRmaxNESelect$ the corresponding quantity. We point out that shift invariance is measured relatively to the norm of the \cmod output, as explained in \cref{remark:normalization}.

On the other hand, the same procedure was applied to the \cmod operators:
\begin{equation}
	\evalDiffShiftCModNESelect := \bignormtwo{\cmodimgbisNEMultiresDepthSelect - \cmodimgNEMultiresDepthSelect} \,/\, \bignormtwo{\cmodimgNEMultiresDepthSelect},
\end{equation}
and $\avgEvalDiffShiftCmodNESelect$ was obtained as before by averaging $\evalDiffShiftCModNESelect$ over the whole dataset.

A visual representation of $\avgEvalDiffShiftRmaxNESelect$ and $\avgEvalDiffShiftCmodNESelect$ are provided in \cref{fig:shiftinvariance} (as well as the other Fourier quadrants). Two observations can be drawn here.
\begin{enumerate*}
	\item When the filter is horizontally oriented, the corresponding output is highly stable with respect to horizontal shifts. This can be explained by noticing that such kernels perform low-pass filtering along the \xcoord-axis. The exact transposed phenomenon occurs for vertical shifts.
	\item Elsewhere, we observe that high discrepancies between \rmax and \cmod outputs (\cref{fig:scdmoment_normdiff_cmodrmax}) are correlated with shift instability of \rmax (\cref{fig:shiftinvariance}, top). This is in line with \eqref{eq:scdmoment_normdiff_cmodrmax} and \eqref{eq:expval_normdiff_rmax_transl} in \cref{th:scdmoment_normdiff_cmodrmax,th:shiftinvariance_rmax}. Note that \cmod outputs are nearly shift invariant regardless the characteristic frequency $\freqvecMpipiMultires{\depth}_\selectOutchannel$ (\cref{fig:shiftinvariance}, bottom), as predicted by \cref{th:shiftinvariance_cmod} \eqref{eq:shiftinvariance_cmod}.
\end{enumerate*}

\section{Conclusion}

In this paper, we investigated the shift invariance properties captured by the max pooling operator, when applied to a convolution layer with Gabor-like kernels. We established a validity domain for near-shift invariance and confirmed our predictions through an experimental setting based on the dual-tree complex wavelet packet transform.
Our results indicate that the \cmod operator can serve as a stable proxy for \rmax, extracting comparable features---except at certain filter frequencies, where potential degeneracies may arise after max pooling.
In this context, a follow-up study applying these principles to real-world architectures was published as a conference paper \citep{LetermeCNNsShiftInvariant2024}.

A link was missing between real- and complex-valued convolutions in CNNs. By comparing the outputs of \cmod and \rmax operators, we established a connection between these two worlds, creating opportunities for extensions of the results obtained for complex wavelet transforms. To paraphrase \citet{Tygert2016}, the correspondence between standard real-valued CNNs (using max pooling) and complex wavelets is no longer ``just a vague analogy.''

\bibliography{preprint}
\bibliographystyle{tmlr}

\appendix

\section{Appendix -- Proofs}
\label{sec:appendix_proofs}

\subsection{Proof of \cref{lemma:lowfreqfun}}
\label{subsec:appendix_proofs_lemma_lowfreqfun}

\begin{proof}
  Applying the Fourier transform on \eqref{eq:lowfreqfun} yields, for any $\freqvecbis \in \mathR^2$,
	\begin{equation}
		\fourierLowfreqfun(\freqvecbis)
			= \fouriertransf{\bigl(
				\inpfunAstWavelet
			\bigr)}(\freqvecbis - \freqvec)
			= \transl_{\freqvec}\bigl(
				\fourierInpfun\,\fourierFlippedWavelet
			\bigr)(\freqvecbis).
	\end{equation}
	By hypothesis on $\wavelet$, we have
	\begin{equation}
		\supp(\fourierInpfun \, \fourierFlippedWavelet) \subset \supp\fourierFlippedWavelet \subset \linftyBall(-\freqvec,\, \supportsizeContinuous/2).
	\end{equation}
	The translation operator $\transl_{\freqvec}$ shifts the support with respect to $\freqvec$, which yields \eqref{eq:suppLowfreqfun}.
\end{proof}

\subsection{Proof of \cref{prop:shiftinvariance_cont}}
\label{subsec:appendix_proofs_prop_shiftinvariance_cont}

\begin{proof}
	Using the 2D Plancherel formula, we compute
	\begin{align*}
		\normLtwo{\transl_{\translvecContinuous} \lowfreqfun - \lowfreqfun}^2
			&= \frac1{4\pi^2} \normLtwo{\fouriertransf{\transl_{\translvecContinuous} \lowfreqfun} - \fourierLowfreqfun}^2 \\
			&= \frac1{4\pi^2} \iint_{\fourierWindowGaborContinuous} \left|
				\fourierLowfreqfun(\freqvecbis)
			\right|^2 \left|
				\rme^{-i\innerprod{\translvecContinuous}{\freqvecbis}} - 1
			\right|^2 \,\rmd^2\freqvecbis \\
			&= \frac1{4\pi^2} \iint_{\fourierWindowGaborContinuous} \left|
				\fourierLowfreqfun(\freqvecbis)
			\right|^2 \bigl(
				2 - 2\cos\innerprod{\translvecContinuous}{\freqvecbis}
			\bigr) \,\rmd^2\freqvecbis \\
			&\leq \frac1{4\pi^2} \iint_{\fourierWindowGaborContinuous} \left|
				\fourierLowfreqfun(\freqvecbis)
			\right|^2 \left|
				\innerprod{\translvecContinuous}{\freqvecbis}
			\right|^2 \,\rmd^2\freqvecbis,
	\end{align*}
	because $\cos \timeval \geq 1-\frac{\timeval^2}{2}$.
	Note that the integral is computed on a compact domain because, according to \cref{lemma:lowfreqfun}, $\supp \fourierLowfreqfun \subset \fourierWindowGaborContinuous$. Next, we use the Cauchy-Schwarz inequality to compute:
	\begin{align*}
		\forall \freqvecbis \in \fourierWindowGaborContinuous,\ \left|\innerprod{\translvecContinuous}{\freqvecbis}\right|
			&\leq \normone{\translvecContinuous} \cdot \norminfty{\freqvecbis} \\
			&\leq \frac{\supportsizeContinuous}{2} \normone{\translvecContinuous}.
	\end{align*}
	Therefore,
	\begin{equation}
		\normLtwo{\transl_{\translvecContinuous} \lowfreqfun - \lowfreqfun}^2
			\leq \frac{\supportsizeContinuous}{4} \normone{\translvecContinuous}^2 \, \normLtwo{\lowfreqfun}^2, 
	\end{equation}
	which yields the result.
\end{proof}

\subsection{Proof of \cref{lemma:fouriersampling}}
\label{subsec:appendix_proofs_lemma_fouriersampling}

\begin{proof}
	Since $\inpfun \in \shannonspaceSamplinterv$, the two-dimensional version of Shannon's sampling theorem \citep[Theorem~3.11, p.~81]{Mallat2009} yields
	\begin{equation}
		\inpfun = \sumoverVectorindices \inpimg[\vectorindex] \, \shanScalingfunSamplintervVectorindex, \qqand \fourierInpfun = \sumoverVectorindices \inpimg[\vectorindex] \, \fouriertransf{\shanScalingfunSamplintervVectorindex}.
	\label{eq:fouriersampling_proof}
	\end{equation}
	Moreover, using \eqref{eq:fourierShanScalingfun}, we can show that, for any $\freqvecbis \in \fourierWindowSampling$, 
	\begin{equation}
		\fouriertransf{\shanScalingfunSamplintervVectorindex}(\freqvecbis) = \fouriertransf{\shanScalingfunSamplinterv}(\freqvecbis) \, \rme^{-i\innerprod{\samplinterv\freqvecbis}{\vectorindex}} = \samplinterv \, \rme^{-i\innerprod{\samplinterv\freqvecbis}{\vectorindex}}.
	\label{eq:fourierShanScalingfunVectorindex}
	\end{equation}
	Therefore, plugging \eqref{eq:fourierShanScalingfunVectorindex} into \eqref{eq:fouriersampling_proof} proves \eqref{eq:fouriersampling_1}.
	
	Then, by combining \eqref{eq:fouriersampling_1} with the Plancherel formula, we get
	\begin{align*}
		\normLtwo{\inpfun}^2
			&= \frac1{4\pi^2} \bignormLtwo{\fourierInpfun}^2 \\
			&= \frac1{4\pi^2} \iint_{\fourierWindowSampling} \bigl|
				\fourierInpfun(\freqvecbis)
			\bigr|^2 \,\rmd^2\freqvecbis \\
			&= \frac1{4\pi^2} \iint_{\fourierWindowSampling} \bigl|
				\samplinterv \, \fourierInpimg(\samplinterv\freqvecbis)
			\bigr|^2 \,\rmd^2\freqvecbis.
	\end{align*}
	The integral is performed on $\fourierWindowSampling$ because $\inpfun \in \shannonspaceSamplinterv$. Then, by applying the change of variable $\freqvecbis' \leftarrow \samplinterv\freqvecbis$, we get
	\begin{align*}
	\normLtwo{\inpfun}^2
		&= \frac1{4\pi^2} \iint_{\linftyBall(\pi)} \bigl|\fourierInpimg(\freqvecbis')\bigr|^2 \,\rmd^2\freqvecbis' \\
		&= \frac1{4\pi^2} \bignormLtwo{\fourierInpimg}^2 = \normtwo{\inpimg}^2,
	\end{align*}
	hence \eqref{eq:fouriersampling_2}, which concludes the proof.
\end{proof}

\subsection{Proof of \cref{prop:discrete2continuous}}
\label{subsec:appendix_proofs_prop_discrete2continuous}

\begin{proof}
	First, $\inpfunInterp$ and $\waveletInterp$ are well defined because $\inpimg \in \realltwoZsq$ and $\complexWeightimg \in \complexltwoZsq$. By construction, $\inpfunInterp$ and $\waveletInterp \in \shannonspaceSamplinterv$. Therefore, according to Shannon's sampling theorem \citep[Theorem~3.11, p.~81]{Mallat2009},
	\begin{equation}
		\inpfunInterp := \samplinterv \sumoverVectorindices \inpfunInterp(\samplinterv\vectorindex) \,\shanScalingfunSamplintervVectorindex \qqand \waveletInterp := \samplinterv \sumoverVectorindices \waveletInterp(\samplinterv\vectorindex) \,\shanScalingfunSamplintervVectorindex.
	\label{eq:shannon_proof}
	\end{equation}
	By uniqueness of decompositions in an orthonormal basis, we get \eqref{eq:shannon}. Moreover, using \eqref{eq:fouriersampling_1} in \cref{lemma:fouriersampling}, we get, for any $\freqvecbis \in \fourierWindowSampling$,
	\begin{equation}
		\fourierWaveletInterp(\freqvecbis) = \samplinterv \, \fourierComplexWeightimg(\samplinterv\freqvecbis).
	\label{eq:fourierWaveletInterp}
	\end{equation}
	Since $\fourierWaveletInterp(\freqvecbis) = 0$ outside $\fourierWindowSampling$, \eqref{eq:fourierWaveletInterp} is true for any $\freqvecbis \in \mathR^2$. Therefore, by hypothesis on $\complexWeightimg$,
	\begin{equation}
		\supp\fourierWaveletInterp \subset \linftyBall\bigl(
			\freqvecMpipi/\samplinterv,\, \supportsizeDiscrete/(2\samplinterv)
		\bigr),
	\end{equation}
	which yields \eqref{eq:gaborfilter}.

	We now prove \eqref{eq:discrete2continuous}. For $\vectorindexInZtwo$, we compute:
	\begin{align*}
	(\inpfunAstWaveletInterp)\left(\sub \samplinterv \vectorindex\right)
		&= \iint_{\mathR^2} \inpfunInterp(\sub \samplinterv \vectorindex - \spatialvec) \,\flippedWaveletInterp(\spatialvec) \,\rmd^2\spatialvec \\
		&= \iint_{\mathR^2} \sumoverVectorindicesbis \inpimg[\vectorindexbis] \shanScalingfunSamplinterv_{\vectorindexbis}(\sub \samplinterv \vectorindex - \spatialvec) \, \flippedWaveletInterp(\spatialvec) \,\rmd^2\spatialvec \\
		&= \sumoverVectorindicesbis \inpimg[\vectorindexbis] \iint_{\mathR^2} \shanScalingfunSamplinterv_{\vectorindexbis}(\sub \samplinterv \vectorindex - \spatialvec) \, \flippedWaveletInterp(\spatialvec) \,\rmd^2\spatialvec.
	\end{align*}
	The sum-integral interchange is possible because $\inpimg$ has a finite support. Then:
	\begin{align}
	(\inpfunAstWaveletInterp) \left(
		\sub \samplinterv \vectorindex
	\right)
		&= \sumoverVectorindicesbis \inpimg[\vectorindexbis] \iint_{\mathR^2} \flippedWaveletInterp(\spatialvec) \, \shanScalingfunSamplinterv\bigl(
			\samplinterv(\sub \vectorindex - \vectorindexbis) - \spatialvec
		\bigr) \,\rmd^2\spatialvec \\
		&= \sumoverVectorindicesbis \inpimg[\vectorindexbis] \, \bigl(
			\flippedWaveletInterp \ast \shanScalingfunSamplinterv
		\bigr)\bigl(
			\samplinterv(\sub \vectorindex - \vectorindexbis)
		\bigr)
	\label{eq:discrete2continuous_proof1}
	\end{align}
	Since $\{\shanScalingfunSamplintervVectorindex\}_{\vectorindexInZtwo}$ is an orthonormal basis of $\shannonspaceSamplinterv$, the definition of $\waveletInterp$ in \eqref{eq:interp} implies, for any $\vectorindexbis' \in \mathZ^2$,
	\begin{equation}
		\flippedComplexWeightimg[\vectorindexbis'] = \innerprod{\waveletInterp}{\shanScalingfunSamplinterv_{-\vectorindexbis'}} = \left(\flippedWaveletInterp \ast \shanScalingfunSamplinterv\right)(\samplinterv\vectorindexbis'),
	\label{eq:discrete2continuous_proof2}
	\end{equation}
	because $\shanScalingfunSamplinterv$ is even.
	Therefore, plugging \eqref{eq:discrete2continuous_proof2} with $\vectorindexbis' \leftarrow (\sub\vectorindex - \vectorindexbis)$ into \eqref{eq:discrete2continuous_proof1} yields
	\begin{equation}
		(\inpfunAstWaveletInterp)\left(\sub \samplinterv \vectorindex\right)
			= \sumoverVectorindicesbis \inpimg[\vectorindexbis] \, \flippedComplexWeightimg[\sub \vectorindex - \vectorindexbis]
			= \bigl(\inpimg \ast \flippedComplexWeightimg\bigr)[\sub \vectorindex],
	\end{equation}
	hence the result.
\end{proof}

\subsection{Proof of \cref{lemma:commut_interp_transl}}
\label{subsec:appendix_proofs_lemma_commut_interp_transl}

\begin{proof}
	Let $\translvecDiscrete \in \mathR^2$. By definition of $\inpfunTranslInterp$ and $\translInpimg$,
		\begin{equation}
			\inpfunTranslInterp = \samplinterv \sumoverVectorindices \translSamplintervVec\inpfunInterp(\samplinterv\vectorindex) \, \shanScalingfunSamplintervVectorindex.
		\label{eq:interp_transl}
		\end{equation}
		On the other hand, $\inpfunInterp \in \shannonspaceSamplinterv$ by construction. Therefore, $\translSamplintervVec\inpfunInterp \in \shannonspaceSamplinterv$. Then, according to Shannon's sampling theorem \citep[Theorem~3.11, p.~81]{Mallat2009}, we get
		\begin{equation}
			\translSamplintervVec\inpfunInterp = \samplinterv \sumoverVectorindices \translSamplintervVec\inpfunInterp(\samplinterv\vectorindex)\, \shanScalingfunSamplintervVectorindex,
		\end{equation}
		which concludes the proof.
\end{proof}

\subsection{Proof of \cref{lemma:normequality}}
\label{subsec:appendix_proofs_lemma_normequality}

\begin{proof}
	Let us write:
	\begin{equation}
		\sumoverVectorindices \bigl|
			\transl_{\translvecContinuous}\lowfreqfun(\subsamplinterv\vectorindex) - \lowfreqfun(\subsamplinterv\vectorindex)
		\bigr|^2 = \sumoverVectorindices \bigl|
			\inpfunbis(\subsamplinterv\vectorindex)
		\bigr|^2 = \frac1{\subsamplinterv^2} \normtwo{\inpimgbis}^2,
	\label{eq:normequality_proof_1}
	\end{equation}
	where we have denoted, for any $\vectorindex \in \mathZ^2$,
	\begin{equation}
		\inpfunbis := \transl_{\translvecContinuous} \lowfreqfun - \lowfreqfun \qqand \inpimgbis[\vectorindex] := \subsamplinterv \inpfunbis(\subsamplinterv\vectorindex).
	\end{equation}
	According to \cref{prop:discrete2continuous} \eqref{eq:gaborfilter}, $\waveletInterp \in \Gaborfilter{\freqvecMpipi / \samplinterv}{\supportsizeDiscrete / \samplinterv}$. Therefore, according to \cref{lemma:lowfreqfun},
	\begin{equation}
		\supp\fourierLowfreqfun \subset \linftyBall\left(\frac{\supportsizeDiscrete}{2\samplinterv}\right).
	\end{equation}
	Moreover, by hypothesis, $\supportsizeDiscrete \leq \pi/\sub$; thus,
	\begin{equation}
		\linftyBall\left(\frac{\supportsizeDiscrete}{2\samplinterv}\right) \subset \linftyBall\left(\frac{\pi}{\subsamplinterv}\right),
	\end{equation}
	which yields \eqref{eq:suppLowfreqfunSubsamp}, and $\inpfunbis \in \shannonspaceSubsamplinterv$. Then, according to \cref{lemma:fouriersampling} \eqref{eq:fouriersampling_2} with $\inpimg \leftarrow \inpimgbis$, $\inpfun \leftarrow \inpfunbis$ and $\samplinterv \leftarrow \subsamplinterv$,
	\begin{equation}
		\normtwo{\inpimgbis} = \normLtwo{\inpfunbis} = \normLtwo{\transl_{\translvecContinuous} \lowfreqfun - \lowfreqfun}.
	\label{eq:normequality_proof_2}
	\end{equation}
	Therefore, plugging \eqref{eq:normequality_proof_2} into \eqref{eq:normequality_proof_1} yields \eqref{eq:normequality_1}.

	Furthermore, according again to \cref{lemma:fouriersampling},
	\begin{equation}
		\normLtwo{\lowfreqfun}^2 = \normtwo{\lowfreqimg}^2,
	\label{eq:normequality_proof_3}
	\end{equation}
	where we have defined, for any $\vectorindex \in \mathZ^2$,
	\begin{equation}
		\lowfreqimg[\vectorindex] := \subsamplinterv \lowfreqfun(\subsamplinterv\vectorindex).
	\label{eq:lowfreqimg}
	\end{equation}
	Then,
	\begin{align*}
		\normtwo{\lowfreqimg}^2
			&= \subsamplinterv^2 \sumoverVectorindices \left|
				\bigl(
					\inpfunAstWaveletInterp
				\bigr)(\subsamplinterv\vectorindex)
			\right|^2 & \mbox{(acc.\@ to \eqref{eq:lowfreqfun_interp})} \\
			&= \subsamplinterv^2 \sumoverVectorindices \left|
				(\inpimg \ast \flippedComplexWeightimg) \downarrow (2\sub) [\vectorindex]
			\right|^2 & \mbox{(acc.\@ to \cref{prop:discrete2continuous} with $\sub \leftarrow 2\sub$)} \\
			&= \subsamplinterv^2 \, \bignormtwo{\cmodopSub\inpimg}^2. & \mbox{(acc.\@ to \eqref{eq:cmod})}
	\end{align*}
	Finally, plugging this result into \eqref{eq:normequality_proof_3} yields \eqref{eq:normequality_2} and concludes the proof.
\end{proof}

\subsection{Proof of \cref{th:shiftinvariance_cmod}}
\label{subsec:appendix_proofs_th_shiftinvariance_cmod}

\begin{proof}
	As in \cref{lemma:normequality}, we consider the low-frequency function $\lowfreqfun$ satisfying \eqref{eq:lowfreqfun_interp}, and denote $\subsamplinterv := 2\sub\samplinterv$. We can write
	\begin{equation}
		|\inpfunAstWaveletInterp| = |\lowfreqfun|
			\qqand
		|\transl_{\samplinterv\translvecDiscrete} \inpfunAstWaveletInterp| = |\transl_{\samplinterv\translvecDiscrete} \lowfreqfun|.
	\label{eq:module_complexconv}
	\end{equation}
	Recall that $\cmodopSub\inpimg = \bigl|
		(\inpimg \ast \flippedComplexWeightimg) \downarrow (2\sub)
	\bigr|$, such as defined in \eqref{eq:cmod}. According to \cref{prop:discrete2continuous} \eqref{eq:discrete2continuous} and \cref{cor:discrete2continuous} \eqref{eq:discrete2continuous_shift} with $\sub \leftarrow 2\sub$, we therefore get
	\begin{align}
		\cmodopSub \inpimg[\vectorindex] &= \left|\lowfreqfun(\subsamplinterv \vectorindex)\right|;
	\label{eq:shiftinvariance_cmod_proof1} \\
		\cmodopSub (\translInpimg)[\vectorindex] &= \left|(\transl_{\samplinterv\translvecDiscrete}\lowfreqfun)(\subsamplinterv \vectorindex)\right|.
	\label{eq:shiftinvariance_cmod_proof2}
	\end{align}
	Then, using \eqref{eq:shiftinvariance_cmod_proof1}, \eqref{eq:shiftinvariance_cmod_proof2} and the reverse triangle inequality,
	\begin{align*}
		\bignormtwo{\cmodopSub (\translInpimg) - \cmodopSub \inpimg}^2
			&= \sumoverVectorindices \Bigl|
				\bigl|
					(\transl_{\samplinterv\translvecDiscrete}\lowfreqfun)(\subsamplinterv\vectorindex)
				\bigr| - \bigl|
					\lowfreqfun(\subsamplinterv\vectorindex)
				\bigr|
			\Bigr|^2 \\
			&\leq \sumoverVectorindices \Bigl|
				(\transl_{\samplinterv\translvecDiscrete}\lowfreqfun)(\subsamplinterv\vectorindex) - \lowfreqfun(\subsamplinterv\vectorindex)
			\Bigr|^2.
	\end{align*}
	Since condition \eqref{eq:condition_bandwidth} is satisfied, we can use \cref{lemma:normequality} \eqref{eq:normequality_1} with $\translvecContinuous \leftarrow \samplinterv\translvecDiscrete$:
	\begin{equation}
		\bignormtwo{\cmodopSub (\translInpimg) - \cmodopSub \inpimg}^2 \leq \frac1{\subsamplinterv^2} \normLtwo{\transl_{\samplinterv\translvecDiscrete}\lowfreqfun - \lowfreqfun}^2
	\end{equation}
	Next, according to \cref{prop:shiftinvariance_cont} with $\supportsizeContinuous \leftarrow \supportsizeDiscrete / \samplinterv$ and $\translvecContinuous \leftarrow \samplinterv \translvecDiscrete$, we then get the following bound:
	\begin{equation}
		\bignormtwo{\cmodopSub (\translInpimg) - \cmodopSub \inpimg}^2
			\leq \frac{\shiftinvCmod(\supportsizeDiscrete\translvecDiscrete)^2}{\subsamplinterv^2} \normLtwo{\lowfreqfun}^2.
	\label{eq:shiftinvariance_cmod_proof3}
	\end{equation}
	Finally, using \cref{lemma:normequality} \eqref{eq:normequality_2} yields \eqref{eq:shiftinvariance_cmod}, which completes the proof.
\end{proof}

\subsection{Proof of \cref{prop:shiftinvariance_normcmod}}
\label{subsec:appendix_proofs_prop_shiftinvariance_normcmod}

\begin{proof}
	Let $\inpimg \in \realltwoZsq$ and $\samplinterv > 0$. We consider $\lowfreqfun \in \complexLtwoRsq$ as the ``low-frequency'' function satisfying \eqref{eq:lowfreqfun_interp}. Again, we introduce $\subsamplinterv := 2\sub \samplinterv $ and $\lowfreqimg \in \complexltwoZsq$ satisfying \eqref{eq:lowfreqimg}. Moreover, for any $\outimg \in \realltwoZsq$, we denote by $\interpshannon{\outimg}{\subsamplinterv}$ the Shannon interpolation of $\outimg$ parameterized by $\subsamplinterv$, analogously to \eqref{eq:interp}:
	\begin{equation}
		\interpshannon{\outimg}{\subsamplinterv} := \sumoverVectorindices \outimg[\vectorindex] \,\shanScalingfunSubsamplintervVectorindex.
	\label{eq:interpbis}
	\end{equation}
	
	On the one hand, \cref{lemma:normequality} provides \eqref{eq:normequality_2}. On the other hand, we seek a similar result with $\inpimg \leftarrow \translInpimg$. For this purpose, \eqref{eq:shiftinvariance_cmod_proof2} can be rewritten
	\begin{equation}
		\cmodopSub(\translInpimg)[\vectorindex]
			= \bigl|\translSubsamplintervSubvec \lowfreqfun(\subsamplinterv\vectorindex)\bigr|,
	\label{eq:cmod_transl}
	\end{equation}
	with $\translvecDiscrete' := \translvecDiscrete / (2\sub) $. Furthermore, according to \cref{lemma:normequality} \eqref{eq:suppLowfreqfunSubsamp}, $\lowfreqfun \in \shannonspaceSubsamplinterv$. Therefore, Shannon's sampling theorem \citep[Theorem~3.11, p.~81]{Mallat2009} with $\samplinterv \leftarrow \subsamplinterv$ yields
	\begin{align*}
		\lowfreqfun
			&= \subsamplinterv \sumoverVectorindices \lowfreqfun(\subsamplinterv\vectorindex) \, \shanScalingfunSubsamplintervVectorindex \\
			&= \sumoverVectorindices \lowfreqimg[\vectorindex] \, \shanScalingfunSubsamplintervVectorindex = \lowfreqfunInterpSubsampl,
	\end{align*}
	where we have used the notations introduced in \eqref{eq:lowfreqimg} and \eqref{eq:interpbis}.
	Then, using \cref{lemma:commut_interp_transl} with $\inpimg \leftarrow \lowfreqimg$, $\translvecDiscrete \leftarrow \translvecDiscrete'$ and $\samplinterv \leftarrow \subsamplinterv$, we get
	\begin{equation}
		\lowfreqfunTranslInterpSubsampl = \transl_{\subsamplinterv\translvecDiscrete'} \lowfreqfunInterpSubsampl = \transl_{\subsamplinterv\translvecDiscrete'} \lowfreqfun.
	\label{eq:interpbis_translbis}
	\end{equation}
	Moreover, \eqref{eq:shannon} (from \cref{prop:discrete2continuous}) with $\inpimg \leftarrow \translSubvecLowfreqimg$ and $\samplinterv \leftarrow \subsamplinterv$ becomes
	\begin{equation}
		\translSubvecLowfreqimg[\vectorindex] = \subsamplinterv \, \lowfreqfunTranslInterpSubsampl(\subsamplinterv\vectorindex),
	\label{eq:interpbisTranslLowfreqfun}
	\end{equation}
	and inserting \eqref{eq:interpbis_translbis} into \eqref{eq:interpbisTranslLowfreqfun} yields
	\begin{equation}
		\translSubvecLowfreqimg[\vectorindex] = \subsamplinterv \, \transl_{\subsamplinterv\translvecDiscrete'} \lowfreqfun(\subsamplinterv\vectorindex).
	\label{eq:translbis}
	\end{equation}
	Therefore, \eqref{eq:cmod_transl} and \eqref{eq:translbis} imply
	\begin{equation}
		\bignormtwo{\cmodopSub(\translInpimg)} = \frac1{\subsamplinterv} \normtwo{\translSubvecLowfreqimg}.
	\label{eq:norm_cmod_transl}
	\end{equation}
	Moreover, since $\lowfreqfun \in \shannonspaceSubsamplinterv$, and according to \eqref{eq:translbis}, we can use \cref{lemma:fouriersampling} with $\samplinterv \leftarrow \subsamplinterv$, $\inpfun \leftarrow \transl_{\subsamplinterv\translvecDiscrete'} \lowfreqfun$ and $\inpimg \leftarrow \translSubvecLowfreqimg$. We get
	\begin{equation}
		\normtwo{\translSubvecLowfreqimg} = \normLtwo{\translSubsamplintervSubvec \lowfreqfun} = \normLtwo{\lowfreqfun},
	\label{eq:norm_translbis}
	\end{equation}
	and plugging \eqref{eq:norm_translbis} into \eqref{eq:norm_cmod_transl} yields
	\begin{equation}
		\bignormtwo{\cmodopSub(\translInpimg)} = \frac1{\subsamplinterv} \normLtwo{\lowfreqfun}.
	\label{eq:shiftinvariance_cmod_proof5}
	\end{equation}
	Finally, considering \cref{lemma:normequality} \eqref{eq:normequality_2} concludes the proof.
\end{proof}

\subsection{Proof of \cref{prop:approx_periodicfun}}
\label{subsec:appendix_proofs_prop_approx_periodicfun}

\begin{proof}
	Let us write:
	\begin{align*}
		(\inpfunAstRealWavelet)(\spatialvec + \searchvec)
			&- \bigl|
			(\inpfunAstWavelet)(\spatialvec)
		\bigr| \, \evalCosfun \\
			&\qquad= \Real\left(
				(
					\inpfunAstWavelet
				)(\spatialvec + \searchvec)
			\right) - \bigl|
				(
					\inpfunAstWavelet
				)(\spatialvec)
			\bigr| \, \Real\left(
				\rme^{-i\innerprod{\freqvec}{\searchvec}} \, \rme^{i\phase(\spatialvec)}
			\right) \\
			&\qquad= \Real\left(
				(
					\inpfunAstWavelet
				)(\spatialvec + \searchvec)
			\right) - \Real\left(
				\bigl|
					(
						\inpfunAstWavelet
					)(\spatialvec)
				\bigr| \, \rme^{i\phase(\spatialvec)} \, \rme^{-i\innerprod{\freqvec}{\searchvec}}
			\right) \\
			&\qquad= \Real\left(
				(
					\inpfunAstWavelet
				)(\spatialvec + \searchvec)
			\right) - \Real\left(
				(
					\inpfunAstWavelet
				)(\spatialvec) \, \rme^{-i\innerprod{\freqvec}{\searchvec}}
			\right) \\
			&\qquad= \Real\left(
				(
					\inpfunAstWavelet
				)(\spatialvec + \searchvec) -
				(
					\inpfunAstWavelet
				)(\spatialvec) \, \rme^{-i\innerprod{\freqvec}{\searchvec}}
			\right).
	\end{align*}
	Therefore,
	\begin{align*}
		\Bigl|
			(\inpfunAstRealWavelet)(\spatialvec + \searchvec) - \bigl|
				(\inpfunAstWavelet)(\spatialvec)
			\bigr| \, \evalCosfun 
		\Bigr|
			&\leq \left|
				(
					\inpfunAstWavelet
				)(\spatialvec + \searchvec) -
				(
					\inpfunAstWavelet
				)(\spatialvec) \, \rme^{-i\innerprod{\freqvec}{\searchvec}}
			\right| \\
			&= \left|
				\lowfreqfun(\spatialvec + \searchvec) \, \rme^{-i\innerprod{\freqvec}{\spatialvec + \searchvec}} - \lowfreqfun(\spatialvec) \, \rme^{-i\innerprod{\freqvec}{\spatialvec + \searchvec}}
			\right|,
	\end{align*}
	which yields \eqref{eq:approx_periodicfun} and concludes the proof.
\end{proof}

\subsection{Proof of \cref{lemma:true_versus_false_maxvalues}}
\label{subsec:appendix_proofs_lemma_true_versus_false_maxvalues}

\begin{proof}
	We apply \cref{prop:approx_periodicfun} with $\searchvec \leftarrow \searchvecMaxSelect$ and $\searchvec \leftarrow \searchvecApproxMaxSelect$, respectively:
	\begin{align}
		\evalMaxExactfunInterp
			&\leq \approxfunInterp\bigl(
				\spatialvecSelect,\, \searchvecMaxSelect
			\bigr) + \left|
				\lowfreqfun\bigl(
					\spatialvecSelect + \searchvecMaxSelect
				\bigr) - \lowfreqfun(\spatialvecSelect)
			\right|;
	\label{eq:true_versus_false_maxvalues_1} \\
		\evalMaxApproxfunInterp
			&\leq \exactfunInterp\bigl(
				\spatialvecSelect,\, \searchvecApproxMaxSelect
			\bigr) + \left|
				\lowfreqfun\bigl(
					\spatialvecSelect + \searchvecApproxMaxSelect
				\bigr) - \lowfreqfun(\spatialvecSelect)
			\right|.
	\label{eq:true_versus_false_maxvalues_2}
	\end{align}
	By construction, we have, for any $\vectorindexbis \in \poolinggrid$,
	\begin{equation}
		\approxfunInterp\bigl(
			\spatialvecSelect,\, \searchvecSelect
		\bigr) \leq \evalMaxApproxfunInterp
		\qqand
		\exactfunInterp\bigl(
			\spatialvecSelect,\, \searchvecSelect
		\bigr) \leq \evalMaxExactfunInterp.
	\end{equation}
	Moreover, by definition, there exists $\vectorindexbis$ and $\vectorindexbis' \in \poolinggrid$ such that $\searchvecMaxSelect = \searchvec_{\vectorindexbis}$ and $\searchvecApproxMaxSelect = \searchvec_{\vectorindexbis'}$. Therefore, \eqref{eq:true_versus_false_maxvalues_1} and \eqref{eq:true_versus_false_maxvalues_2} yield, respectively,
	\begin{align}
		\evalMaxExactfunInterp
			&\leq \evalMaxApproxfunInterp + \left|
				\lowfreqfun\bigl(
					\spatialvecSelect + \searchvecMaxSelect
				\bigr) - \lowfreqfun(\spatialvecSelect)
			\right|;
	\label{eq:true_versus_false_maxvalues_3} \\
		\evalMaxApproxfunInterp
			&\leq \evalMaxExactfunInterp + \left|
				\lowfreqfun\bigl(
					\spatialvecSelect + \searchvecApproxMaxSelect
				\bigr) - \lowfreqfun(\spatialvecSelect)
			\right|,
	\label{eq:true_versus_false_maxvalues_4}
	\end{align}
	which yields \eqref{eq:true_versus_false_maxvalues} and concludes the proof.
\end{proof}

\subsection{Proof of \cref{prop:diffmodpool}}
\label{subsec:appendix_proofs_prop_diffmodpool}

\begin{proof}
	Let us write:
	\begin{align*}
		\bignormtwo{\detailCmodimg - \detailRmaximg}^2
			&= \sumoverVectorindices\left(
				\detailCmodimg[\vectorindex] - \detailRmaximg[\vectorindex]
			\right)^2 \\
			&= \sumoverVectorindices\left(
				\detailCmodimg[\vectorindex]
				- \detailCmodimg[\vectorindex] \maxVectorindexInPoolinggrid \evalCosfunInpimgGrid
			\right. \\
			&\qquad\qquad\qquad\qquad \left.
				+ \detailCmodimg[\vectorindex] \maxVectorindexInPoolinggrid \evalCosfunInpimgGrid
				- \detailRmaximg[\vectorindex]
			\right)^2 \\
			&= \sumoverVectorindices\left(
				\deltaopSubGrid\inpimg[\vectorindex] + \evalMaxApproxfunInterp - \evalMaxExactfunInterp
			\right)^2,
	\end{align*}
	according to \eqref{eq:delta_outimg_cmod}, \eqref{eq:evalMaxExactfunInterp} and \eqref{eq:evalMaxApproxfunInterp}. Then, using the triangle inequality, we get
	\begin{equation}
		\bignormtwo{\detailCmodimg - \detailRmaximg} \leq \bignormtwo{\deltaopSubGrid\inpimg} + \left(
			\sumoverVectorindices\left(
				\evalMaxApproxfunInterp - \evalMaxExactfunInterp
			\right)^2
		\right)^{1/2}.
	\label{eq:diffmodpool_proof1}
	\end{equation}
	Furthermore, \cref{lemma:true_versus_false_maxvalues} yields
	\begin{align}
		\sumoverVectorindices\left(
			\evalMaxApproxfunInterp - \evalMaxExactfunInterp
		\right)^2
		&\leq \sumoverVectorindices \, \max\limits_{
			\searchvec \in \left\{
				\searchvecMaxSelect,\, \searchvecApproxMaxSelect
			\right\}
		}\Bigl|
			\lowfreqfun(
				\spatialvecSelect + \searchvec
			) - \lowfreqfun(\spatialvecSelect)
		\Bigr|^2 \\
		&\leq \sumoverVectorindices\Bigl|
			\lowfreqfun(
				\spatialvecSelect + \searchvec_0
			) - \lowfreqfun(\spatialvecSelect)
		\Bigr|^2,
	\end{align}
	according to \cref{hyp:bound_sumofdiffs}. Next, since \eqref{eq:condition_bandwidth} is satisfied, we can use \cref{lemma:normequality} \eqref{eq:normequality_1} with $\translvecContinuous \leftarrow \searchvec_0$. We get
	\begin{align*}
		\sumoverVectorindices\left(
			\evalMaxApproxfunInterp - \evalMaxExactfunInterp
		\right)^2
			&\leq \frac1{4\sub^2\samplinterv^2} \normLtwo{\transl_{\searchvec_0}\lowfreqfun - \lowfreqfun}^2 \\
			&\leq \shiftinvCmod(\supportsizeDiscrete\searchvec_0/\samplinterv)^2 \, \frac1{4\sub^2\samplinterv^2} \normLtwo{\lowfreqfun}^2 & \mbox{(acc.\@ to \cref{prop:shiftinvariance_cont})} \\
			&= \shiftinvCmod(\supportsizeDiscrete\searchvec_0/\samplinterv)^2 \, \bignormtwo{\cmodopSub\inpimg}^2. & \mbox{(acc.\@ to \cref{lemma:normequality} \eqref{eq:normequality_2})}
	\end{align*}
	Since, according to \cref{hyp:bound_sumofdiffs}, $\normtwo{\searchvec_0} = \sqrt{2}\gridhalfsize\sub\samplinterv$, it comes that $\normone{\searchvec_0} = 2\gridhalfsize\sub\samplinterv$. Therefore,
	\begin{equation}
		\shiftinvCmod(\supportsizeDiscrete\searchvec_0/\samplinterv)^2 = \frac{\supportsizeDiscrete^2\normone{\searchvec_0}^2}{4\samplinterv^2} = (\gridhalfsize\sub\supportsizeDiscrete)^2,
	\end{equation}
	which yields
	\begin{equation}
		\sumoverVectorindices\left(
			\evalMaxApproxfunInterp - \evalMaxExactfunInterp
		\right)^2 \leq \evalLowerboundDiffRmaxCmod^2 \, \bignormtwo{\cmodopSub\inpimg}^2.
	\label{eq:diffmodpool_proof2}
	\end{equation}
	Finally, plugging \eqref{eq:diffmodpool_proof2} into \eqref{eq:diffmodpool_proof1} concludes the proof.
\end{proof}

\subsection{Proof of \cref{prop:indep}}
\label{subsec:appendix_proofs_prop_indep}

\begin{proof}
	We suppose that \cref{hyp:indep_phase_modulus} is satisfied and we consider $\spatialvec \in \mathR^2$. For a given $\boundVectorindex \in \nonzeroMathN$, we introduce the random variable 
	\begin{equation}
		{\normtwoStochMagnitudeInpimgBounded} := \sqrt{\sumoverArbitrarygrid \stochMagnitudeInterp(\spatialvec_{\vectorindexbis})^2}.
	\end{equation}
	According to \cref{hyp:indep_phase_modulus}, $\stochComplexPhaseInterp(\spatialvec)$ is jointly independent of $\stochMagnitudeInterp(\spatialvec_{\vectorindexbis})$ for $\vectorindexbis \in \range{-\boundVectorindex}{\boundVectorindex}^2$. Therefore, by composition, $\stochComplexPhaseInterp(\spatialvec)$ is also independent of $\normtwoStochMagnitudeInpimgBounded$. Moreover, according to \eqref{eq:modulus_complexconv_stoch} and \eqref{eq:norm_outimg_cmod}, $\normtwoStochMagnitudeInpimgBounded$ converges almost surely towards $\normtwoStochMagnitudeInpimg$, which proves independence between $\stochComplexPhaseInterp(\spatialvec)$ and $\normtwoStochMagnitudeInpimg$.
	
	Next, we prove conditional independence between $\stochComplexPhaseInterp(\spatialvec)$ and $\stochMagnitudeInterp(\spatialvec)$ given $\normtwoStochMagnitudeInpimg$. According to \cref{hyp:indep_phase_modulus},
	\begin{equation}
		\Bigl(
			\stochMagnitudeInterp(\spatialvec),\, \normtwoStochMagnitudeInpimgBounded
		\Bigr) \indep \stochComplexPhaseInterp(\spatialvec),
	\end{equation}
	where $\indep$ stands for independence. This is because $\normtwoStochMagnitudeInpimgBounded$ only depends on a finite number of $\stochMagnitudeInterp(\spatialvec_{\vectorindexbis})$. Therefore,
	\begin{equation}
		\stochComplexPhaseInterp(\spatialvec) \indep \stochMagnitudeInterp(\spatialvec) \bigm|\normtwoStochMagnitudeInpimgBounded.
	\end{equation}
	Finally, since $\normtwoStochMagnitudeInpimgBounded$ converges almost surely towards $\normtwoStochMagnitudeInpimg$, it comes that $\stochComplexPhaseInterp(\spatialvec)$ and $\stochMagnitudeInterp(\spatialvec)$ are conditionally independent given $\normtwoStochMagnitudeInpimg$.
\end{proof}

\subsection{Proof of \cref{th:scdmoment_normdiff_cmodrmax}}
\label{subsec:appendix_proofs_th_scdmoment_normdiff_cmodrmax}

\begin{proof}
	We consider $\vectorindexInZtwo$. By construction, $\stochOneminusCosmaxInterp(\spatialvecSelect) := 1 - \stochCosmaxInterp(\spatialvecSelect)$ only depends on $\stochComplexPhaseInterp(\spatialvecSelect)$. Therefore, under \cref{hyp:indep_phase_modulus}, \cref{prop:indep} implies
	\begin{equation}
		\stochOneminusCosmaxInterp(\spatialvecSelect) \indep \stochMagnitudeInterp(\spatialvecSelect) \bigm| \normtwoStochMagnitudeInpimg^2 \qqand \stochOneminusCosmaxInterp(\spatialvecSelect) \indep \normtwoStochMagnitudeInpimg^2.
	\label{eq:condindep}
	\end{equation}
	Additionally, we introduce
	\begin{equation}
		\normtwoStochDeltaOutimgCmodInpimg := \normtwo{\deltaopSubGrid\inpimg},
	\end{equation}
	where $\deltaopSubGrid\inpimg$ is defined in \eqref{eq:delta_outimg_cmod_stoch}. Then, using the linearity of $\Expval$, we get
	\begin{align*}
		\ExpvalDeltaOutimgCmodCondMagnitude
			&= \sumoverVectorindices \condexpval{
				\deltaopSubGrid[\vectorindex]^2
			}{
				\normtwoStochMagnitudeInpimg^2 = \evalEnergyOutimgCmod
			} \\
			&= \sumoverVectorindices \condexpval{
				\cmodopSubSelect\inpimg[\vectorindex]^2 \, \bigl(
					1 - \stochCosmaxInterp(\spatialvecSelect)
				\bigr)^2
			}{
				\normtwoStochMagnitudeInpimg^2 = \evalEnergyOutimgCmod
			} \\
			&= \sumoverVectorindices \condexpval{
				\stochMagnitudeInterp(\spatialvecSelect)^2 \, \stochOneminusCosmaxInterp(\spatialvecSelect)^2
			}{
				\normtwoStochMagnitudeInpimg^2 = \evalEnergyOutimgCmod
			} \quad \mbox{
				(acc.\@ to \eqref{eq:modulus_complexconv_stoch} and \eqref{eq:oneminuscos_stoch})
			} \\
			&= \sumoverVectorindices  \condexpval{
				\stochMagnitudeInterp(\spatialvecSelect)^2
			}{
				\normtwoStochMagnitudeInpimg^2 = \evalEnergyOutimgCmod
			} \, \Expval\bigl[
				\stochOneminusCosmaxInterp(\spatialvecSelect)^2
			\bigr] \quad \mbox{
				(acc.\@ to \eqref{eq:condindep}).
			}
	\end{align*}
	Using the monotone convergence theorem, we get
	\begin{equation}
		\ExpvalDeltaOutimgCmodCondMagnitude = \condexpval{
			\sumoverVectorindices\stochMagnitudeInterp(\spatialvecSelect)^2
		}{
			\normtwoStochMagnitudeInpimg^2 = \evalEnergyOutimgCmod
		} \, \Expval\bigl[
			\stochOneminusCosmaxInterp(\spatialvecSelect)^2
		\bigr].
	\end{equation}
	According to \eqref{eq:modulus_complexconv_stoch} and \eqref{eq:norm_outimg_cmod}, we have
	\begin{equation}
		\sumoverVectorindices \stochMagnitudeInterp(\spatialvecSelect)^2 = \bignormtwo{\detailCmodimg}^2 = \normtwoStochMagnitudeInpimg^2.
	\end{equation}
	Therefore, we get
	\begin{align*}
		\ExpvalDeltaOutimgCmodCondMagnitude
			&= \condexpval{
				\normtwoStochMagnitudeInpimg^2
			}{
				\normtwoStochMagnitudeInpimg^2 = \evalEnergyOutimgCmod
			} \, \Expval\bigl[
				\stochOneminusCosmaxInterp(\spatialvecSelect)^2
			\bigr] \\
			&= \evalEnergyOutimgCmod \cdot \Expval\bigl[
				\stochOneminusCosmaxInterp(\spatialvecSelect)^2
			\bigr].
	\end{align*}
	Under \cref{hyp:uniformdist}, \cref{prop:expval_oneminuscos} yields
	\begin{equation}
		\ExpvalDeltaOutimgCmodCondMagnitude
			= \evalEnergyOutimgCmod \cdot \evalProxyfunRmaxCmod^2.
	\end{equation}
	Moreover, we can reformulate $\avgStochOneminusCosmaxInpimg$ such as defined in \eqref{eq:normdelta_cmod_stoch}: $\avgStochOneminusCosmaxInpimg = \normtwoStochDeltaOutimgCmodInpimg / \normtwoStochMagnitudeInpimg$. Therefore,
	\begin{equation}
		\ExpvalDiffRmaxCmodCondMagnitude
			= \frac1{\evalEnergyOutimgCmod} \, \ExpvalDeltaOutimgCmodCondMagnitude = \evalProxyfunRmaxCmod^2.
	\label{eq:condexpval_diffrmaxcmod}
	\end{equation}
	According to \eqref{eq:condexpval_diffrmaxcmod}, the conditional expected value of $\avgStochOneminusCosmaxInpimg^2$ remains the same whatever the outcome of $\normtwoStochMagnitudeInpimg^2$. Thus, the law of total expectation states that
	\begin{equation}
		\ExpvalOneminusCosmax = \Expval\left[\Expval\bigl[\avgStochOneminusCosmaxInpimg^2 \bigm| \normtwoStochMagnitudeInpimg^2\bigr]\right] = \evalProxyfunRmaxCmod^2.
	\end{equation}

	Since we have assumed \cref{hyp:bound_sumofdiffs}, we can apply \cref{prop:diffmodpool}. Using the definition of $\avgStochDiffRmaxCmodInpimg$ \eqref{eq:normdiff_cmodrmax_stoch} and $\avgStochOneminusCosmaxInpimg$ \eqref{eq:normdelta_cmod_stoch}, we get
	\begin{equation}
		\avgStochDiffRmaxCmodInpimg \leq \avgStochOneminusCosmaxInpimg + \evalLowerboundDiffRmaxCmod.
	\end{equation}
	Then,
	\begin{equation}
		\ExpvalDiffRmaxCmod \leq \ExpvalOneminusCosmax + 2\evalLowerboundDiffRmaxCmod \, \Expval\bigl[
			\avgStochOneminusCosmaxInpimg
		\bigr] + \evalLowerboundDiffRmaxCmod^2.
	\end{equation}
	According to Jensen's inequality,
	\begin{equation}
		\Expval\bigl[
			\avgStochOneminusCosmaxInpimg
		\bigr] \leq \sqrt{\ExpvalOneminusCosmax} = \evalProxyfunRmaxCmod.
	\end{equation}
	Thus,
	\begin{equation}
		\ExpvalDiffRmaxCmod \leq \evalProxyfunRmaxCmod^2 + 2\evalLowerboundDiffRmaxCmod\evalProxyfunRmaxCmod + \evalLowerboundDiffRmaxCmod^2,
	\end{equation}
	which yields \eqref{eq:scdmoment_normdiff_cmodrmax}.
\end{proof}

\subsection{Proof of \cref{lemma:hyps_tranls}}
\label{subsec:appendix_proofs_lemma_hyps_tranls}

\begin{proof}
	First, we show that, for any $\spatialvec \in \mathR^2$,
	\begin{align}
		\stochMagnitudeTranslInterp(\spatialvec)
			&= \translSamplintervVec\stochMagnitudeInterp(\spatialvec);
	\label{eq:commut_interp_transl_magnitude}\\
		\stochComplexPhaseTranslInterp(\spatialvec)
			&= \translSamplintervVec\stochComplexPhaseInterp(\spatialvec).
	\label{eq:commut_interp_transl_complexphase}
	\end{align}
	According to \cref{lemma:commut_interp_transl}, and since the convolution product commutes with translations, we have
	\begin{equation}
		\bigl(
			\inpfunAstWaveletTranslInterp
		\bigr)(\spatialvec) = \translSamplintervVec\bigl(
			\inpfunAstWaveletInterp
		\bigr)(\spatialvec).
	\end{equation}
	Then, using \eqref{eq:magnitude_phase_stoch}, the above expression becomes
	\begin{equation}
		\stochMagnitudeTranslInterp(\spatialvec) \times \stochComplexPhaseTranslInterp(\spatialvec) = (\translSamplintervVec\stochMagnitudeInterp)(\spatialvec) \times (\translSamplintervVec\stochComplexPhaseInterp)(\spatialvec),
	\end{equation}
	which yields \eqref{eq:commut_interp_transl_magnitude} and \eqref{eq:commut_interp_transl_complexphase}, by uniqueness of the magnitude-phase decomposition.
	Finally, we remind that 
	\begin{equation}
		\translSamplintervVec\stochMagnitudeInterp(\spatialvec) = \stochMagnitudeInterp(\spatialvec - \samplinterv\translvecDiscrete)
		\qqand
		\translSamplintervVec\stochComplexPhaseInterp(\spatialvec) = \stochComplexPhaseInterp(\spatialvec - \samplinterv\translvecDiscrete).
	\end{equation}
	Then, considering hypotheses \cref{hyp:uniformdist,hyp:indep_phase_modulus} with $\spatialvec \leftarrow \spatialvec - \samplinterv\translvecDiscrete$ yields the result.
\end{proof}

\subsection{Proof of \cref{th:shiftinvariance_rmax}}
\label{subsec:appendix_proofs_th_shiftinvariance_rmax}

\begin{proof}
	Using the triangle inequality, we compute
	\begin{multline}
		\bignormtwo{\rmaxopSubGrid(\translInpimg) - \detailRmaximg} \\
		\leq \bignormtwo{\cmodopSub(\translInpimg)} \, \avgStochDiffRmaxCmodTranslInpimg + \bignormtwo{\detailCmodimg} \, \avgStochDiffRmaxCmodInpimg
		+ \bignormtwo{\cmodopSub(\translInpimg) - \detailCmodimg},
	\label{eq:shiftinvariance_rmax_proof}
	\end{multline}
	where $\avgStochDiffRmaxCmodInpimg$ and $\avgStochDiffRmaxCmodTranslInpimg$ are defined in \eqref{eq:normdiff_cmodrmax_stoch}. According to \eqref{eq:condition_bandwidth}, we can apply \cref{prop:shiftinvariance_normcmod} on the first term of \eqref{eq:shiftinvariance_rmax_proof}:
	\begin{equation}
		\bignormtwo{\cmodopSub(\translInpimg)} = \bignormtwo{\detailCmodimg}.
	\end{equation}
	Moreover, we can apply \cref{th:shiftinvariance_cmod} to the third term of \eqref{eq:shiftinvariance_rmax_proof}:
	\begin{equation}
		\bignormtwo{\cmodopSub (\translInpimg) - \cmodopSub \inpimg} \leq \evalShiftinvCmod \, \bignormtwo{\cmodopSub \inpimg}.
	\end{equation}
	We therefore get
	\begin{equation}
		\bignormtwo{\rmaxopSubGrid(\translInpimg) - \detailRmaximg}
		\leq \left[\avgStochDiffRmaxCmodTranslInpimg + \avgStochDiffRmaxCmodInpimg + \evalShiftinvCmod\right] \bignormtwo{\detailCmodimg}.
	\end{equation}
		Then, by linearity of $\Expval$, we get
	\begin{equation}
		\ExpvalDiffShiftRmax \leq \Expval\bigl[\avgStochDiffRmaxCmodTranslInpimg\bigr] + \Expval\bigl[\avgStochDiffRmaxCmodInpimg\bigr] + \evalShiftinvCmod,
	\label{eq:expval_normdiff_rmax_transl_0}
	\end{equation}
	where $\avgStochDiffShiftRmaxInpimg$ has been introduced in \eqref{eq:normdiff_rmax_transl_stoch}.

	For any stochastic process $\inpimg'$ satisfying \cref{hyp:uniformdist,hyp:indep_phase_modulus}, \cref{th:scdmoment_normdiff_cmodrmax} and Jensen's inequality yield:
	\begin{equation}
		\Expval\bigl[\avgStochDiffRmaxCmodInpimgbis\bigr] \leq \evalLowerboundDiffRmaxCmod + \evalProxyfunRmaxCmod.
	\label{eq:expval_normdiff_cmodrmax}
	\end{equation}
	According to \cref{lemma:hyps_tranls}, \cref{hyp:uniformdist,hyp:indep_phase_modulus} are also satisfied for $\inpimg \leftarrow \translInpimg$. Therefore, \eqref{eq:expval_normdiff_cmodrmax} is valid for both $\inpimg' \leftarrow \inpimg$ and $\inpimg' \leftarrow \translInpimg$, and plugging it into \eqref{eq:expval_normdiff_rmax_transl_0} concludes the proof.
\end{proof}

\section{Appendix -- Theoretical Foundations for our Hypotheses}
\label{sec:appendix_invariance_hyps}

In this section, we provide theoretical arguments for justifying \cref{hyp:uniformdist,hyp:indep_phase_modulus}. As will be discussed, these hypotheses rely on some degree of shift invariance for input images, which implies the notions of stationarity and phase-shift-equivariance for stochastic processes.

\subsection{Stationary and Local Stationarity}

Given $\nevalpoints \in \nonzeroMathN$, we define \emph{$\nevalpoints$-th order stationarity} of a given stochastic process $\stochInpfun$ as stated by \citet[p.~152]{Park2018}: For any $\nevalpoints' \leq \nevalpoints$, $(\spatialvec_0,\, \dots,\, \spatialvec_{\nevalpoints' - 1}) \in (\mathR^2)^{\nevalpoints'}$ and $\translvecContinuous \in \mathR^2$, the joint distribution of $\bigl(
	\stochInpfun(\spatialvec_0),\, \dots,\, \stochInpfun(\spatialvec_{\nevalpoints' - 1})
\bigr)$ is identical to the one of $\bigl(
	\transl_{\translvecContinuous}\stochInpfun(\spatialvec_0),\, \dots,\, \transl_{\translvecContinuous}\stochInpfun(\spatialvec_{\nevalpoints' - 1})
\bigr)$. Additionally, \emph{strict-sense stationarity} is defined as $\nevalpoints$-th order stationarity for any $\nevalpoints \in \nonzeroMathN$.

Strict-sense stationarity suggests that any translated version of a given image is equally likely. However, this property is seldom fully achieved for images \citep{Tygert2016}. First, $\inpimg$ has fixed boundaries. Consequently, any realization of $\stochInpfunInterp(\spatialvec)$ quickly vanishes as $\norm{\spatialvec}$ tends to $\infty$. Furthermore, depending on which category the image belongs to, the pixel distribution is likely to vary across various regions. For instance, we can expect the main subject to be located at the center of the image. We refer readers to \citet{Torralba2003} for more details on statistical properties of images from natural versus man-made objects.
For this reason, we introduce a notion of local stationarity as follows.
\begin{definition}
	\label{def:locallystationary}
	Given $\localityparam \geq 0$, a stochastic process $\stochInpfun$ is \emph{$\localityparam$-locally stationary} to the $\nevalpoints$-th order if, for any $0 < \nevalpoints'' \leq \nevalpoints' \leq \nevalpoints$, any $(\spatialvec_0,\, \dots,\, \spatialvec_{\nevalpoints' - 1}) \in (\mathR^2)^{\nevalpoints'}$, any displacement vector $\translvecContinuous \in \mathR^2$, and any pair of measurable sets $\measurablesetRealmultaxes,\, \measurablesetRealmultaxesbis \subset \mathR^{\nevalpoints''} \times \mathR^{\nevalpoints' - \nevalpoints''}$,
	\begin{multline}
		\left|\Bigcondproba{
			\Bigl(
				\transl_{\translvecContinuous}\stochInpfun(\spatialvec_0),\, \dots,\, \transl_{\translvecContinuous}\stochInpfun(\spatialvec_{\nevalpoints'' - 1})
			\Bigr)^\top \in \measurablesetRealmultaxes
		}{
			\Bigl(
				\transl_{\translvecContinuous}\stochInpfun(\spatialvec_{\nevalpoints''}),\, \dots,\, \transl_{\translvecContinuous}\stochInpfun(\spatialvec_{\nevalpoints' - 1})
			\Bigr)^\top \in \measurablesetRealmultaxesbis
		} \right. \\
		\left. - \Bigcondproba{
			\Bigl(
				\stochInpfun(\spatialvec_0),\, \dots,\, \stochInpfun(\spatialvec_{\nevalpoints'' - 1})
			\Bigr)^\top \in \measurablesetRealmultaxes
		}{
			\Bigl(
				\stochInpfun(\spatialvec_{\nevalpoints''}),\, \dots,\, \stochInpfun(\spatialvec_{\nevalpoints' - 1})
			\Bigr)^\top \in \measurablesetRealmultaxesbis
		}\right| \leq \localityparam\normtwo{\translvecContinuous}.
	\label{eq:locallystationary}
	\end{multline}
	Moreover, $\stochInpfun$ is \emph{strict-sense $\localityparam$-locally stationary} if this property is satisfied for any $\nevalpoints \in \mathN$.
\end{definition}

\begin{remark}
	The special case where $\localityparam = 0$ corresponds to the standard definition of $\nevalpoints$-th order stationarity. Additionally, the use of conditional probabilities is essential for defining local stationarity, as it helps prevent instabilities that may arise when conditioning on low-probability events.
\end{remark}

In what follows, we consider the following narrowband stochastic process:
\begin{equation}
	\stochHighfreqfunInterp: \spatialvec \mapsto (\stochInpfunAstWaveletInterp)(\spatialvec).
\end{equation}
We assume that $\stochInpfunInterp$, and therefore $\stochHighfreqfunInterp$, is nearly shift-invariant for displacement vectors that are much smaller than the image ``characteristic'' size in the continuous domain, which is equal to $\samplinterv\imgsize$, where, as a reminder, $\imgsize \in \mathN$ denotes the support size of input images and $\samplinterv > 0$ denotes the sampling interval. More formally, $\stochHighfreqfunInterp$ is assumed to be strict-sense $\localityparam$-locally stationary with
\begin{equation}
	\localityparam := \frac1{\samplinterv\imgsize}.
\label{eq:localityparam}
\end{equation}

\subsection{Translations and Phase Shifts}

We consider the following stochastic processes:
\begin{equation}
	\stochLowfreqfunInterp: \spatialvec \mapsto (\stochInpfunAstWaveletInterp)(\spatialvec) \, \rme^{i\innerprod{\freqvec}{\spatialvec}}
	\qqand
	\stochHighfreqfunInterp: \spatialvec \mapsto (\stochInpfunAstWaveletInterp)(\spatialvec).
\end{equation}
To justify our hypotheses, we also need to characterize $\transl_{\translvecContinuous}\stochHighfreqfunInterp$ as a function of $\stochHighfreqfunInterp$. Having:
\begin{equation}
	\transl_{\translvecContinuous}\stochHighfreqfunInterp(\spatialvec)
		= \transl_{\translvecContinuous}\stochLowfreqfunInterp(\spatialvec) \, \rme^{-i\innerprod{\freqvec}{\spatialvec}} \, \rme^{i\innerprod{\freqvec}{\translvecContinuous}},
\end{equation}
we get
\begin{equation}
	\bigl|
		\transl_{\translvecContinuous}\stochHighfreqfunInterp(\spatialvec) - \stochHighfreqfunInterp(\spatialvec) \, \rme^{i\innerprod{\freqvec}{\translvecContinuous}}
	\bigr| = \bigl|
		\transl_{\translvecContinuous}\stochLowfreqfunInterp(\spatialvec) - \stochLowfreqfunInterp(\spatialvec)
	\bigr|.
\end{equation}
According to \cref{lemma:lowfreqfun}, the support of $\fourierStochLowfreqfunInterp$ is contained within the ball $\linftyBall\left(
	\frac{\supportsizeDiscrete}{2\samplinterv}
\right)$. Intuitively, we can define a ``minimal wavelength'' $\wavelength := 2\pi\samplinterv / \supportsizeDiscrete$ such that, if $\normtwo{\translvecContinuous} \ll \wavelength$, we can approximate $\transl_{\translvecContinuous}\stochLowfreqfunInterp(\spatialvec) \approx \stochLowfreqfunInterp(\spatialvec)$, and therefore
\begin{equation}
	\transl_{\translvecContinuous}\stochHighfreqfunInterp(\spatialvec) \approx \stochHighfreqfunInterp(\spatialvec) \, \rme^{i\innerprod{\freqvec}{\translvecContinuous}}.
\end{equation}
If the two terms were strictly identical for any $\spatialvec \in \mathR^2$, we would get, for any $\nevalpoints \in \mathN$, $(\spatialvec_0,\, \dots,\, \spatialvec_{\nevalpoints}) \in (\mathR^2)^{\nevalpoints}$, and any measurable set $\measurablesetRealmultaxes \subset \mathR^{\nevalpoints}$,
\begin{equation}
	\bigl(
		\stochHighfreqfunInterp(\spatialvec_0),\, \dots,\, \stochHighfreqfunInterp(\spatialvec_{\nevalpoints - 1})
	\bigr)^\top \in \measurablesetUnitcircle
		\iif \bigl(
			\transl_{\translvecContinuous}\stochHighfreqfunInterp(\spatialvec_0),\, \dots,\, \transl_{\translvecContinuous}\stochHighfreqfunInterp(\spatialvec_{\nevalpoints - 1})
		\bigr)^\top \in \expiFreqvecTranslvec \measurablesetUnitcircle,
\end{equation}
and therefore,
\begin{equation}
	\Proba\left\{
		\bigl(
			\stochHighfreqfunInterp(\spatialvec_0),\, \dots,\, \stochHighfreqfunInterp(\spatialvec_{\nevalpoints - 1})
		\bigr)^\top \in \measurablesetUnitcircle
	\right\} = \Proba\left\{
		\bigl(
			\transl_{\translvecContinuous}\stochHighfreqfunInterp(\spatialvec_0),\, \dots,\, \transl_{\translvecContinuous}\stochHighfreqfunInterp(\spatialvec_{\nevalpoints - 1})
		\bigr)^\top \in \expiFreqvecTranslvec \measurablesetUnitcircle
	\right\}.
\label{eq:probaequality}
\end{equation}
Instead, we relax the above equality \eqref{eq:probaequality} and introduce the following definition.
\begin{definition}
	\label{def:locallypsequivariant}
	Given $\localityparam \geq 0$, a stochastic process $\stochInpfun$ is \emph{$\localityparam$-locally phase-shift-equivariant} to the $\nevalpoints$-th order with respect to the frequency vector $\freqvec \in \mathR^2$ if, for any $0 < \nevalpoints'' \leq \nevalpoints' \leq \nevalpoints$, any $(\spatialvec_0,\, \dots,\, \spatialvec_{\nevalpoints' - 1}) \in (\mathR^2)^{\nevalpoints'}$, any displacement vector $\translvecContinuous \in \mathR^2$, and any pair of measurable sets $\measurablesetRealmultaxes,\, \measurablesetRealmultaxesbis \subset \mathR^{\nevalpoints''} \times \mathR^{\nevalpoints' - \nevalpoints''}$,
	\begin{multline}
		\left|\Bigcondproba{
			\Bigl(
				\transl_{\translvecContinuous}\stochInpfun(\spatialvec_0),\, \dots,\, \transl_{\translvecContinuous}\stochInpfun(\spatialvec_{\nevalpoints'' - 1})
			\Bigr)^\top \in \expiFreqvecTranslvec\measurablesetRealmultaxes
		}{
			\Bigl(
				\transl_{\translvecContinuous}\stochInpfun(\spatialvec_{\nevalpoints''}),\, \dots,\, \transl_{\translvecContinuous}\stochInpfun(\spatialvec_{\nevalpoints' - 1})
			\Bigr)^\top \in \expiFreqvecTranslvec\measurablesetRealmultaxesbis
		}\right. \\
		\left. - \Bigcondproba{
			\Bigl(
				\stochInpfun(\spatialvec_0),\, \dots,\, \stochInpfun(\spatialvec_{\nevalpoints'' - 1})
			\Bigr)^\top \in \measurablesetRealmultaxes
		}{
			\Bigl(
				\stochInpfun(\spatialvec_{\nevalpoints''}),\, \dots,\, \stochInpfun(\spatialvec_{\nevalpoints' - 1})
			\Bigr)^\top \in \measurablesetRealmultaxesbis
		}
		\right| \leq \localityparam\normtwo{\translvecContinuous}.
	\label{eq:localphaseshiftequivariance}
	\end{multline}
	Moreover, $\stochInpfun$ is \emph{strict-sense $\localityparam$-locally phase-shift-equivariant} if this property is satisfied for any $\nevalpoints \in \mathN$.
\end{definition}

In what follows, we will assume that the stochastic process $\stochHighfreqfunInterp$ is nearly phase-shift-equivariant for displacement vectors that are much smaller than the minimal wavelength $\wavelength$. More formally, $\stochHighfreqfunInterp$ is assumed to be strict-sense $\localityparam'$-locally phase-shift-equivariant with
\begin{equation}
	\localityparam' := 1/\wavelength = \frac{\supportsizeDiscrete}{2\pi\samplinterv}.
\label{eq:localityparambis}
\end{equation}

\subsection{Justification for \cref{hyp:uniformdist}}
\label{subsec:appendix_invariance_hyps_1}

We then consider the following proposition, which states that the probability measure of $\stochComplexPhaseInterp(\spatialvec)$ is nearly invariant with respect to phase shifts.
\begin{proposition}
	\label{prop:uniformdist}
	We assume that $\stochHighfreqfunInterp$ is $\localityparam$-locally stationary (\cref{def:locallystationary}) and $\localityparam'$-locally phase-shift-equivariant with respect to $\freqvec$ (\cref{def:locallypsequivariant}), both to the first order. Then, for any measurable set $\measurablesetUnitcircle \subset \unitcircle$,
	\begin{equation}
		\forall \anglevalbis \in \zerotwopi,\, \left|
			\probaMeasure(\measurablesetUnitcircle) - \probaMeasure(\rme^{i\anglevalbis} \measurablesetUnitcircle)
		\right| \leq 2\pi(\localityparam + \localityparam') / \normtwo{\freqvec}.
	\label{eq:phaseshiftinvariance}
	\end{equation}
	where $\probaMeasure: \measurablesetUnitcircle \mapsto \Proba\left\{
		\stochComplexPhaseInterp(\spatialvec) \in \measurablesetUnitcircle
	\right\}$ denotes the probability measure of $\stochComplexPhaseInterp(\spatialvec)$.
\end{proposition}

\begin{proof}
	Let $\anglevalbis \in \zerotwopi$. We compute:
	\begin{align*}
		\left|
			\probaMeasure(\measurablesetUnitcircle) - \probaMeasure(\rme^{i\anglevalbis} \measurablesetUnitcircle)
		\right|
			&= \left|
				\Proba\left\{
					\stochComplexPhaseInterp(\spatialvec) \in \measurablesetUnitcircle
				\right\} - \Proba\left\{
					\stochComplexPhaseInterp(\spatialvec) \in \rme^{i\anglevalbis}\measurablesetUnitcircle
				\right\}
			\right| \\
			&= \left|
				\Proba\bigl\{
					\stochComplexPhaseInterp(\spatialvec) \in \measurablesetUnitcircle
				\bigr\} - \Proba\bigl\{
					\stochComplexPhaseInterp(\spatialvec) \in \rme^{i\innerprod{\freqvec}{\translvecContinuous}}\measurablesetUnitcircle
				\bigr\}
			\right|,
	\end{align*}
	where we have denoted $\translvecContinuous := \anglevalbis \, \freqvec / \normtwo{\freqvec}^2$. Then, using the triangle inequality,
	\begin{align*}
		\left|
			\probaMeasure(\measurablesetUnitcircle) - \probaMeasure(\rme^{i\anglevalbis} \measurablesetUnitcircle)
		\right|
			&\leq \left|
				\Proba\bigl\{
					\stochComplexPhaseInterp(\spatialvec) \in \measurablesetUnitcircle
				\bigr\} - \Proba\bigl\{
					\transl_{\translvecContinuous}\stochComplexPhaseInterp(\spatialvec) \in \rme^{i\innerprod{\freqvec}{\translvecContinuous}}\measurablesetUnitcircle
				\bigr\}
			\right| \\
			&\qquad\qquad + \left|
				\Proba\bigl\{
					\stochComplexPhaseInterp(\spatialvec) \in \rme^{i\innerprod{\freqvec}{\translvecContinuous}}\measurablesetUnitcircle
				\bigr\} - \Proba\bigl\{
					\transl_{\translvecContinuous}\stochComplexPhaseInterp(\spatialvec) \in \rme^{i\innerprod{\freqvec}{\translvecContinuous}}\measurablesetUnitcircle
				\bigr\}
			\right| \\
			&= \left|
				\Proba\bigl\{
					\stochHighfreqfunInterp(\spatialvec) \in \complexphasefun^{-1}(\measurablesetUnitcircle)
				\bigr\} - \Proba\bigl\{
					\transl_{\translvecContinuous}\stochHighfreqfunInterp(\spatialvec) \in \complexphasefun^{-1}\bigl(
						\rme^{i\innerprod{\freqvec}{\translvecContinuous}}\measurablesetUnitcircle
					\bigr)
				\bigr\}
			\right| \\
			&\qquad\qquad + \left|
				\Proba\bigl\{
					\stochHighfreqfunInterp(\spatialvec) \in \complexphasefun^{-1}\bigl(
						\rme^{i\innerprod{\freqvec}{\translvecContinuous}}\measurablesetUnitcircle
					\bigr)
				\bigr\} - \Proba\bigl\{
					\transl_{\translvecContinuous}\stochHighfreqfunInterp(\spatialvec) \in \complexphasefun^{-1}\bigl(
						\rme^{i\innerprod{\freqvec}{\translvecContinuous}}\measurablesetUnitcircle
					\bigr)
				\bigr\}
			\right|,
	\end{align*}
	where we have denoted $\complexphasefun: z \mapsto \rme^{i\angle(z)}$. We can easily show that $\complexphasefun^{-1}\bigl(
		\rme^{i\innerprod{\freqvec}{\translvecContinuous}}\measurablesetUnitcircle
	\bigr) = \rme^{i\innerprod{\freqvec}{\translvecContinuous}} \complexphasefun^{-1}(\measurablesetUnitcircle)$. Therefore, the above expression can be re-written:
	\begin{multline}
		\left|
			\probaMeasure(\measurablesetUnitcircle) - \probaMeasure(\rme^{i\anglevalbis} \measurablesetUnitcircle)
		\right| \leq \left|
			\Proba\bigl\{
				\stochHighfreqfunInterp(\spatialvec) \in \measurablesetRealaxis
			\bigr\} - \Proba\bigl\{
				\transl_{\translvecContinuous}\stochHighfreqfunInterp(\spatialvec) \in \rme^{i\innerprod{\freqvec}{\translvecContinuous}} \measurablesetRealaxis
			\bigr\}
		\right| \\
		+ \left|
			\Proba\bigl\{
				\stochHighfreqfunInterp(\spatialvec) \in \measurablesetRealaxis'
			\bigr\} - \Proba\bigl\{
				\transl_{\translvecContinuous}\stochHighfreqfunInterp(\spatialvec) \in \measurablesetRealaxis'
			\bigr\}
		\right|,
	\label{eq:uniformdist_proof}
	\end{multline}
	where we have denoted $\measurablesetRealaxis := \complexphasefun^{-1}(\measurablesetUnitcircle)$ and $\measurablesetRealaxis' := \complexphasefun^{-1}\bigl(
		\rme^{i\innerprod{\freqvec}{\translvecContinuous}}\measurablesetUnitcircle
	\bigr)$. Then, by applying \eqref{eq:locallystationary} and \eqref{eq:localphaseshiftequivariance} with $\nevalpoints' = 1$ to \eqref{eq:uniformdist_proof}, we get:
	\begin{equation}
		\left|
			\probaMeasure(\measurablesetUnitcircle) - \probaMeasure(\rme^{i\anglevalbis} \measurablesetUnitcircle)
		\right| \leq (\localityparam + \localityparam') \normtwo{\translvecContinuous}.
	\end{equation}
	Finally, using the definition of $\translvecContinuous$ and bounding $\anglevalbis$ by $2\pi$ yields the result.
\end{proof}

If we replace $\localityparam$ and $\localityparam'$ in \eqref{eq:phaseshiftinvariance} by their respective values in \eqref{eq:localityparam} and \eqref{eq:localityparambis}, we get,
\begin{equation}
	\forall \anglevalbis \in \zerotwopi,\, \left|
		\probaMeasure(\measurablesetUnitcircle) - \probaMeasure(\rme^{i\anglevalbis} \measurablesetUnitcircle)
	\right| \leq \frac{2\pi}{\samplinterv\imgsize} \times \normtwo{\freqvec}^{-1} + \frac{\supportsizeDiscrete}{\samplinterv} \times \normtwo{\freqvec}^{-1}.
\end{equation}
We recall that $\freqvec := \freqvecMpipi / \samplinterv$. Then, by applying the constraints stated in \eqref{eq:condition_characfreq} and \eqref{eq:condition_characfreq_fourierwindow}, we get
\begin{equation}
	\forall \anglevalbis \in \zerotwopi,\, \left|
		\probaMeasure(\measurablesetUnitcircle) - \probaMeasure(\rme^{i\anglevalbis} \measurablesetUnitcircle)
	\right| \ll 1.
\end{equation}
Therefore, $\probaMeasure$ is almost invariant to phase shifts. Since the only probability measure satisfying the phase-shift invariance property is the uniform probability measure,\footnote{
	Any probability measure defined on $\unitcircle$ is a Radon measure. Therefore, according to Haar's theorem \citep{Halmos2013}, there exists a unique probability measure on $\unitcircle$ satisfying the phase-shift invariance property, and it turns out that the uniform probability measure is one such candidate.
}
we deduce that $\stochComplexPhaseInterp(\spatialvec)$ follows a near-uniform distribution on $\unitcircle$. For the sake of simplicity, in \cref{hyp:uniformdist} we have assumed a strictly-uniform distribution.

\subsection{Justification for \cref{hyp:indep_phase_modulus}}
\label{subsec:appendix_invariance_hyps_2}

Let $\nevalpoints \in \nonzeroMathN$ and $\spatialvec,\, \spatialvecbis_0,\, \dots,\, \spatialvecbis_{\nevalpoints-1} \in \mathR^2$. To simplify notations, we consider the random vector
\begin{equation}
	\randMagnitudeMulteval := \bigl(
		\stochMagnitudeInterp(\spatialvecbis_0),\, \dots,\, \stochMagnitudeInterp(\spatialvecbis_{\nevalpoints - 1})
	\bigr)^\top
\end{equation}
with outcomes in $\mathR_+^\nevalpoints$.
This section is organized as follows. Using reasoning similar to that in \cref{prop:uniformdist}, we show that, for any measurable subset $\measurablesetRealposmultaxes \subset \mathR_+^\nevalpoints$, $\stochComplexPhaseInterp$ follows a near-uniform probability distribution conditionally to $\randMagnitudeMulteval \in \measurablesetRealposmultaxes$. Since we already assumed that $\stochComplexPhaseInterp$ follows an unconditional uniform distribution, we deduce that $\stochComplexPhaseInterp$ and $\randMagnitudeMulteval$ are nearly independent.

\begin{proposition}
	\label{prop:indep_phase_modulus}
	We assume that $\stochHighfreqfunInterp$ is $\localityparam$-locally stationary (\cref{def:locallystationary}) and $\localityparam'$-locally phase-shift-equivariant with respect to $\freqvec$ (\cref{def:locallypsequivariant}), both in the strict sense. Then, for any measurable sets $\measurablesetUnitcircle \subset \unitcircle$ and $\measurablesetRealposmultaxes \subset \mathR_+^{\nevalpoints}$,
	\begin{equation}
		\forall \anglevalbis \in \zerotwopi,\, \left|
			\probaMeasure_{\measurablesetRealposmultaxes}(\measurablesetUnitcircle) - \probaMeasure_{\measurablesetRealposmultaxes}(\rme^{i\anglevalbis} \measurablesetUnitcircle)
		\right| \leq 2\pi(\localityparam + \localityparam') / \normtwo{\freqvec}.
	\label{eq:condphaseshiftinvariance}
	\end{equation}
	where $\probaMeasure_{\measurablesetRealposmultaxes}: \measurablesetUnitcircle \mapsto \condproba{\stochComplexPhaseInterp(\spatialvec) \in \measurablesetUnitcircle}{\randMagnitudeMulteval \in \measurablesetRealposmultaxes}$ denotes the conditional probability measure of $\stochComplexPhaseInterp(\spatialvec)$.
\end{proposition}

\begin{proof}
	Let $\anglevalbis \in \zerotwopi$. We compute:
	\begin{align*}
		\left|
			\probaMeasure_{\measurablesetRealposmultaxes}(\rme^{i\anglevalbis} \measurablesetUnitcircle) - \probaMeasure_{\measurablesetRealposmultaxes}(\measurablesetUnitcircle)
		\right|
			&= \left|
			\bigcondproba{
				\stochComplexPhaseInterp(\spatialvec) \in \rme^{i\anglevalbis}\measurablesetUnitcircle
			}{
				\randMagnitudeMulteval \in \measurablesetRealposmultaxes
			} -
			\bigcondproba{
				\stochComplexPhaseInterp(\spatialvec) \in \measurablesetUnitcircle
			}{
				\randMagnitudeMulteval \in \measurablesetRealposmultaxes
			}
			\right| \\
			&= \left|
			\bigcondproba{
				\stochComplexPhaseInterp(\spatialvec) \in \rme^{i\innerprod{\freqvec}{\translvecContinuous}}\measurablesetUnitcircle
			}{
				\randMagnitudeMulteval \in \measurablesetRealposmultaxes
			} -
			\bigcondproba{
				\stochComplexPhaseInterp(\spatialvec) \in \measurablesetUnitcircle
			}{
				\randMagnitudeMulteval \in \measurablesetRealposmultaxes
			}
			\right|,
	\end{align*}
	where we have denoted
	\begin{equation}
		\translvecContinuous := \anglevalbis \, \freqvec / \normtwo{\freqvec}^2
	\label{eq:indep_phase_modulus_proof_translvec}
	\end{equation}
	Then, using the triangle inequality,
	\begin{multline}
		\left|
			\probaMeasure_{\measurablesetRealposmultaxes}(\rme^{i\anglevalbis} \measurablesetUnitcircle) - \probaMeasure_{\measurablesetRealposmultaxes}(\measurablesetUnitcircle)
		\right| \\
		\leq \left|
		\bigcondproba{
			\transl_{\translvecContinuous}\stochComplexPhaseInterp(\spatialvec) \in \rme^{i\innerprod{\freqvec}{\translvecContinuous}}\measurablesetUnitcircle
		}{
			\transl_{\translvecContinuous}\randMagnitudeMulteval \in \measurablesetRealposmultaxes
		} -
		\bigcondproba{
			\stochComplexPhaseInterp(\spatialvec) \in \rme^{i\innerprod{\freqvec}{\translvecContinuous}}\measurablesetUnitcircle
		}{
			\randMagnitudeMulteval \in \measurablesetRealposmultaxes
		}
		\right| \\
		+ \left|
		\bigcondproba{
			\transl_{\translvecContinuous}\stochComplexPhaseInterp(\spatialvec) \in \rme^{i\innerprod{\freqvec}{\translvecContinuous}}\measurablesetUnitcircle
		}{
			\transl_{\translvecContinuous}\randMagnitudeMulteval \in \measurablesetRealposmultaxes
		} -
		\bigcondproba{
			\stochComplexPhaseInterp(\spatialvec) \in \measurablesetUnitcircle
		}{
			\randMagnitudeMulteval \in \measurablesetRealposmultaxes
		}
		\right|,
	\label{eq:indep_phase_modulus_proof}
	\end{multline}
	where we have denoted
	\begin{equation}
		\transl_{\translvecContinuous}\randMagnitudeMulteval := \bigl(
			\transl_{\translvecContinuous}\stochMagnitudeInterp(\spatialvecbis_0),\, \dots,\, \transl_{\translvecContinuous}\stochMagnitudeInterp(\spatialvecbis_{\nevalpoints - 1})
		\bigr)^\top.
	\end{equation}

	Let us split this expression for the sake of readability. The first term after the $\leq$ sign can be equivalently written as follows:
	\begin{multline}
		\left|
		\bigcondproba{
			\transl_{\translvecContinuous}\stochComplexPhaseInterp(\spatialvec) \in \rme^{i\innerprod{\freqvec}{\translvecContinuous}}\measurablesetUnitcircle
		}{
			\transl_{\translvecContinuous}\randMagnitudeMulteval \in \measurablesetRealposmultaxes
		} -
		\bigcondproba{
			\stochComplexPhaseInterp(\spatialvec) \in \rme^{i\innerprod{\freqvec}{\translvecContinuous}}\measurablesetUnitcircle
		}{
			\randMagnitudeMulteval \in \measurablesetRealposmultaxes
		}
		\right| \\
		= \left|
			\bigcondproba{
				\transl_{\translvecContinuous}\stochHighfreqfunInterp(\spatialvec) \in \measurablesetRealaxis'
			}{
				\bigl(
					\transl_{\translvecContinuous}\stochHighfreqfunInterp(\spatialvecbis_0),\, \dots,\, \transl_{\translvecContinuous}\stochHighfreqfunInterp(\spatialvecbis_{\nevalpoints - 1})
				\bigr)^\top \in \measurablesetRealmultaxesbis
			}
		\right. \\
		\left. - \bigcondproba{
			\stochHighfreqfunInterp(\spatialvec) \in \measurablesetRealaxis'
		}{
			\bigl(
				\stochHighfreqfunInterp(\spatialvecbis_0),\, \dots,\, \stochHighfreqfunInterp(\spatialvecbis_{\nevalpoints - 1})
			\bigr)^\top \in \measurablesetRealmultaxesbis
		}
		\right|,
	\label{eq:indep_phase_modulus_proof_2}
	\end{multline}
	where we have denoted
	\begin{equation}
		\measurablesetRealaxis' := \complexphasefun^{-1}\bigl(
			\rme^{i\innerprod{\freqvec}{\translvecContinuous}}\measurablesetUnitcircle
		\bigr) \qqand \measurablesetRealmultaxesbis := \modulusmultfun^{-1}(\measurablesetRealposmultaxes),
	\end{equation}
	with
	\begin{equation}
		\complexphasefun: z \mapsto \rme^{i\angle(z)} \qqand \modulusmultfun: (z_0,\, \dots,\, z_{\nevalpoints - 1})^\top \mapsto (|z_0|,\, \dots,\, |z_{\nevalpoints - 1}|)^\top.
	\end{equation}
	Therefore, according the local stationarity hypothesis, we apply \eqref{eq:locallystationary} to \eqref{eq:indep_phase_modulus_proof_2}, which yields
	\begin{equation}
		\left|
		\bigcondproba{
			\transl_{\translvecContinuous}\stochComplexPhaseInterp(\spatialvec) \in \rme^{i\innerprod{\freqvec}{\translvecContinuous}}\measurablesetUnitcircle
		}{
			\transl_{\translvecContinuous}\randMagnitudeMulteval \in \measurablesetRealposmultaxes
		} -
		\bigcondproba{
			\stochComplexPhaseInterp(\spatialvec) \in \rme^{i\innerprod{\freqvec}{\translvecContinuous}}\measurablesetUnitcircle
		}{
			\randMagnitudeMulteval \in \measurablesetRealposmultaxes
		}
		\right| \leq \localityparam\normtwo{\translvecContinuous}.
	\label{eq:indep_phase_modulus_proof_A}
	\end{equation}

	Next, the second term after the $\leq$ sign in \eqref{eq:indep_phase_modulus_proof} can be equivalently written as follows:
	\begin{multline}
		\left|
		\bigcondproba{
			\transl_{\translvecContinuous}\stochComplexPhaseInterp(\spatialvec) \in \rme^{i\innerprod{\freqvec}{\translvecContinuous}}\measurablesetUnitcircle
		}{
			\transl_{\translvecContinuous}\randMagnitudeMulteval \in \measurablesetRealposmultaxes
		} -
		\bigcondproba{
			\stochComplexPhaseInterp(\spatialvec) \in \measurablesetUnitcircle
		}{
			\randMagnitudeMulteval \in \measurablesetRealposmultaxes
		}
		\right| \\
		= \left|
			\bigcondproba{
				\transl_{\translvecContinuous}\stochHighfreqfunInterp(\spatialvec) \in \complexphasefun^{-1}\bigl(
					\rme^{i\innerprod{\freqvec}{\translvecContinuous}}\measurablesetUnitcircle
				\bigr)
			}{
				\bigl(
					\transl_{\translvecContinuous}\stochHighfreqfunInterp(\spatialvecbis_0),\, \dots,\, \transl_{\translvecContinuous}\stochHighfreqfunInterp(\spatialvecbis_{\nevalpoints - 1})
				\bigr)^\top \in \modulusmultfun^{-1}(\measurablesetRealposmultaxes)
			}
		\right. \\
		\left. - \bigcondproba{
			\stochHighfreqfunInterp(\spatialvec) \in \complexphasefun^{-1}(
				\measurablesetUnitcircle
			)
		}{
			\bigl(
				\stochHighfreqfunInterp(\spatialvecbis_0),\, \dots,\, \stochHighfreqfunInterp(\spatialvecbis_{\nevalpoints - 1})
			\bigr)^\top \in \modulusmultfun^{-1}(\measurablesetRealposmultaxes)
		}
		\right|.
	\label{eq:indep_phase_modulus_proof_3}
	\end{multline}
	As explained in the proof of \cref{prop:uniformdist}, we have
	\begin{equation}
		\complexphasefun^{-1}\bigl(
			\rme^{i\innerprod{\freqvec}{\translvecContinuous}}\measurablesetUnitcircle
		\bigr) = \rme^{i\innerprod{\freqvec}{\translvecContinuous}} \complexphasefun^{-1}(\measurablesetUnitcircle).
	\end{equation}
	Moreover, we can easily show that
	\begin{equation}
		\modulusmultfun^{-1}(
			\measurablesetRealposmultaxes
		) = \rme^{i\innerprod{\freqvec}{\translvecContinuous}}\modulusmultfun^{-1}(
			\measurablesetRealposmultaxes
		).
	\end{equation}
	Therefore, \eqref{eq:indep_phase_modulus_proof_3} can be re-written:
	\begin{multline}
		\left|
		\bigcondproba{
			\transl_{\translvecContinuous}\stochComplexPhaseInterp(\spatialvec) \in \rme^{i\innerprod{\freqvec}{\translvecContinuous}}\measurablesetUnitcircle
		}{
			\transl_{\translvecContinuous}\randMagnitudeMulteval \in \measurablesetRealposmultaxes
		} -
		\bigcondproba{
			\stochComplexPhaseInterp(\spatialvec) \in \measurablesetUnitcircle
		}{
			\randMagnitudeMulteval \in \measurablesetRealposmultaxes
		}
		\right| \\
		= \left|
			\bigcondproba{
				\transl_{\translvecContinuous}\stochHighfreqfunInterp(\spatialvec) \in \rme^{i\innerprod{\freqvec}{\translvecContinuous}}\measurablesetRealaxis
			}{
				\bigl(
					\transl_{\translvecContinuous}\stochHighfreqfunInterp(\spatialvecbis_0),\, \dots,\, \transl_{\translvecContinuous}\stochHighfreqfunInterp(\spatialvecbis_{\nevalpoints - 1})
				\bigr)^\top \in \rme^{i\innerprod{\freqvec}{\translvecContinuous}}\measurablesetRealmultaxesbis
			}
		\right. \\
		\left. - \bigcondproba{
			\stochHighfreqfunInterp(\spatialvec) \in \measurablesetRealaxis
		}{
			\bigl(
				\stochHighfreqfunInterp(\spatialvecbis_0),\, \dots,\, \stochHighfreqfunInterp(\spatialvecbis_{\nevalpoints - 1})
			\bigr)^\top \in \measurablesetRealmultaxesbis
		}
		\right|,
	\label{eq:indep_phase_modulus_proof_4}
	\end{multline}
	where we have denoted
	\begin{equation}
		\measurablesetRealaxis := \complexphasefun^{-1}(
			\measurablesetUnitcircle
		) \qqand \measurablesetRealmultaxesbis := \modulusmultfun^{-1}(\measurablesetRealposmultaxes).
	\end{equation}
	Therefore, according the local phase-shift equivariance hypothesis, we apply \eqref{eq:localphaseshiftequivariance} to \eqref{eq:indep_phase_modulus_proof_4}, which yields
	\begin{equation}
		\left|
		\bigcondproba{
			\transl_{\translvecContinuous}\stochComplexPhaseInterp(\spatialvec) \in \rme^{i\innerprod{\freqvec}{\translvecContinuous}}\measurablesetUnitcircle
		}{
			\transl_{\translvecContinuous}\randMagnitudeMulteval \in \measurablesetRealposmultaxes
		} -
		\bigcondproba{
			\stochComplexPhaseInterp(\spatialvec) \in \measurablesetUnitcircle
		}{
			\randMagnitudeMulteval \in \measurablesetRealposmultaxes
		}
		\right| \leq \localityparam'\normtwo{\translvecContinuous}.
	\label{eq:indep_phase_modulus_proof_B}
	\end{equation}

	Finally, plugging \eqref{eq:indep_phase_modulus_proof_A} and \eqref{eq:indep_phase_modulus_proof_B} into \eqref{eq:indep_phase_modulus_proof}, using the definition of $\translvecContinuous$ in \eqref{eq:indep_phase_modulus_proof_translvec}, and bounding $\anglevalbis$ by $2\pi$, yields the result.
\end{proof}

By applying the constraints sated in \eqref{eq:condition_characfreq} and \eqref{eq:condition_characfreq_fourierwindow}, we get (see \cref{subsec:appendix_invariance_hyps_1}),
\begin{equation}
	\forall \anglevalbis \in \zerotwopi,\, \left|
		\probaMeasure_{\measurablesetRealposmultaxes}(\measurablesetUnitcircle) - \probaMeasure_{\measurablesetRealposmultaxes}(\rme^{i\anglevalbis} \measurablesetUnitcircle)
	\right| \ll 1.
\end{equation}
Therefore, $\stochComplexPhaseInterp(\spatialvec)$ follows a near-uniform conditional distribution on $\unitcircle$ given $\randMagnitudeMulteval \in \measurablesetRealposmultaxes$.

Furthermore, according to \cref{subsec:appendix_invariance_hyps_1}, $\stochComplexPhaseInterp(\spatialvec)$ also follows a near-uniform unconditional distribution. Therefore, we get, for any measurable sets $\measurablesetUnitcircle \subset \unitcircle$ and $\measurablesetRealposmultaxes \subset \mathR_+^\nevalpoints$,
\begin{equation}
	\bigcondproba{
		\stochComplexPhaseInterp(\spatialvec) \in \measurablesetUnitcircle
	}{
		(\randMagnitudeMulteval \in \measurablesetRealposmultaxes)
	} \approx \Proba\bigl\{
		\stochComplexPhaseInterp(\spatialvec) \in \measurablesetUnitcircle
	\bigr\}.
\end{equation}
We deduce that $\stochComplexPhaseInterp(\spatialvec)$ and $\randMagnitudeMulteval$ are nearly independent. For the sake of simplicity, in \cref{hyp:indep_phase_modulus} we have assumed strict independence.

\section{Appendix -- Details on \dtcwpt}

A description of the transform itself is provided in \cref{subsec:appendix_dtcwpt_background}. Then, \cref{subsec:appendix_dtcwpt_conv} shows that \dtcwpt performs convolutions with a subsampling factor $\subDepth$ which depends on the decomposition depth $\depth$. Finally, the Gabor-like nature of the convolution kernels is established in \cref{subsec:appendix_dtcwpt_properties}.

\subsection{Background}
\label{subsec:appendix_dtcwpt_background}

We provide a brief overview of the classical, real-valued 2D wavelet packet transform (WPT) algorithm \citep[p.~377]{Mallat2009}, before introducing the redundant, complex-valued and oriented \dtcwpt \citep{Bayram2008}.

\subsubsection{Discrete Wavelet Packet Transform}
\label{subsubsec:appendix_dtcwpt_background_wpt}

Given a pair of low- and high-pass 1D orthogonal filters $\qmfLow,\, \qmfHigh \in \realltwoZ$ satisfying a \emph{quadrature mirror filter} (QMF) relationship, we consider a separable 2D filter bank (FB), denoted by $\twodmultqmf := (\twodqmfHigh_\selectOutchannel)_{\selectOutchannel \in \zeroto{3}}$, defined by
\begin{align}
	\twodqmfHigh_0 &= \qmfLow \otimes \qmfLow; & \twodqmfHigh_1 &= \qmfLow \otimes \qmfHigh; & \twodqmfHigh_2 &= \qmfHigh \otimes \qmfLow; & \twodqmfHigh_3 &= \qmfHigh \otimes \qmfHigh.
\label{eq:2dfb}
\end{align}

Let $\inpimg \in \realltwoZ$. The decomposition starts with $\wptimgMultires{0}{0} = \inpimg$. Given $\selectDepth \in \mathN$, suppose that we have computed $4^\selectDepth$ sequences of wavelet packet coefficients at stage $\selectDepth$, denoted by $\wptimgMultiresSelectbis \in \realltwoZ$ for each $\selectOutchannel \in \zeroto{4^\selectDepth-1}$. They are referred to as \emph{feature maps}.

At stage $\selectDepth+1$, we compute a new representation of $\inpimg$ with increased frequency resolution---and decreased spatial resolution. It is obtained by further decomposing each feature map $\wptimgMultiresSelectbis$ into four sub-sequences, using subsampled (or strided) convolutions with kernels $\twodqmfHigh_\selectInchannel$, for each $\selectInchannel \in \zeroto{3}$:
\begin{equation}
	\forall \selectInchannel \in \zeroto{3},\; \wptimgMultires{\selectDepth + 1}{4\selectOutchannel+\selectInchannel} = \bigl(
        \wptimgMultiresSelectbis \ast \flipped{\twodqmfHigh_\selectInchannel}
    \bigr) \downarrow 2.
\label{eq:wpt}
\end{equation}

The algorithm stops after reaching the desired number of stages $\depth > 0$---referred to as \emph{decomposition depth}. Then,
\begin{equation}
	\iter{\dtmultimg}{\depth} := \bigl(
		\wptimgMultires{\depth}{\selectOutchannel}
	\bigr)_{\selectOutchannelInrangeWPT}
\label{eq:BBXitJ}
\end{equation}
constitutes a multichannel representation of $\inpimg$ in an orthonormal basis, from which the original image can be retrieved.

\subsubsection{Dual-Tree Complex Wavelet Packet Transform}
\label{subsubsec:appendix_dtcwpt_background_dtcwpt}

Despite having interesting properties such as sparse signal representation, WPT is unstable with respect to small shifts and suffers from a poor directional selectivity. To overcome this, \citet{Kingsbury2001} designed a new type of discrete wavelet transform, where images are decomposed in a redundant frame of nearly-analytic, complex-valued waveforms. It was later extended to the wavelet packet framework by \citet{Bayram2008}. The latter operation, referred to as \emph{dual-tree complex wavelet packet transform} (\dtcwpt), is performed as follows.

Let $(\qmfLowReal,\, \qmfHighReal)$ and $(\qmfLowImag,\, \qmfHighImag)$ denote two pairs of QMFs as defined in \cref{subsubsec:appendix_dtcwpt_background_wpt}, satisfying the \emph{half-sample delay} condition:
\begin{equation}
	\forall \omega \in \mpipi,\, \fourierQmfLowImag(\omega) = \rme^{-i \omega / 2} \, \fourierQmfLowReal(\omega).
\label{eq:halfsampleshift}
\end{equation}
Then, for any $\selectInchannel \in \zeroto{3}$, we build a 2D FB $\twodmultqmf_\selectInchannel := (\twodqmfHigh_{\selectInchannel,\, \selectOutchannel})_{\selectOutchannel \in \zeroto{3}}$ similarly to \eqref{eq:2dfb}:
\begin{align}
	\twodqmfHigh_{\selectInchannel,\, 0} &= \rmh_\selectFBOned \otimes \rmh_\selectFBOnedbis; &
	\twodqmfHigh_{\selectInchannel,\, 1} &= \rmh_\selectFBOned \otimes \rmg_\selectFBOnedbis; &
	\twodqmfHigh_{\selectInchannel,\, 2} &= \rmg_\selectFBOned \otimes \rmh_\selectFBOnedbis; &
	\twodqmfHigh_{\selectInchannel,\, 3} &= \rmg_\selectFBOned \otimes \rmg_\selectFBOnedbis,
\label{eq:2dfbdt}
\end{align}
where $\selectFBOned,\, \selectFBOnedbis \in \binaryset$ are defined such that $\selectInchannel = 2 \times \selectFBOned + \selectFBOnedbis$.%
\footnote{
	Actually, the FB design requires some technicalities which are not described here.
}

Let $\depth > 0$ denote a decomposition depth. Using each of the four FBs $\twodmultqmf_{0-3}$ as defined above, we assume that we have decomposed an input image $\inpimg$ into four multichannel WPT representations
$\iter{\dtmultimg}{\depth}_{0-3}$, each of which satisfies \eqref{eq:wpt} and \eqref{eq:BBXitJ}. Then, for any $\selectOutchannel \in \zeroto{4^\depth - 1}$, the following complex feature maps are computed:
\begin{equation}
\setlength\arraycolsep{2pt}
	\begin{pmatrix}
		\wptimgNEMultires{\depth}{\selectOutchannel} \\ \wptimgSEMultires{\depth}{\selectOutchannel}
	\end{pmatrix} =
	\begin{pmatrix}
		1 & -1 \\
		1 & 1
	\end{pmatrix}
	\begin{pmatrix}
		\wptimgFBMultires{0}{\depth}{\selectOutchannel} \\ \wptimgFBMultires{3}{\depth}{\selectOutchannel}
	\end{pmatrix}
	- i
	\begin{pmatrix}
		1 & 1 \\
		1 & -1
	\end{pmatrix}
	\begin{pmatrix}
		\wptimgFBMultires{2}{\depth}{\selectOutchannel} \\ \wptimgFBMultires{1}{\depth}{\selectOutchannel}
	\end{pmatrix}.
\label{eq:dtcwpt}
\end{equation}
As explained in \cref{subsec:appendix_dtcwpt_properties}, the feature maps of dual-tree coefficients have their Fourier transform restricted to a compact region of the frequency plane, and as such can be considered as Gabor-like coefficients. In the above expression, the arrow points to the Fourier quadrant where energy is concentrated. Furthermore, in the specific case where input images are real-valued, $\wptimgSWMultires{\depth}{\selectOutchannel}$ and $\wptimgNWMultires{\depth}{\selectOutchannel}$ are defined as the complex conjugates of the above feature maps, and therefore do not need to be explicitly computed. Then,
\begin{equation}
	\iter{\dtmultimg}{\depth} := \bigl(
		\wptimgNEMultires{\depth}{\selectOutchannel},\, \wptimgSEMultires{\depth}{\selectOutchannel},\, \wptimgSWMultires{\depth}{\selectOutchannel},\, \wptimgNWMultires{\depth}{\selectOutchannel}
	\bigr)_{\selectOutchannelInrangeWPT}
\end{equation}
constitutes a complex-valued, four-time redundant multichannel representation of $\inpimg$ from which the original image can be reconstructed.

\subsection{Convolution Operators}
\label{subsec:appendix_dtcwpt_conv}

We now show that \dtcwpt performs subsampled convolutions with Gabor-like filters, whose characteristics will be specified. First, we state the following lemma concerning the real-valued WPT algorithm, such as introduced in \cref{subsubsec:appendix_dtcwpt_background_wpt}. It is a simple reformulation of the well-known result that two successive convolutions can be written as another convolution with a wider kernel.

\begin{lemma}
	For any $\selectOutchannelInbigrangeWPT$, there exists $\weightimgMultires{\depth}_{\selectOutchannel} \in \realltwoZsq$ such that
	\begin{equation}
		\wptimgMultires{\depth}{\selectOutchannel} = \bigl(
			\inpimg \ast \flippedWeightimgMultires{\depth}_{\selectOutchannel}
		\bigr) \downarrow 2^\depth.
	\label{eq:conv_wpt}
	\end{equation}
\label{lemma:conv_wpt}
\end{lemma}

\begin{proof}
	We introduce the upsampling operator: $(\inpimg \uparrow \sub )[\vectorindex] := \inpimg[\vectorindex / \sub ]$ if $\vectorindex / \sub  \in \mathZ^2$, and $0$ otherwise. We also consider the ``identity'' filter $\idimg \in \realltwoZsq$ such that $\idimg[\bzero] = 1$ and $\idimg[\vectorindex] = 0$ otherwise.
	First, for any $\RMU,\, \RMV \in \realltwoZsq$ and any $s,\, t \in \mathN^\ast$, we have
	\begin{equation}
		((\RMU \downarrow s) \ast \RMV) \downarrow t = \left(\RMU \ast (\RMV \uparrow s)\right) \downarrow (st).
	\end{equation}
	Then, a simple reasoning by induction yields the result, with
	\begin{equation}
		\weightimgMultires{0}_0 := \idimg;
		\qquad
		\weightimgMultires{\selectDepth + 1}_{4\selectOutchannel+\selectInchannel} := \weightimgMultires{\selectDepth}_{\selectOutchannel} \ast \big(\twodqmfHigh_{\selectInchannel} \uparrow 2^\selectDepth\big)
	\end{equation}
	for any $\selectOutchannel \in \zeroto{\selectDepth-1}$ and any $\selectInchannel \in \zeroto{3}$.
\end{proof}

Based on \cref{lemma:conv_wpt}, the following proposition introduces complex kernels characterizing \dtcwpt.

\begin{proposition}
	For any $\selectOutchannelInbigrangeWPT$, there exists $\complexWeightimgNEMultires{\depth}_{\selectOutchannel} \in \complexltwoZsq$ such that \eqref{eq:conv_dtcwpt} is satisfied.
	Identical results are obtained with the three other Fourier quadrants.
\label{prop:conv_dtcwpt}
\end{proposition}

\begin{proof}
	For each of the four filter banks $\selectFBInrange$, and any channel $\selectOutchannelInbigrangeWPT$, \cref{lemma:conv_wpt} provides a convolution kernel $\weightimgFBMultires{\selectFB}{\depth}{\selectOutchannel} \in \realltwoZsq$ such that
	\begin{equation}
		\wptimgFBMultires{\selectFB}{\depth}{\selectOutchannel} = \left(
			\inpimg \ast \flippedWeightimgFBMultires{\selectFB}{\depth}{\selectOutchannel}
		\right) \downarrow 2^\depth.
	\label{eq:conv_dualwpt}
	\end{equation}
	Then, the result is obtained by plugging \eqref{eq:conv_dualwpt} into \eqref{eq:dtcwpt} for all $\selectFBInrange$, and by denoting
	\begin{equation}
		\setlength\arraycolsep{2pt}
			\begin{pmatrix}
				\complexWeightimgNEMultires{\depth}{\selectOutchannel} \\ \complexWeightimgSEMultires{\depth}{\selectOutchannel}
			\end{pmatrix} =
			\begin{pmatrix}
				1 & -1 \\
				1 & 1
			\end{pmatrix}
			\begin{pmatrix}
				\weightimgFBMultires{0}{\depth}{\selectOutchannel} \\ \weightimgFBMultires{3}{\depth}{\selectOutchannel}
			\end{pmatrix}
			+ i
			\begin{pmatrix}
				1 & 1 \\
				1 & -1
			\end{pmatrix}
			\begin{pmatrix}
				\weightimgFBMultires{2}{\depth}{\selectOutchannel} \\ \weightimgFBMultires{1}{\depth}{\selectOutchannel}
			\end{pmatrix}.
		\label{eq:dtcwpt_weightimg}
		\end{equation}
\end{proof}

\begin{remark}
\label{remark:conv_dtcwpt_continuous}
	\dtcwpt, computed on a discrete image $\inpimg$, approximates the decomposition of a continuous 2D signal $\inpfun \in \realLtwoRsq$ into a tight frame
	\begin{equation}
		\complexMultwaveletFinal{\depth}
        := \biguplus\limits_{\selectOutchannel=0}^{4^\depth - 1} \waveletMultiresComplexFamily{\depth}{\selectOutchannel}.
	\label{eq:complexframemultiresPseudocosine2}
	\end{equation}
	In this context, the feature maps of dual-tree wavelet packet coefficients satisfy
	\begin{equation}
		\wptimgNEMultires{\depth}{\selectOutchannel}[\vectorindex] \approx \Bigl(
			\inpfun \ast {\flippedWaveletNEMultiresZero{\depth}{\selectOutchannel}}^\ast
		\Bigr)(2^\depth \vectorindex),
		\qqwith
		\waveletNEMultiresZero{\depth}{\selectOutchannel} := \waveletNEMultires{\depth}{\selectOutchannel}{\bzero}.
	\label{eq:conv_dtcwpt_continuous}
	\end{equation}
	Expression \eqref{eq:conv_dtcwpt_continuous} is only an approximation because of implementation technicalities that occur in practice. A ``perfect'' dual-tree transform should be initialized with four different inputs $\inpimgFB{0-3}$. Instead, all four WPT decompositions operate on the same input image $\inpimg$, leading to non-analytic outputs for small values of $\depth$. In order to counterbalance this shortcoming, the first stage of \dtcwpt decomposition must be performed with a special set of filters that satisfy the \emph{one-sample delay} condition. We refer to \citet{Selesnick2005} for more details on this matter.
\end{remark}

\subsection{Gabor-Like Convolution Kernels}
\label{subsec:appendix_dtcwpt_properties}

In this section, we show that the convolution kernels $\complexWeightimgNEMultires{\depth}_{\selectOutchannel}$ and $\complexWeightimgSEMultires{\depth}_{\selectOutchannel}$, introduced in \eqref{eq:conv_dtcwpt}, approximately behave as Gabor-like filters, as defined in \eqref{eq:gaborfilt_discrete}. To begin with, we assume that $\qmfLowReal$ is a Shannon filter, which is associated with a sinc scaling function \citep{Shannon1949}. Let $\depth \in \nonzeroMathN$ denote the number of decomposition stages. The following proposition states that \dtcwpt tiles the frequency plane with square windows.

\begin{proposition}
	\label{prop:supp_dtcwpt}%
	There exists a permutation $\bigl(\permutvecDepthSelect\bigr)_{\selectOutchannelInrangeWPT}$ of $\bigrange{0}{2^\depth - 1}^2$ such that, for any $\selectOutchannelInbigrangeWPT$,
	\begin{equation}
		\waveletNEMultiresZero{\depth}{\selectOutchannel} \in \GaborfilterDt,
	\label{eq:dtkernel_gaborfilt}
	\end{equation}
	where $\waveletNEMultiresZero{\depth}{\selectOutchannel}$ has been introduced in \cref{remark:conv_dtcwpt_continuous}, and where we have defined
	\begin{equation}
		\freqvecMpipiMultires{\depth}_\selectOutchannel := \left(
			\permutvecDepthSelect + \frac12
		\right) \frac{
			\pi
		}{2^\depth}
			\qqand
		\supportsizeDiscrete_\depth := \frac{\pi}{2^\depth}.
	\label{eq:supp_dtcwpt}
	\end{equation}
	We remind the reader that $\GaborfilterGen$, defined in \eqref{eq:gaborfilt_continuous}, denotes a space of Gabor-like filters in the continuous framework.
\end{proposition}

\begin{proof}
	The atoms $\waveletNEMultiresZero{\depth}{\selectOutchannel}$ of the wavelet packet tight frame $\complexMultwaveletFinal{\depth}$ can be written as the tensor product of two 1D wavelet packets:
	\begin{equation}
		\waveletNEMultiresZero{\depth}{\selectOutchannel} = \waveletOnedMultires{\depth}{\selectOutchannel_1} \otimes \waveletOnedMultires{\depth}{\selectOutchannel_2},
	\label{eq:dtcwpt_tensorprod}
	\end{equation}
	for some indices $\selectOutchannel_1$ and $\selectOutchannel_2 \in \bigsetof{2^\depth}$. Moreover, for any $\selectOutchannel' \in \bigsetof{2^\depth}$, we have
	\begin{equation}
		\waveletOnedMultires{\depth}{\selectOutchannel'} = \waveletOnedRealMultires{\depth}{\selectOutchannel'} + i \, \waveletOnedImagMultires{\depth}{\selectOutchannel'},
	\end{equation}
	where $\waveletOnedRealMultires{\depth}{\selectOutchannel'} \in \realLtwoR$ is an atom of the standard Shannon wavelet packet orthonormal basis, and $\waveletOnedImagMultires{\depth}{\selectOutchannel'}$ is the one-dimensional Hilbert transform of $\waveletOnedRealMultires{\depth}{\selectOutchannel'}$. Therefore, since the Hilbert transform suppresses negative frequencies, we get
	\begin{equation}
		\fourierWaveletOnedMultires{\depth}{\selectOutchannel'} = 2 \, \fourierWaveletOnedRealMultires{\depth}{\selectOutchannel'} \, \characfun_{\mathR_+}.
	\end{equation}
	Consequently, according to the Coifman-Wickerhauser theorem \citep[pp.~384-385]{Mallat2009}, there exists $\selectInchannel \in \bigsetof{2^\depth}$ such that
	\begin{equation}
		\supp \fourierWaveletOnedMultires{\depth}{\selectOutchannel'} \subset \interval{
			\frac{\selectInchannel\pi}{2^\depth}
		}{
			\frac{(\selectInchannel + 1)\pi}{2^\depth}
		}.
	\end{equation}
	Finally, the tensor product \eqref{eq:dtcwpt_tensorprod} yields the result.
\end{proof}

According to \cref{prop:supp_dtcwpt}, each atom $\waveletNEMultiresZero{\depth}{\selectOutchannel}$, for $\selectOutchannelInbigrangeWPT$, is supported in a square window of size $\supportsizeDiscreteDepth \times \supportsizeDiscreteDepth$ included in the top-right quadrant of the Fourier domain. Similar results can be obtained for the three remaining quadrants, with $\waveletSEMultiresZero{\depth}{\selectOutchannel}$, $\waveletSWMultiresZero{\depth}{\selectOutchannel}$ and $\waveletNWMultiresZero{\depth}{\selectOutchannel}$.
We would like to deduce from \cref{prop:supp_dtcwpt} that the discrete filter $\complexWeightimgNEMultires{\depth}_{\selectOutchannel} \in \complexltwoZsq$ satisfies the Gabor property \eqref{eq:dtkernel_gaborfilt_discrete}. However, as mentioned in \cref{remark:conv_dtcwpt_continuous}, \eqref{eq:conv_dtcwpt_continuous} is only an approximation.
In fact, the Fourier support of $\complexWeightimgNEMultires{\depth}_{\selectOutchannel}$ is contained in four square regions of size $\supportsizeDiscrete_\depth$ (one in each quadrant), its energy becoming negligible outside the top-right quadrant when $\depth$ increases.
Nevertheless, employing, in the first stage, a specific pair of low-pass filters satisfying the one-sample delay condition \citep{Selesnick2005} yields near-analytic solutions even for small values of $\depth$. We therefore assume that \eqref{eq:dtkernel_gaborfilt_discrete} is a reasonable approximation if $\depth \geq 2$.

\begin{remark}
	\label{remark:fouriersupportsize}
	\Cref{prop:supp_dtcwpt} tiles the top-right Fourier quadrant with $4^\depth$ square cells of size $\supportsizeDiscreteDepth := \pi / 2^\depth$. However, the Shannon wavelet is poorly suited for sparse image representations, because of its slow decay rate. Moreover, it deviates from what is typically observed in freely-trained CNNs, because $\complexWeightimgNEMultires{\depth}_{\selectOutchannel}$ must be approximated with very large filters to avoid numerical instabilities. Practical implementations of \dtcwpt use fast-decaying filters such as these associated to Meyer wavelets \citep{Meyer1985}, or finite-length filters that approximate the half-sample delay condition \citep{Selesnick2005}. Therefore, energy is leaking outside the square cells tiling the Fourier domain. To counterbalance this, we increase the window size up to
	\begin{equation}
		\supportsizeDiscreteDepth := \frac{\pi}{2^{\depth-1}} = \pi / \subDepth,
	\end{equation}
	and suggest that \eqref{eq:dtkernel_gaborfilt_discrete} remains a reasonable approximation. Therefore, the conditions to apply \cref{th:shiftinvariance_cmod,th:scdmoment_normdiff_cmodrmax,th:shiftinvariance_rmax} are approximately satisfied in this context.

	\begin{table}
		\renewcommand{\arraystretch}{1.2}
		\begin{center}
			\begin{footnotesize}
				\addtolength{\tabcolsep}{-1pt}
				\begin{sc}
					\begin{tabular}{|r|c|r|r|}
							\hline
						\multicolumn{1}{|c|}{Depth $\depth$} & \multicolumn{1}{c|}{Bandwidth $\supportsizeDiscreteDepth$} & \multicolumn{1}{c|}{Mean} & \multicolumn{1}{c|}{Std} \\
							\hline
						$2$ & $\pi / 2$ & $0.98$ & $0.00$ \\
						$3$ & $\pi / 4$ & $0.95$ & $0.02$ \\
							\hline
					\end{tabular}
				\end{sc}
			\end{footnotesize}
		\end{center}
		\caption[%
			Gabor-likeness of \dtcwpt Filters%
		]{Energy concentration of the \dtcwpt filters within a Fourier window of size $\supportsizeDiscreteDepth \times \supportsizeDiscreteDepth$, with $\supportsizeDiscreteDepth := \pi / 2^{\depth - 1}$.}
	\label{tab:energyratio_fourierwindow}
	\end{table}
	
	In order to numerically assess this assumption, we measured the maximum percentage of energy within a square window of size $\supportsizeDiscreteDepth \times \supportsizeDiscreteDepth$ in the Fourier domain:
	\begin{equation}
		\energyratioSelect^{\topright} := \frac{
			\max_{
				\freqvecMpipi \in \mpipi^2
			} \BignormLtwo{
				\characfun_{\linftyBall(\freqvecMpipi,\, \supportsizeDiscreteDepth/2)}\fourierComplexWeightimgNEMultires{\depth}_{\selectOutchannel}
			}^2
		}{
			\BignormLtwo{
				\fourierComplexWeightimgNEMultires{\depth}_{\selectOutchannel}
			}^2
		},
	\label{eq:compute_kernelbandwidth}
	\end{equation}
	where the $l^\infty$-ball $\linftyBall(\freqvecMpipi,\, \supportsizeDiscreteDepth/2)$ is defined in the quotient space $\mpipi^2 / \allowbreak(2\pi\mathZ^2)$, as explained in \cref{remark:quotientspace}. If \eqref{eq:dtkernel_gaborfilt_discrete} is perfectly satisfied, then $\energyratioSelect^{\topright} = 1$. The statistics computed over the collection $\bigl(
		\energyratioSelect^{\topright},\, \energyratioSelect^{\bottomright}
	\bigr)_{\selectOutchannelInrangeWPT}$ are reported in \cref{tab:energyratio_fourierwindow}.
\end{remark}

\begin{remark}
	For ``boundary filters'', \ie, when $\bignorminfty{\freqvecMpipiMultires{\depth}_\selectOutchannel} = \bigl(1 - 2^{-(\depth+1)}\bigr) \, \pi$, \cref{remark:quotientspace} states that a small fraction of the filter's energy remains located at the far end of the Fourier domain---see also \citet{Bayram2008}. Therefore, these filters do not strictly comply with the conditions of \cref{th:shiftinvariance_cmod,th:scdmoment_normdiff_cmodrmax,th:shiftinvariance_rmax}. We nevertheless include them in our experiments.
\end{remark}

\end{document}